\documentclass[twoside,11pt]{article}
\usepackage{jmlr2e}
    \usepackage{graphicx,url}
    \usepackage[linesnumbered,ruled,algo2e]{algorithm2e}
    
    \usepackage{etoolbox}

    \usepackage{subfigure}

\usepackage{hyperref}

\hypersetup{
  colorlinks = true,
  urlcolor = blue, 
  linkcolor = blue,
  citecolor = blue,
  pdfborder={0 0 0},
  pagebackref=true
}

\usepackage{algpseudocode}

\usepackage{booktabs} 
\usepackage{caption}
\usepackage[usestackEOL]{stackengine}
\usepackage{enumitem}

\usepackage{etoc}

\usepackage{mathrsfs}
\usepackage{upgreek}
\usepackage{adjustbox}
\usepackage{url}

\usepackage{xr}

\makeatletter
 
\usepackage{graphics}
\usepackage{graphicx}
\usepackage{psfrag}

\usepackage{color}

\usepackage{amssymb,bbm}

\usepackage{nicefrac}

 \usepackage{tabularx}%

\usepackage{enumitem}
\usepackage{booktabs}
\usepackage{caption}

\usepackage{mathtools}
\graphicspath{../fig/}

\newcommand{\qtext}[1]{\quad\text{#1}\quad}

\newcommand{\Fgrad}{\ensuremath{F^{\tiny{\mbox{GRD}}}}}
\newcommand{\Fnewt}{\ensuremath{F^{\tiny{\mbox{NWT}}}}}
\newcommand{\stepsize}{\ensuremath{h}}

\newcommand{\event}{\mathcal{E}}

\newcommand{\samples}{\ensuremath{n}}
\newcommand{\obs}{\samples}
\newcommand{\real}{\ensuremath{\mathbb{R}}}

\newcommand{\smallthreshold}{\alpha}

\newcommand{\order}[1]{\ensuremath{\mathcal{O}\parenth{#1}}}

\newcommand{\thetastar}{\ensuremath{\theta^*}}

\newcommand{\NORMAL}{\ensuremath{\mathcal{N}}}

\newcommand{\brackets}[1]{\left[ #1 \right]}
\newcommand{\parenth}[1]{\left( #1 \right)}

\newcommand{\braces}[1]{\left\{ #1 \right \}}
\newcommand{\abss}[1]{\left| #1 \right |}

\newcommand{\ceil}[1]{\left\lceil #1 \right\rceil}

\newcommand{\numsteps}{T}
\newcommand{\totalnumsteps}{S}
\newcommand{\seqalpha}[1]{\lambda_{#1}}

\newcommand{\imax}{\ind_\smallthreshold}

\newcommand{\sech}{\text{sech}}
\newcommand{\constfn}{c'}

\newcommand{\iter}{t}

\newcommand{\usedim}{\ensuremath{d}}

\newcommand{\ind}{\ensuremath{\ell}}

\newcommand{\Rspace}{\ensuremath{\mathbb{R}}}

\newcommand{\Ncal}{\ensuremath{\mathcal{N}}}

\newcommand{\ball}{\ensuremath{\mathbb{B}}}

\newcommand{\fastcont}{\ensuremath{\kappa}}

\newcommand{\FitDensity}{\ensuremath{f}_{\theta}}

\newcommand{\normDensity}{\ensuremath{\phi}}

\newcommand{\sd}{\ensuremath{\sigma}}

\newcommand{\mydefn}{\ensuremath{:=}}

\newcommand{\likelihood}{\ensuremath{\mathcal{L}}}
\newcommand{\samlikelihood}{\ensuremath{\mathcal{L}_{n}}}

\newcommand{\likelihoodres}{\ensuremath{\bar{\mathcal{L}}}}
\newcommand{\likelihoodsing}{\ensuremath{\widetilde{\mathcal{L}}}}
\newcommand{\mullikelihoodsing}{\ensuremath{\widetilde{\mathcal{L}}}}
\newcommand{\regular}{\ensuremath{L}}

\newcommand{\unicon}{c}

\newcommand{\unicontwo}{c_2}

\newcommand{\defn}{:=}
\newcommand{\rdefn}{=:}
\newcommand{\etal}{{et al.}}

\newcommand{\enorm}[1]{\ensuremath{\| #1 \|}} 

\newcommand{\inprod}[2]{\ensuremath{\langle #1 , \, #2 \rangle}}

\newcommand{\Exs}{\ensuremath{{\mathbb{E}}}}
\newcommand{\Prob}{\ensuremath{{\mathbb{P}}}}

\newcommand{\polyfunc}{\ensuremath{f}}

\newcommand{\learnrate}{\ensuremath{\eta}}

\newcommand{\opindex}[2]{\mathrm{#1}^{#2}}

\newcommand{\popgd}{\opindex{\pop}{\mathrm{GD}}}
\newcommand{\nopgd}{\popgd_\obs}

\newcommand{\popga}{\opindex{\pop}{{\mathrm{GA}}}}
\newcommand{\nopga}{\popga_{\obs}}

\newcommand{\popcnm}{\opindex{\pop}{\scriptsize{\mathrm{CNM}}}}
\newcommand{\nopcnm}{\popcnm_{\obs}}

\newcommand{\bigc}{C}
\newcommand{\tfun}{\widetilde{\mathcal{T}}}
\newcommand{\counter}{k}

\newcommand{\radius}{r}
\newcommand{\lradius}{\rho}
\newcommand{\ballnotag}{\ensuremath{\mathbb{B}}}
\newcommand{\ballstar}{\ensuremath{\mathbb{B}}(\tstar,\lradius)}
\newcommand{\lipschitz}{\ensuremath{{\mathrm{LL}}(\lradius)}}

\newcommand{\fastnotag}{\texttt{\small{FAST}}}
\newcommand{\slownotag}{\texttt{\small{SLOW}}}
\newcommand{\slow}{\texttt{\slownotag($\parbeta$)}}
\newcommand{\fast}{\texttt{\fastnotag($\fastcont$)}}

\newcommand{\stabilitynotag}{\texttt{\small{STA}}}
\newcommand{\instabilitynotag}{\texttt{\small{UNS}}}

\newcommand{\stability}{\texttt{\stabilitynotag($\pargamma$)}}
\newcommand{\instability}{\texttt{\instabilitynotag($\pargamma$)}}

\newcommand{\annulus}{\mathbb{A}}
\newcommand{\Ann}{\annulus}

\newcommand{\noiseunstable}{\widetilde{\noise}}

\newcommand{\resga}{\opindex{M}{{\mathrm{GA}}}}
\newcommand{\samresga}{\resga_{\obs}}

\newcommand{\cnm}{\mathrm{\scriptsize{CNM}}}

\newcommand{\polygd}{\opindex{Q}{{\mathrm{GD}}}}
\newcommand{\sampolygd}{\polygd_{\obs}}

\newcommand{\polycnm}{\opindex{Q}{\cnm}}
\newcommand{\sampolycnm}{\polycnm_{\obs}}

\newcommand{\popem}{\opindex{G}{\mathrm{EM}}}
\newcommand{\nopem}{\popem_{\obs}}

\newcommand{\newt}{\tiny{NM}}
\newcommand{\newem}{\opindex{G}{\mathrm{\newt}}}
\newcommand{\samnewem}{\newem_{\obs}}

\newcommand{\resnm}{\opindex{M}{\mathrm{\newt}}}
\newcommand{\samresnm}{\resnm_{\obs}}

\newcommand{\polynm}{\opindex{Q}{\mathrm{\newt}}}
\newcommand{\sampolynm}{\polynm_{\obs}}

\newcommand{\popnm}{\opindex{\pop}{{\mathrm{\newt}}}}
\newcommand{\nopnm}{\popnm_{\obs}}

\newcommand{\newdndelta}{\ensuremath{r_{*}}}

\newlength{\widebarargwidth}

\makeatletter
\long\def\@makecaption#1#2{
        \vskip 0.8ex
        \setbox\@tempboxa\hbox{\small {\bf #1:} #2}
        \parindent 1.5em  
        \dimen0=\hsize
        \advance\dimen0 by -3em
        \ifdim \wd\@tempboxa >\dimen0
                \hbox to \hsize{
                        \parindent 0em
                        \hfil
                        \parbox{\dimen0}{\def\baselinestretch{0.96}\small
                                {\bf #1.} #2
                                }
                        \hfil}
        \else \hbox to \hsize{\hfil \box\@tempboxa \hfil}
        \fi
        }
\makeatother

\long\def\comment#1{}
\definecolor{carnelian}{rgb}{0.7, 0.11, 0.11}
\definecolor{battleshipgrey}{rgb}{0.52, 0.52, 0.51}
\definecolor{darkgray}{rgb}{0.66, 0.66, 0.66}
\definecolor{indiagreen}{rgb}{0.07, 0.53, 0.03}
\definecolor{darkgreen}{rgb}{0.0, 0.2, 0.13}
\definecolor{darkspringgreen}{rgb}{0.09, 0.45, 0.27}
\definecolor{dukeblue}{rgb}{0.0, 0.0, 0.61}
\definecolor{olivedrab7}{rgb}{0.24, 0.2, 0.12}
\definecolor{darkblue}{rgb}{0.0, 0.0, 0.55}
\definecolor{darkscarlet}{rgb}{0.34, 0.01, 0.1}
\definecolor{candyapplered}{rgb}{1.0, 0.03, 0.0}
\definecolor{ao(english)}{rgb}{0.0, 0.5, 0.0}
\definecolor{applegreen}{rgb}{0.55, 0.71, 0.0}

\newcommand{\empirical}{empirical }

\newcommand{\widgraph}[2]{\includegraphics[keepaspectratio,width=#1]{#2}}

\newcommand{\noise}{\ensuremath{\varepsilon}}
\newcommand{\pop}{\ensuremath{F}}
\newcommand{\nop}{\ensuremath{F_n}}

\newcommand{\mydndelta}{\specialeps}

\newcommand{\paralpha}{\ensuremath{\alpha}}
\newcommand{\parbeta}{\ensuremath{\beta}}
\newcommand{\pargamma}{\ensuremath{\gamma}}
\newcommand{\pargammaMod}{\ensuremath{{|\gamma|}}}

\newcommand{\parfinal}{\ensuremath{\nu}}
\newcommand{\tol}{\delta}
\newcommand{\radii}{\ensuremath{\mathcal{R}}}
\newcommand{\tvar}{\theta}
\newcommand{\tvarn}{\tvar_{\obs}}
\newcommand{\tstar}{\tvar^\star}

\newcommand{\parnuone}{\ensuremath{\parfinal}}
\newcommand{\parnutwo}{\ensuremath{\parfinal'}}

\newcommand{\tfunUnstable}{\ensuremath{\widetilde{\mathcal{T}}}}

\newcommand{\innerradius}{\widetilde{\lradius}}
\newcommand{\inradius}{\lradius_{\textrm{in}}}
\newcommand{\outradius}{\lradius_{\textrm{out}}}

\newcommand{\newnoise}{\xi} 
\newcommand{\estimate}{\widehat{\tvar}_\obs}
\newcommand{\specialeps}{\ensuremath{\noise(\obs, \delta^*)}}

\newcommand{\numobs}{\obs}
\newcommand{\betapar}{\parbeta}
\newcommand{\ds}{\displaystyle}
\newcommand{\localmle}{\widehat{\tvar}_ {\obs,\mathrm{MLE}}}

\makeatletter
\long\def\@makecaption#1#2{
        \vskip 0.8ex
        \setbox\@tempboxa\hbox{\small {\bf #1:} #2}
        \parindent 1.5em  
        \dimen0=\hsize
        \advance\dimen0 by -3em
        \ifdim \wd\@tempboxa >\dimen0
                \hbox to \hsize{
                        \parindent 0em
                        \hfil 
                        \parbox{\dimen0}{\def\baselinestretch{0.96}\small
                                {\bf #1.} #2
                                } 
                        \hfil}
        \else \hbox to \hsize{\hfil \box\@tempboxa \hfil}
        \fi
        }
\makeatother

\ShortHeadings{Instability, Computational Efficiency and Statistical Accuracy}{Ho, Khamaru, Dwivedi, Wainwright, Jordan, Yu}
\firstpageno{1}

\begin{document}



\title{Instability, Computational Efficiency and Statistical Accuracy}

\author{\name Nhat Ho$^{\star}$ \email minhnhat@utexas.edu \\
\addr Department of Statistics and Data Sciences \\
University of Texas, Austin
\AND 
\name Koulik Khamaru$^{\star}$\email koulik@berkeley.edu \\
\addr Department of Statistics \\
University of California, Berkeley
\AND 
\name Raaz Dwivedi$^{\star}$ \email raaz.rsk@berkeley.edu \\
       \addr Department of EECS\\
       University of California, Berkeley
       \AND 
       \name Martin J. Wainwright \email wainwrig@berkeley.edu \\
       \addr Department of EECS, Department of Statistics \\
       University of California, Berkeley
       \AND
       \name Michael I. Jordan \email jordan@cs.berkeley.edu \\
       \addr Department of EECS, Department of Statistics \\
       University of California, Berkeley
       \AND
       \name Bin Yu  \email binyu@berkeley.edu \\
       \addr Department of EECS, Department of Statistics \\
       University of California, Berkeley}

\editor{Francis Bach, David Blei, and Bernhard Sch{\"o}lkopf}

\maketitle

\begin{abstract}
 Many statistical estimators are defined as the fixed point of a
  data-dependent operator, with estimators based on minimizing a
  cost function being an important special case.  The limiting
  performance of such estimators depends on the properties of the
  population-level operator in the idealized limit of infinitely many
  samples.  We develop a general framework that yields bounds on
  statistical accuracy based on the interplay between the
  deterministic convergence rate of the algorithm at the population
  level, and its degree of (in)stability when applied to an empirical
  object based on $n$ samples.  Using this framework, we analyze both
  stable forms of gradient descent and some higher-order and unstable
  algorithms, including Newton's method and its cubic-regularized
  variant, as well as the EM algorithm. We provide applications of our
  general results to several concrete classes of models, including
  Gaussian mixture estimation, non-linear regression models, and informative
  non-response models.  We exhibit cases in which an unstable
  algorithm can achieve the same statistical accuracy as a stable
  algorithm in exponentially fewer steps---namely, with the number of
  iterations being reduced from polynomial to logarithmic in sample
  size $n$.
\end{abstract}
\let\thefootnote\relax\footnotetext{$\star$ Raaz Dwivedi, Nhat Ho,
  and Koulik Khamaru contributed equally to this work.}




\section{Introduction} 
\label{sec:introduction}

The interplay between the stability and computational efficiency of
optimization algorithms has long been a fundamental problem in
statistics and machine learning.  The stability of the algorithm, a
classical desideratum, is often believed to be a necessity for
obtaining efficient statistical estimators. Such a belief rules out
the use of a variety of faster algorithms due to their instability.
This paper shows that this popular belief can be misleading: the
situation is more subtle in that there are various settings in which
unstable algorithms may be preferable to their stable counterparts.

Recent years have seen a significant body of work involving
performance of various machine-learning algorithms when applied to
statistical estimation problems.  Examples include sparse signal
recovery~\citep{hale2008fixed,garg2009gradient,beck2009fast,becker2011nesta},
more general forms of
M-estimation~\citep{agarwal2012fast,zhang2012general,loh2015regularized},
principal component
analysis~\citep{amini2008high,ma2013sparse,yuan2013truncated},
regression with concave
penalties~\citep{loh2015regularized,wang2014optimal}, phase retrieval
problems retrieval
(e.g.,~\citep{Candes-2012,Candes-2015,CheWai15,zhang2017nonconvex,chen2018gradient}),
and mixture model
estimation~\citep{Siva_2017,Fanny-2017,Cai_2018,yi2015regularized}. 

A unifying theme in these works is to study, in a finite-sample
setting, the computational efficiency of different algorithms and the
statistical accuracy of the resulting estimates.  For estimators based
on solving optimization problems that are convex, standard algorithms
and theory can be applied.  However, many modern estimators arise from
non-convex optimization problems, in which case the associated
algorithms become more complex to understand.  But evidence is
accumulating for the practical and theoretical advantages of such
algorithms.  For instance, the paper~\citep{agarwal2012fast}
established the fast convergence of projected gradient descent (GD)
for high-dimensional signal recovery in a weakly convex setting,
whereas the papers~\citep{loh2015regularized,wang2014optimal} provided
similar guarantees for a class of non-convex learning problems.  Other
work has demonstrated fast convergence of the truncated power method
for PCA~\citep{yuan2013truncated}, analyzed the behavior of projected
gradient methods for low-rank matrix recovery~\citep{CheWai15}, and
characterized the behavior of gradient descent for phase-retrieval
problems~\citep{chen2018gradient}.  Additionally, there is also a
recent line on work on the fast convergence of EM for various types of
mixture models~\citep{Siva_2017,Fanny-2017,Cai_2018}.  Finally, there
is a line of
work~\citep{hardt16,chen2018stability,kuzborskij2018data,charles2018stability}
that provides statistical error bounds for generic machine learning
problems (with certain assumptions on loss functions) in terms of
estimators obtained via iterative optimization algorithms (e.g.,
stochastic gradient methods).


\subsection{Population-to-sample or stability-based analysis} 
\label{sub:population_to_sample_or_stability_based_analysis}

The analysis in these works falls into two distinct categories.  The
first is a \emph{direct analysis}, in which one directly characterizes
the behavior of the iterates of the algorithm on the finite-sample
objective.  A long line of papers has used the direct approach
(e.g.,~\citep{agarwal2012fast,loh2015regularized,wang2014optimal,zhang2012general,yuan2013truncated})
to demonstrate that certain optimization algorithms converge at
geometric rates to a local neighborhood of the true parameter, with
the radius proportional to the statistical minimax risk.  The second
kind of analysis is more indirect and can be referred to as
\emph{population-to-sample analysis} or \emph{stability-based
analysis} where one analyzes the algorithmic convergence of
population-level iterates, and derives statistical errors for the
sample-level updates via uniform laws for stability/perturbation
bounds.  These approaches have been used to analyze the performance of
EM and its variants in several statistical settings, see the
papers~\citep{Siva_2017,Cai_2018,Fanny-2017,yi2015regularized,Raaz_Ho_Koulik_2018,Raaz_Ho_Koulik_2018_second}
and the references therein.  In general settings, it has been used to
derive statistical errors for iterates from stochastic optimization
methods~\citep{hardt16,chen2018stability,kuzborskij2018data,charles2018stability}.

The contributions of this paper build upon the stability-based
analysis, so let us discuss it in a little more detail.  Let $\pop$
and $\nop$ denote the operators that define the iterates at the
population level, corresponding to the idealized limit of an infinite
sample size, and sample-level based on a dataset of size $n$. Suppose
$\tstar$ denotes the parameter of interest, such that the
population-level iterates defined as $\tvar^t = \pop(\tvar^{t-1})$ for
$t=1, 2, \ldots$ with initialization $\tvar^0$, i.e., $\tvar^t =
\pop^t(\tvar^0)$, converge to $\tstar$ as $t\to \infty$.  Of interest
is to characterize the best possible estimate of $\tstar$ obtained
from the sample-based (noisy) iterates, defined as $\theta_{n}^{t}
=\nop^t (\theta^0)$ (with initialization $\theta^0$), and possibly
characterize the change in the error$\enorm{\nop^t(\tvar^0) - \tstar}$
as a function of the iteration $\iter$ and the sample size $\obs$.
The population-to-sample or the stability analysis proceeds by using
the following decomposition:

\begin{align}
\label{eq:ftml}
    \nop^t(\theta^0) - \tstar
    &= \underbrace{\pop^t(\theta^0) - \tstar}_{=:\varepsilon_\text{opt}^t}
    +
    \underbrace{\nop^t(\theta^0)
    - \pop^t(\theta^0)}_{=:\varepsilon_\text{stab}^t}.
\end{align}
Given this decomposition, the analysis proceeds in two steps:\vspace{-1.2mm}
\begin{itemize}
\item The first step is a deterministic convergence analysis of the
  algorithm to the true parameter at the population-level, namely, obtain
  a control on the \emph{optimization error} $\varepsilon_\text{opt}^t$
  as
  a function of $t$. 
\item The second step is to perform a stability analysis of the difference
between the population and the sample-based iterates, namely, obtain a control
on the \emph{perturbation/stability error} $\varepsilon_\text{stab}^t$ as
a function of $t$.
\end{itemize}
The ultimate convergence guarantee---what statistical error can be achieved
with the sample-based operator $\nop$, and in how many iterations---is then
derived based on the interplay between the two errors in equation~\eqref{eq:ftml}, 
namely, $\varepsilon_\text{opt}^t$ and $\varepsilon_\text{stab}^t$. \\

\paragraph{The ERM-based approach:}
We remark that the decomposition in equation~\eqref{eq:ftml} is different
from that used when invoking the uniform laws for the empirical
risk minimizer (ERM). Assuming the sample-based iterates converges to the
ERM, i.e., $\lim_{\iter\to \infty}F_n^{\iter}
(\theta^0) =
\widehat{\theta}_{\textrm{ERM}}$, the typical decomposition in the ERM-based
approach is given
by 
\begin{align*}
 F_n^{\iter}(\theta^0) -\tstar = \underbrace{F_n^{\iter} (\theta^0) - \widehat{\theta}_
 {\textrm{ERM}}}_{=:\varepsilon_{\textrm{opt-sample}}^t} + 
 \underbrace{\widehat{\theta}_{\textrm{ERM}} - \tstar}_{=:\varepsilon_{\textrm{unif-gen}}}.
\end{align*}
Here the first term in the RHS corresponds to the \emph{optimization
  error at the sample-level} at iteration $\iter$ and the second term
corresponds to the (iteration-independent) \emph{uniform
  generalization bound.}  Depending on the application, a precise
characterization of either of these terms can be non-trivial;
moreover, applying uniform bounds to control the term
$\varepsilon_{\textrm{unif-gen}}$ may lead to bounds that are overly
loose.  In such settings, the population-to-sample or stability-based
analysis can prove to be a useful alternative.

%

\subsection{Past works focus on stable methods} 
\label{sub:the_role_of_stability}
Most of the past work with the population-to-sample analysis
has focused on algorithms whose updates are \emph{stable}, meaning that
the perturbation error between sample-level and population-level iterates
decays to zero as
the iterates approach the true parameter.  For example, the
papers~\citep{Siva_2017,Cai_2018,Fanny-2017,yi2015regularized} used
this framework for problems where the population updates converge at a
geometric rate to the true parameter, and iterates based on $\obs$
samples yield an estimate within $\obs^{-1/2}$ of the true parameter.
On the other hand, other
papers~\citep{Raaz_Ho_Koulik_2018,Raaz_Ho_Koulik_2018_second} have
shown that with over-specified Gaussian mixtures, the EM algorithm,
which is a stable algorithm, takes a large number of steps to find an
estimate whose statistical error is of order $n^{-1/4}$ or
$n^{-1/8}$. Although for those problems the larger final statistical error of 
EM is minimax optimal, several natural questions remained unanswered: Can
an algorithm converge to a statistically optimal
estimate in significantly fewer steps than EM for over-specified
mixtures?  Moreover, will the faster algorithm continue to be stable?
Besides the analysis in recent works~\citep{Raaz_Ho_Koulik_2018,Raaz_Ho_Koulik_2018_second}
relied heavily on the facts that the EM updates had closed-form analytical
expressions. To our best knowledge, general statistical guarantees
for a generic stable or unstable algorithm (without a closed-formed expression)
when the algorithmic convergence is slow, are not present
in the literature.

In past work,~\cite{chen2018gradient} provided a trade-off
between stability and number of iterations to converge.  In particular, they showed that the minimax error of a problem class
forces a trade-off between the two errors in equation~\eqref{eq:ftml},
$\varepsilon_\text{opt}^t$ and $\varepsilon_\text{stab}^t$, for any
iterative algorithm used for solving it. In simple words, given the
minimax error, an algorithm that converges quickly is necessarily
unstable\footnote{There is a subtle difference in the definition of
  (in)stability used in Chen~\etal's work~\citep{chen2018stability}
  compared to ours.  In their work, stability refers to a \emph{slow}
  growth in the error $\enorm{\pop^\iter (\tvar)-\nop^\iter(\tvar)}$
  with number of iterations $\iter$, where slow is defined in a
  relative sense with other methods.  In our case, we use stability
  for the settings when $\enorm{\pop(\tvar)-\nop (\tvar)}$ decreases
  with $\enorm{\theta - \tstar}$ as $\tvar\to \tstar$.}, and
conversely, a stable algorithm cannot converge quickly. Their work,
however, did not address the following converse questions: Under what conditions does an algorithm, either stable or unstable, achieve a
statistically optimal rate? When is an unstable algorithm to be preferred to a stable counterpart?

Such questions about the trade-off between stability, computational
efficiency and the statistical error upon convergence are of special
interest for singular problems in which the Fisher information
matrices are degenerate. Singular problems appear in a wide range of
statistical settings, including mode estimation~\citep{Chernoff-1964},
robust regression~\citep{Rousseeuw-1984}, stochastic utility
models~\citep{Manski-1975}, informative non-response in missing
data~\citep{Heckman_1976, Diggle_1994}, high-dimensional linear
regression~\citep{Hastie-Tibshirani-Wainwright}, and over-specified
mixture models~\citep{Chen1992, Rousseau-2011, Nguyen-13}.  Several
papers have shown that maximum likelihood estimates for singular
problems have much lower accuracy than the classical parametric rate
$\obs^{-1/2}$; problems that exhibit slow rates of this type include
stochastic frontier models~\citep{Lee_1986,Lee_1993}, certain classes
of parametric models~\citep{Rotnitzky_2000}, and in strongly or weakly
identifiable mixture models~\citep{Chen1992, Nguyen-13,
  Ho-Nguyen-AOS-17}.  Nevertheless, the computational aspects of
parameter estimation and the trade-offs with stability in such models
are not well understood at the current time.

\subsection{Our contributions}
This paper lays out a general framework to address the questions
raised above. Making use of the population-to-sample approach
and a generalization of the localization argument from our previous
works~\citep{Raaz_Ho_Koulik_2018,Raaz_Ho_Koulik_2018_second}, we derive
tight bounds on the statistical error of the final iterate produced by
an algorithm. The final error and the number of steps taken depend on
two things: (i) the rate of convergence of the corresponding
population-level iterates, and (ii) the (in)stability of the
sample-level iterates with respect to that at the population-level.  As a first
contribution, our statistical guarantees for slowly converging stable
algorithms and (fast/slow converging) unstable algorithms complement
the findings of~\cite{Siva_2017} for fast
converging stable algorithms (Theorems~\ref{theorem:slow_meta_theorem} and
\ref{theorem:fast_unstab_meta_theorem}). We provide an overview of these
general results in Table~\ref{tab:summary_rates}. 

The second contribution extends the work of~\cite{chen2018stability} by showing how the final
statistical errors achieved by stable and unstable algorithms can be
used to directly compare and contrast the (dis)advantages between the
two (Section~\ref{sec:specific_models}). Our third contribution is an
explicit demonstration of the fact that unstable methods can converge
in significantly fewer steps when compared to stable methods, while
still yielding statistically optimal estimates
(Corollaries~\ref{cor:grad_infor_response},
\ref{corollary:Newton_mixture} and \ref{cor:grad_single_index}).  In
particular, applying our framework to three estimation
problems---single index models with known link, informative
non-response models, and Gaussian mixture models---we show that while
the (unstable) Newton method converges after on the order of $\log
\obs$ steps, there is some $q > 0$ such that gradient descent---which
we show to be a stable method---takes on the order of $\obs^{q}$
steps.  Finally, we also establish that our guarantees are
unimprovable in general on both statistical accuracy, and the
iteration complexity. 


\paragraph{Organization:} The remainder of our paper is
organized as follows. We begin in
Section~\ref{sec:motivation_and_problem_set_up} with simulations that
illustrate the phenomena to be investigated in this paper.  We then
introduce some definitions and discuss different properties of the
sample and population operators.
Section~\ref{sec:general_convergence} is devoted to statements of our
general computational and statistical guarantees with detailed proofs
presented in Appendix~\ref{sec:Proof}. In
Section~\ref{sec:specific_models}, we apply our general results to
demonstrate the trade-off between stable and unstable methods for
several examples.  We conclude with a discussion of potential future
work in Section~\ref{sec:discussion}.  Proofs of supporting lemmas and
technical results are provided in the appendices.


 \paragraph{Notation:} A few remarks on notation: for a pair
of sequences $\{a_n\}_{n \geq 1}$ and $\{b_n\}_{n \geq 1}$, we write
$a_{n} \succsim b_{n}$ or $a_n = \Omega(b_n)$ to mean that there is a universal constant $c$
such that $a_{n} \geq c b_{n}$ for all $n \geq 1$.  We write $a_{n}
\asymp b_{n}$ if both $a_{n} \succsim b_{n}$ and $a_{n} \precsim
b_{n}$ hold. We use $\ceil{x}$ to denote the smallest integer greater
than or equal to $x$ for any $x \in \Rspace$. In the paper, we use $c,
c', c_{i}, c_{i}'$ when $i \geq 1$ to denote the universal
constants. Note that the values of universal constants may change from
line to line. Finally, for our operator notation, we use the subscript
$n$ to distinguish a sample-based operator (e.g., $\nop, \samnewem,
\samresga$) from its corresponding population-based analog
(respectively $\pop, \newem, \resga$).

\begin{table}[t]

  \resizebox{1. \textwidth}{!}
  {

    {\renewcommand{\arraystretch}{2}

    \begin{tabular}{ccccc}
        \toprule
        {\bf Operator Properties} & \Centerstack{\bf Optimization \\
        \bf Rate} &
        {\bf Stability} 
        & \Centerstack{\bf  Iterations for \\ \bf convergence } & 
        \Centerstack{\bf  Statistical error \\  \bf on convergence}   \\

          \midrule
          \bf{General expressions} \\
        Fast, stable
        \citep{Siva_2017}
        & $\fast$ & $\stabilitynotag\texttt{(0)}$
        & $\log(1/{\noise(\obs, \tol)})$

        &  $\noise(\obs, \tol)$  

        \\[2mm]
       Slow, stable (Thm. \ref{theorem:slow_meta_theorem}) 
       & $\slow$ & $\stability$
       & $\specialeps^{-\frac{1}{1 + \parbeta -
    \pargamma\parbeta}}$

        & $[\specialeps]^{\frac{\parbeta}{1 + \parbeta -\pargamma
            \parbeta}}$

        \\[2mm]

        Fast, unstable (Thm. \ref{theorem:fast_unstab_meta_theorem})
        & $\fast$ & $\instability$ & $\log(1/{\noise(\obs, \tol)})$
        & $[\noise(\obs, \tol)]^{\frac{1}{1+\abss{\pargamma}}}$

        \\[2mm]

        Slow, unstable (Thm. \ref{theorem:fast_unstab_meta_theorem})
        & $\slow$ & $\instability$ 
        & $[\noise(\obs, \tol)]^{- \frac{1} {1 + \parbeta}}$
        & $[\noise(\obs, \tol)]^{\frac{\parbeta}{1+\parbeta+\abss{\pargamma}\parbeta}}$

        \\[2ex] \midrule 
        \Centerstack{\bf{Specific examples}\\ $\noise(\obs, \tol)= \log(1/\tol)/\sqrt{\obs}$} 
        \\

        \fast, \stabilitynotag\texttt{(0)}
        & $\ds e^{-\fastcont \iter}$ &  $\ds \frac{1}{\sqrt{\obs}}$
        & $\log \obs $ & $n^{-1/2}$

        \\[2mm]

        \slownotag\texttt{($\frac12$)}, \stabilitynotag\texttt{(1)} 
        & $\ds \frac{1}{\sqrt \iter}$ &  $\ds \frac{\radius}{\sqrt{\obs}}$
        & $n^{1/2}$ & $n^{-1/4}$

        \\[2mm]

        \fast, \instabilitynotag\texttt{(-1)}

        & $\ds e^{-\fastcont \iter}$ &  $\ds \frac{1}{\radius\sqrt{\obs}}$
        & $\log \obs $ & $n^{-1/4}$
    
        \\[2mm]

        \slownotag\texttt{($\frac12$)}, \instabilitynotag\texttt{(-1)}

        & $\ds \frac{1}{\sqrt \iter}$ &  $\ds \frac{1}{\radius\sqrt{\obs}}$
        & $n^{1/3} $ & $n^{-1/8}$

        \\[2ex] \bottomrule \hline
    \end{tabular}
    }
    }
    \noindent\caption{\textrm{A high-level overview of our results.
        The notation in the problem set-up (columns 2 and 3) is
        formalized in Section~\ref{sub:problem_set_up}, and the formal
        results (columns 4 and 5) are discussed in
        Section~\ref{sec:general_convergence}.  In the top panel, we
        provide general expressions from our results, and in the
        bottom panel, we provide some explicit expressions for few
        specific settings.  The second and third columns respectively
        denote the optimization and stability properties of the
        operator, and the last two columns provide the expressions for
        iterations for convergence, and the final statistical errors
        of the estimate returned the sample-based (noisy) operator
        (see equations~\eqref{eq:sample_size} for the definition of
        $\delta^*$).  For the bottom panel, we use $\parbeta=\frac12,
        \pargamma=0, -1$ with the noise function $\noise(\obs, \tol)=
        \log(1/\tol)/\sqrt{\obs}$. For brevity, we omit log-factors
        (in $\obs, \tol$) and universal constants for the expressions
        in the bottom panel.}}
    \label{tab:summary_rates}
    \vspace{-5mm}
\end{table}


\section{Motivation and problem set-up} 
\label{sec:motivation_and_problem_set_up}

We begin in Section~\ref{ssub:single_index_brief} by motivating the
analysis to follow by showing and discussing the results of some
computational studies for the class of non-linear regression models.  These
results demonstrate a wide range of possible convergence rates, and
associated stability (or instability) of the operator to
perturbations.  With this intuition in hand, we then turn to
Section~\ref{sub:problem_set_up}, in which we set up the definitions
that underlie our analysis.  In particular, we state the (i) local
Lipschitz condition, and (ii) local convergence behavior for the
population-level operator $\pop$, and (iii) the stability and
instability condition of the sample-level operator $\nop$ with respect
to $\pop$.



\subsection{A vignette on non-linear regression}
\label{ssub:single_index_brief}

We first consider a certain class of statistical estimation problems
in which there are interesting differences between algorithms.  Here
we keep the discussion very brief; see
Section~\ref{subsec:single_index} for a more detailed discussion.  We
consider a simple type of non-linear regression model, one based on a
function $f: \real^\usedim \rightarrow \real$ that can be written in
the form $f(x) = g \left( \inprod{x}{\theta} \right)$ for some
parameter vector $\theta \in \real^\usedim$, and some univariate
function $g: \real \rightarrow \real$.  In the simplest setting
considered here, the univariate function $g$ is known, and we have a
parametric family of functions as $\theta$ ranges over
$\real^\usedim$; when $g$ is unknown, we have a semi-parametric
family.  Now suppose that we are given a collection of pairs $\{(X_i,
Y_i) \}_{i=1}^\numobs$, generated from a noisy regression model of the
form
\begin{align}
\label{EqnSingleIndex}  
  Y_i & = g \left( \inprod{X_i}{\tstar} \right) + \newnoise_i, \qquad
  \mbox{for $i = 1, \ldots, \obs$.}
\end{align}
Here $\newnoise_i$ is a zero-mean noise variable with variance
$\sigma^2$, which we assume to be independent of $X_i$.  The single
index regression model~\eqref{EqnSingleIndex} has been studied
extensively in the literature (e.g.,~\citep{Carroll-1997,Ich93}).

When $g$ is known, a natural procedure for estimating $\theta$ is
based on minimizing the least squares objective function
\begin{align}
\label{EqnSingleIndexEmpirical}
\likelihood_\obs(\theta) & \defn \frac{1}{\numobs} \sum_{i=1}^\numobs
\left \{ Y_i - g \left( \inprod{X_i}{\theta} \right) \right \}^2.
\end{align}
When the variables $\newnoise_i$ are Gaussian, then this objective
coincides (up to scaling and constant factors) with the negative
log-likelihood function, so that minimizing it yields the maximum
likelihood estimate.

Under suitable regularity conditions on $g$ in a neighborhood of
$\tstar$, it is known that it is possible to estimate $\tstar$ at the
usual parametric rate of $n^{-1/2}$.  However, problems can arise when
the signal-to-noise ratio (SNR), as measured by the ratio
$\|\tstar\|_2/\sigma$, tends to zero.  In particular, consider a
function $g$ whose derivative vanishes at zero---that is, $g'(0) = 0$.
For instance, the function $g(t) = t^2$, which arises in the
application of the non-linear regression framework to the problem of
phase retrieval, has this property.  Taking the limit of low SNR
amounts to trying to estimate the vector $\tstar = 0$ based on
observations from the model~\eqref{EqnSingleIndex}.  For this type of
singular statistical model, we see many interesting differences
between algorithms that might be used to minimize the least-squares
criterion~\eqref{EqnSingleIndexEmpirical}.

\begin{figure}[h!]
\begin{adjustbox}{width=0.95\textwidth,center=\textwidth} 
  \begin{tabular}{cc}
    \widgraph{0.5\textwidth, trim={0.2cm, 0, 0, 0},
      clip}{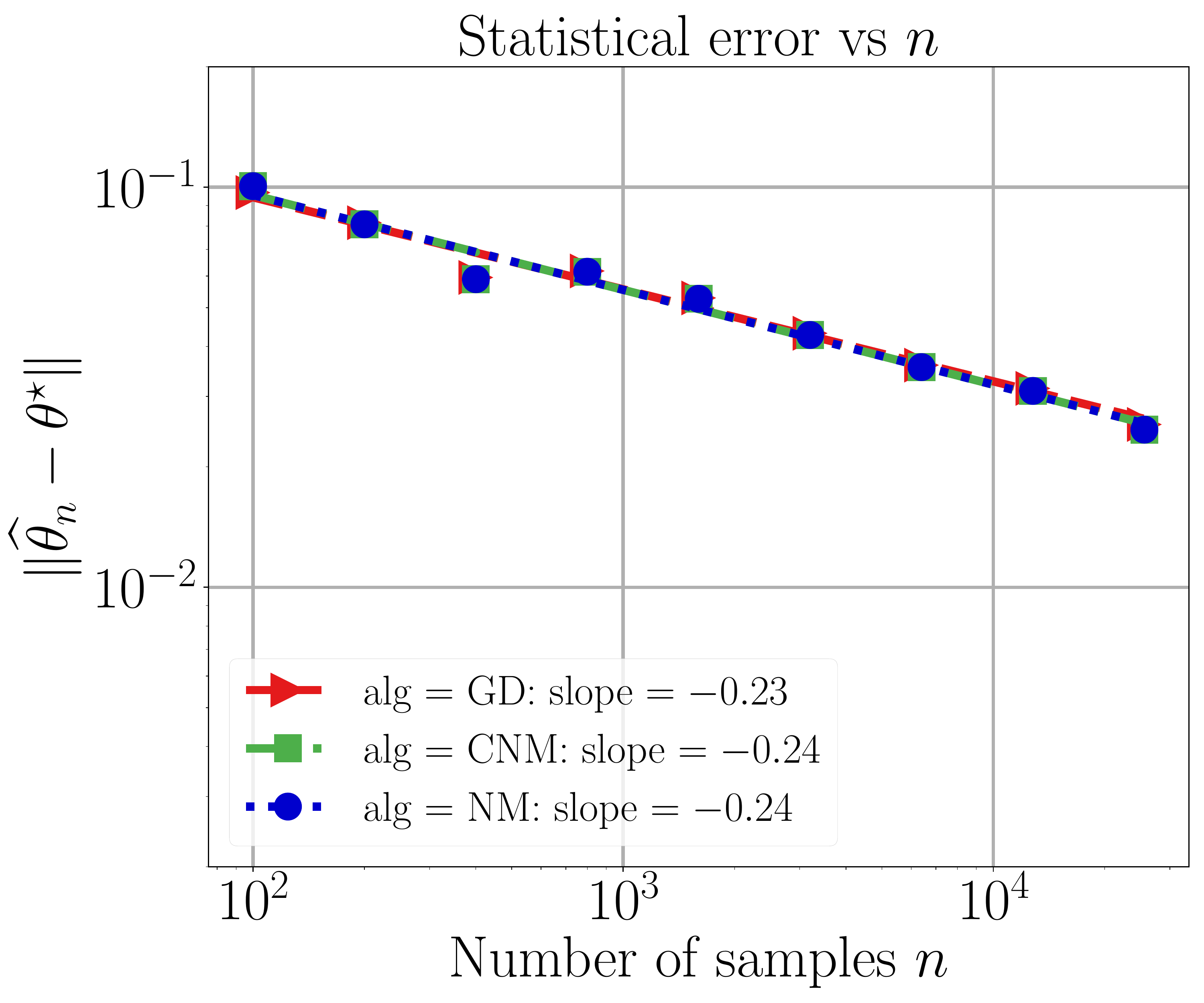} & \widgraph{0.5\textwidth,
      trim={0.2cm, 0, 0, 0}, clip}{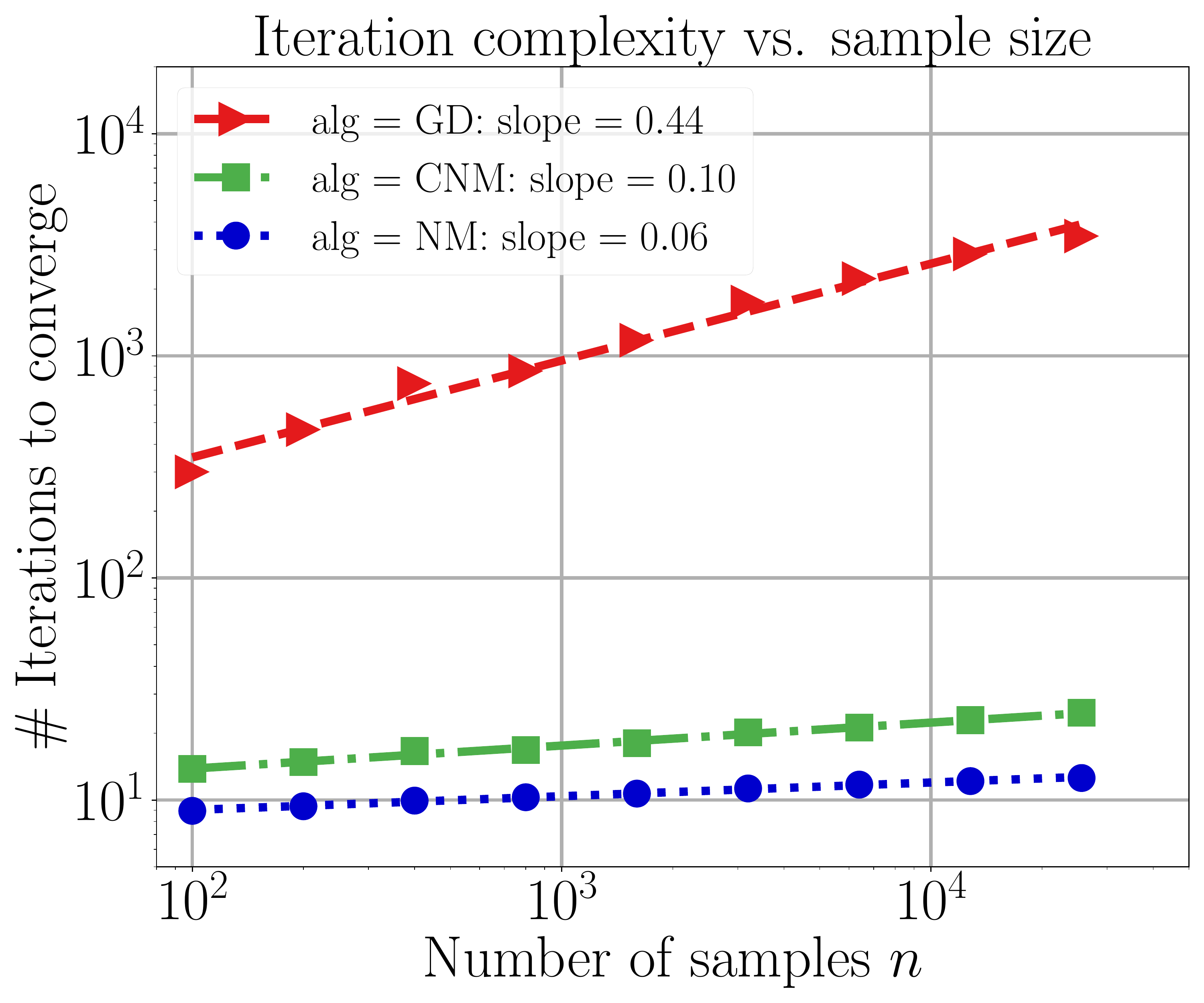} \\ (a)
    & (b)
  \end{tabular}
  \end{adjustbox}
  \caption{Plots characterizing the behavior of different algorithms,
    namely gradient descent (GD), cubic-regularized Newton's method
    (CNM), and the vanilla Newton's method (NM) for the non-linear
    regression model when $\tstar = 0$.  (a) Log-log plots of the
    Euclidean distance $\|\widehat{\theta}_\obs - \tstar \|_2$ versus
    the sample size.  It shows that all the algorithms converge to an
    estimate at Euclidean distance of the order $n^{-1/4}$ from the
    true parameter $\tstar$.  (b) Log-log plots for the number of
    iterations taken by different algorithms to converge to the final
    estimate. Newton's method takes the least number of steps. On the
    other hand, gradient descent takes significantly larger number of
    steps, with an empirical scaling close to $\sqrt{n}$.  }
  \label{FigSingleIndexEmpirical}
\end{figure}

More concretely, let us consider three standard optimization
algorithms that might be applied to the
objective~\eqref{EqnSingleIndexEmpirical}: (i) gradient descent; (ii)
Newton's method, and; (iii) cubic-regularized Newton's method.  See
Appendix~\ref{subsec:proof:corollary:grad_single_index} for a precise
description of these algorithms and the associated updates in
application to this model.

\paragraph{Statistical and iteration complexity of optimization algorithms:}
For each procedure, we are interested both in the associated
statistical error---that is, the Euclidean distance between their output and the true parameter $\tstar$---and their iteration
complexity, meaning the number of iterations required to converge.  In
order to gain some understanding, we performed some simulations for
non-linear regression based on the function $g(t) = t^2$ in
dimension $d = 1$, over a range of sample sizes $\obs$.
Figure~\ref{FigSingleIndexEmpirical} provides some plots that
summarize some results from these simulations.  Panel (a) plots the
Euclidean error associated with the estimate versus the sample size
$\obs$ on a log-log plot, along with associated least-squares fits to
these data.  As can be seen, all three methods lie upon a line with
slope $-1/4$ on the log-log scale, showing that the statistical error
decays at the rate $\obs^{-1/4}$.  This ``slow rate''---to be
contrasted with the usual $\obs^{-1/2}$ parametric rate---is a
consequence of the singularity in the model.  Panel (b) plots the
iteration complexity of the three algorithms versus the sample sizes, again on a log-log plot.  For a given problem based on $\obs$ samples,
the iteration complexity is the number of iterations required for the
distance between the iterate and $\tstar$ to drop below $\obs^{-1/4}$.
Here we see some interesting differences, with the gradient method
having an empirical iteration complexity that grows as $\approx
\obs^{0.44}$, based on our fits, with the two forms of Newton's method
having much milder growth in iteration complexity.  In the theory to
follow, we will prove that iteration complexity for the gradient
method scales at most like $\sqrt{\obs}$, that of the
cubic-regularized Newton method scales as $\obs^{1/6}$, whereas that
of Newton's method scales only as $\log \obs$. (See
Corollary~\ref{cor:grad_single_index} for a precise statement.)

\begin{figure}[h!]
  \begin{tabular}{cc}
    \widgraph{0.45\textwidth}{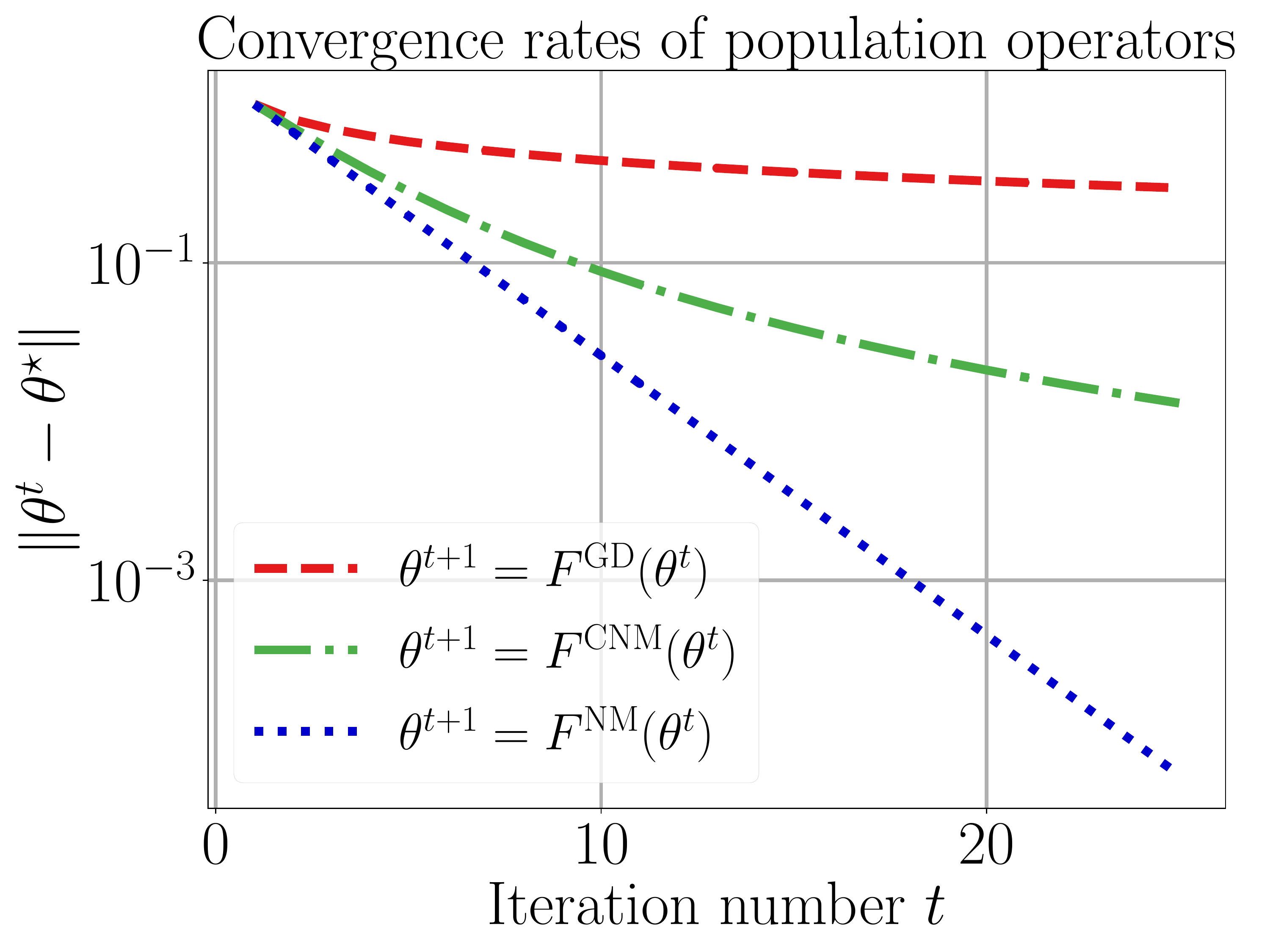}
    & 
    \widgraph{0.45\textwidth}{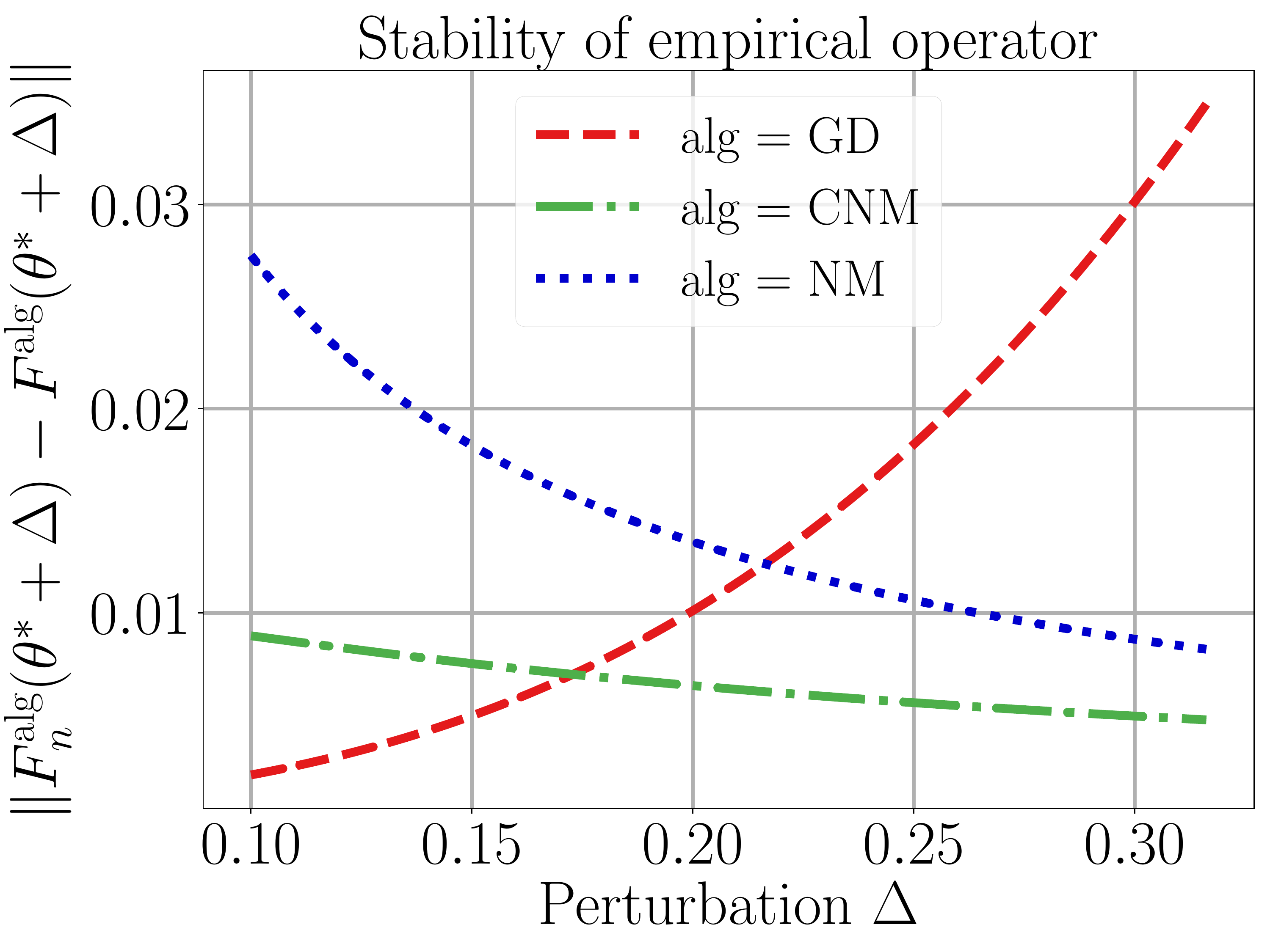}
    \\
    (a) & (b)
  \end{tabular}
  \caption{Exploration of the population level updates, and their
    connection to the empirical updates for the non-linear regression problem.
    (a) Plots showing the convergence rate of the error
    $\enorm{\tvar^t-\tstar}$ for different algorithms---namely
    gradient descent (GD), standard Newton's method (NM), and
    cubic-regularized Newton's method (CNM)---applied at the
    population level (limit of infinite sample size).  Notice the
    log-scale on the $y$-axis.  The sequence from the Newton's method
    converges a geometric rate to $\tstar$, whereas the gradient
    method converges at a sub-linear rate.  (b) Plots showing the
    scaling of the perturbation error $\enorm{ \nop(\tstar + \Delta
      )-\pop(\tstar + \Delta)}$ versus the perturbation $\Delta$.  For
    an unstable operator, the perturbation error can increase as
    $\enorm{\Delta} \rightarrow 0$, with Newton's method showing a
    strong version of such instability.  In contrast, the gradient
    descent method is a stable procedure in this setting.}
  \label{FigSingleIndexPopulation}
\end{figure}

\paragraph{Behavior of optimization operators:}
The plots in Figure~\ref{FigSingleIndexEmpirical} all concern the behavior of algorithms in practice, as applied to the \empirical
objective function, and our ultimate goal is to provide a theoretical
explanation of phenomena of these types.  In order to do so, our analysis makes use of the population-level algorithms obtained in the
limit of infinite sample size; i.e., $\obs \to \infty$.  In the special case of the non-linear regression model considered here, we refer the
Appendix~\ref{subsec:proof:corollary:grad_single_index} for the precise forms of these
operators (cf.
equations~\eqref{eq:pop_gd_single_idx}--\eqref{eq:pop_cnm_single_idx}).
The plots in Figure~\ref{FigSingleIndexPopulation} illustrate the two
properties of the operators that underlie our theoretical analysis:
convergence rate of the population operators (panel (a)), and the
stability of the \empirical operators relative to the population
version (panel (b)).

The plots in panel (a) reveal that the three algorithms differ
dramatically in their convergence rate at the population level.  The
ordinary Newton updates converge at a geometric rate, with the
distance to the optimum $\tstar$ decreasing as $\kappa^t$ with the
number of iterations $t$, where $\kappa \in (0,1)$ is a contraction
coefficient.  In contrast, the other two algorithms exhibit an inverse
polynomial rate of convergence, with the distance to optimality
decreasing at the rate $1/t^\betapar$ for some exponent $\betapar >
0$.  In the analysis to follow, we prove that gradient descent has
inverse polynomial decay with exponent $\betapar = 1/2$, whereas the
cubic-regularized Newton updates exhibit inverse polynomial decay with
exponent $\betapar = 2$.

In Corollary~\ref{cor:grad_single_index} and its proof, we characterize
the optimization rate (algorithmic rate of convergence), the stability
and the final statistical error obtained by these three methods. 
For reader's convenience, we summarize these results in Table~\ref{tab:summary_explicit}.

\begin{table}[h!]
    \centering
  {

    {
    \renewcommand{\arraystretch}{2}

    \begin{tabular}{ccccc}

        \toprule

       {\bf Algorithm} & \shortstack{\bf Optimization 
        Rate} &
        {\bf Stability} 
        & \shortstack{\bf  Iterations for \\ \bf convergence } & 
        \shortstack{\bf  Statistical error \\  \bf on convergence}   
        \\
        \midrule

        Gradient descent
        & $\ds \frac{1}{\sqrt \iter}$ &  $\ds \frac{\radius}{\sqrt{\obs}}$
        & $n^{1/2}$ & $n^{-1/4}$

        \\[2mm]

        Newton's method

        & $\ds e^{-\fastcont \iter}$ &  $\ds \frac{1}{\radius\sqrt{\obs}}$
        & $\log \obs $ & $n^{-1/4}$
    
        \\[2mm]

        \shortstack{Cubic-regularized \\ Newton's method}

        & $\ds \frac{1}{\iter^2}$ &  $\ds \frac{1}{\sqrt{\radius}\sqrt{\obs}}$
        & $n^{1/6} $ & $n^{-1/4}$

        \\[2ex] \bottomrule \hline

    \end{tabular}
    }}
    \caption{Overview of results illustrated in  Figures~\ref{FigSingleIndexEmpirical}
    and \ref{FigSingleIndexPopulation} for non-linear regression model with
    the link function $g(t) = t^2$ and $\tstar=0$.
    By characterizing the optimization rate and stability precisely, and
    invoking our general theory (summarized in Table~\ref{tab:summary_rates}), we
    establish that while the three methods differ significantly
    in terms of their optimization rate and stability, they achieve the
    same statistical error upon convergence, albeit by taking different
    number of iterations to converge. We omit logarithmic factors and universal
    constants for brevity. See Corollary~\ref{cor:grad_single_index}
    and its proof for precise details.
    }
    \label{tab:summary_explicit}
\end{table}


\subsection{Problem set-up} 
\label{sub:problem_set_up}

Having provided a high-level overview of the phenomena that motivate
our analysis, let us now set up the problem more abstractly, and
introduce some key definitions.  Consider an operator $\pop$ that maps
a space $\Theta$ to itself; typical examples of the space $\Theta$
that we consider are subsets of the Euclidean space $\real^d$, and
subsets of symmetric matrices. Let $\tstar$ be a fixed point of the
operator---i.e., an element $\tstar \in \Theta$ such that
$\pop(\tstar) = \tstar$.  The challenge is that we do not have access
to the operator $\pop$ directly, but rather can observe only a random
operator $\nop$ that can be understood as a noisy estimate of $\pop$.\footnote{For ease of exposition, going forward the index $n$ is synonymous with the sample size that defines the operator $\nop$; while our general results, namely, Theorems~\ref{theorem:slow_meta_theorem} and \ref{theorem:fast_unstab_meta_theorem} do not rely on use of this simplification.}
Throughout, we call $\pop$ the \emph{population operator} and $\nop$
the \emph{\empirical operator}. Using the empirical operator, we
generate a sequence of iterates via the fixed-point updates
\begin{align}
\tvar_n^{\iter+1} & = \nop (\tvar_n^{\iter}) \qquad \mbox{for $\iter =
  1, 2, \ldots$,}
\end{align}
with a suitable initialization $\tvar_n^0 \in \Theta$.  Our goal is to
determine conditions under which the sequence $\{ \tvar_n^t \}_{t \geq
  0}$ approaches a suitably defined neighborhood of $\tstar$. More
precisely, for any given triple $(\pop, \nop,t)$ we provide a sharp
characterization of the optimality gap $\|\tvar_n^{\iter} - \tstar
\|_2$ as a function of the iteration count $\iter$ and the error $\|
\pop - \nop \|_2$ of the \empirical operator $\nop$.

One interesting class of problems where the operators $\pop$ and
$\nop$ arise naturally is estimation problems in statistics and
machine learning.  More concretely, consider the problem of finding
the unique minimizer $\tstar$ of an objective function $\likelihood:
\Theta \rightarrow \real$.  In practice, we do not know the true
objective function $\likelihood$, instead we have access to an
approximate (random) objective function $\likelihood_\numobs$, which
is an unbiased estimate of the true objective function $\likelihood$.
Given the pair $(\likelihood, \likelihood_\numobs)$, we can obtain
different operators $\pop$ by applying various optimization algorithms
to minimize $\likelihood$, including gradient methods, proximal
methods, the EM algorithm and related majorization-minimization
algorithms, as well as Newton and other higher-order methods.  The
noisy operators $\nop$ are obtained by applying the same optimization
algorithms to the approximate objective function
$\likelihood_\numobs$.


\subsubsection{Properties of the operator $\pop$} 
\label{ssub:properties_for_the_operator}

We begin by formalizing some properties of the operator $\pop$. We
assume that the operator $\pop$ has a unique fixed point $\tstar$ and
we study its behavior in the local neighborhood of the Euclidean
ball
\begin{align}
  \ballstar & : = \Big \{ \theta \in \Theta \mid \|\theta - \tstar\|_2
  \leq \lradius \Big \}
\end{align}
centered at $\tstar$.  Our first condition is a standard Lipschitz
condition on the operator $\pop$.  In particular, we say that the
operator $\pop$ is \emph{$1$-Lipschitz} in $\| \cdot \|$ norm
over the ball $\ballstar$ if
\begin{align}
  \label{eq:lipschitz}
  \enorm{\pop(\tvar_1)-\pop(\tvar_2)} \leq 
  \enorm{\tvar_1-\tvar_2} \quad \text{for all }\tvar_1, \tvar_2 \in
  \ballstar.
\end{align}
In words, the $1$-Lipschitz condition guarantees that the operator $\pop$
is non-expansive with respect to perturbations of its argument.

Our next two definitions distinguish between fast and slow rates of
convergence.  The first definition captures an especially favorable
property of operator $\pop$; namely, it is locally contractive around
the fixed point $\tstar$. The second definition considers a
substantially slower (sub-linear) rate of convergence of the operator
$\pop$. 

\begin{definition}[Fast convergence]
\label{def:fast_convergence}
For a contraction coefficient $\kappa \in (0, 1)$, the operator $\pop$
is \mbox{$\fast$-convergent} on the ball $\ballstar$ if
\begin{align}
  \label{eq:fast_convergence}
  \enorm{\pop^{\iter}(\tvar_0) - \tstar} \leq \kappa^\iter \,
  \enorm{\tvar_0 - \tstar} \quad \mbox{for all iterations $\iter = 1,
    2, \ldots$,}
\end{align}
and for all $\tvar_0 \in \ballstar$.
\end{definition}

\begin{definition}[Slow convergence]
\label{def:slow_convergence}
Given an exponent $\parbeta > 0$, the operator $\pop$ is
$\slow$-convergent over the ball $\ballstar$ means that
\begin{align}
  \label{eq:slow_convergence}
  \enorm{\pop^{\iter}(\tvar_0) - \tstar} \leq \frac{\unicon}{\iter^
    \parbeta} \quad \mbox{for all iterations $\iter = 1, 2, \ldots$,}
\end{align}
and for all $\tvar_0 \in \ballstar$, where $\unicon$ is a universal
constant.
\end{definition}

These notions of fast and slow convergence are ubiquitous in analysis of iterative methods, especially in the optimization literature. For example,  when the operator $\pop$ corresponds to gradient descent for some objective $\likelihood$, a sufficient condition for fast convergence is local strong convexity of the objective $\likelihood$, and if $\likelihood$ is just convex, $\pop$ satisfies slow convergence. 
Let us now illustrate these definitions with a simple example.
\begin{example}[Fast versus slow convergence]
  \label{ExaPopulation}
  Consider the function $\likelihood(\theta) = \frac{\theta^{2p}}{2p}$
  for some positive integer $p \geq 1$. Note that for any $p \geq 1$,
  the function $\likelihood(\cdot)$ has a unique global minimum at
  $\tstar = 0$. The first two derivatives of $\likelihood(\cdot)$ are
  given by
  \begin{align*}
    \likelihood'(\theta) = \theta^{2p-1}, \quad \mbox{and} \quad
    \likelihood''(\theta) = (2p-1) \theta^{2p-2}.
  \end{align*}
Consequently, a gradient descent update with a constant stepsize
$\stepsize > 0$ takes the form
\begin{align}
\label{eq:simple_gd}
  \Fgrad(\theta) &= \theta - \stepsize \likelihood'(\theta) \; = \;
  \theta \big( 1 - \stepsize \theta^{2p-2} \big).
\end{align}
Thus, when $p = 1$, for any $\stepsize \in (0,1)$, this gradient descent
update is a $\fast$-convergent algorithm with $\kappa = 1 - \stepsize$.
On the other hand, for any $p \geq 2$, it can be shown that
gradient descent is $\slow$-convergent with parameter 
$\betapar =  \frac{1}{2p-2}$ in the ball
$\ballstar$ with $\thetastar=0$ and $\lradius= h^{-\frac{1}{2p-2}}$.

Now, let us consider Newton's method with step size one,
namely the update
\begin{align}
\label{eq:simple_newton}
  \Fnewt(\theta) & = 
  \theta - \big( \likelihood''(\theta) \big)^{-1}\likelihood'(\theta) \;
  = \; \theta - \frac{\theta^{2p-1}}{(2p-1) \theta^{2p-2}} \; 
  = \; \theta
\left(1 - \frac{1}{2p-1} \right).
\end{align}
For $p = 1$, this update converges in a single step (simply because
the quadratic approximation that underlies Newton's method is exact in
this special case).  For $p \geq 2$, the pure Newton update is
\mbox{$\fast$-convergent} with $\kappa = 1 - \frac{1}{2p-1}$ for all 
$\theta \in \real$.
\end{example}


\subsubsection{From the empirical operator $\nop$ to the population operator $\pop$} 
\label{ssub:properties_of_the_operator_}
In this section, we introduce some key concepts that characterize the (in)-stability of the sample operator $\nop$ with
respect to the population operator $\pop$. Given a pair of operators
$(\nop, \pop)$ and a tolerance parameter $\delta \in (0,1)$,
our definitions involve a \emph{perturbation function} $\noise(\cdot)$
that maps the triple $(\nop, \pop, \delta)$  to a positive (deterministic) scalar  $\noise(\obs, \tol)$. In general, we impose the following conditions on the perturbation
function $\noise(\cdot)$:
\begin{itemize}
\item It is decreasing in $\obs$ for any fixed $\tol$, and is
  monotonically increasing in $\tol$ for any fixed
  $\obs$.
\item For any fixed $\tol \in (0, 1)$, we have $\noise(\obs, \tol) \to
  0$ as $\obs \to \infty$, and similarly, for any fixed $\obs > 0$, we
  have $\noise(\obs, \tol) \to \infty$ as $\tol \to 0$.
\end{itemize}
Note that $\noise(n, \delta) = \sqrt{\log(1/\delta)/n}$ would satisfy these requirements.
Given some choices of perturbation
function, we can define our first stability condition as follows:
 
\begin{definition}[$\stability$-Stability]
For a given parameter $\pargamma \geq 0$, the operator $\nop$ is
$\stability$-stable over $\ballstar$ with noise function $\noise(\cdot)$
means that, for any radius $\radius \in (0, \lradius)$ and tolerance
$\tol \in (0, 1)$, we have
\begin{align}
  \label{eq:sample_stability}
  \Prob \Big [ \sup \limits_{\tvar \in \ball(\tstar, \radius)}
    \enorm{ \nop(\tvar) - \pop( \tvar)} \leq \unicontwo \min \Big \{
    \radius^\pargamma \noise(\obs, \tol), \radius \Big \} \Big]
  \geq 1-\tol,
\end{align}
for some positive universal constant $\unicontwo$. 
\end{definition}
 Informally,
the stability condition~\eqref{eq:sample_stability} guarantees that
with high probability, the error $\enorm{ \nop(\tvar) - \pop( \tvar)}$
is upper bounded by $\unicontwo \min \{ \radius^\pargamma \noise(\obs,
\tol), \radius \}$ uniformly over a disk of radius $\radius$.  Note
moreover that the upper bound decays to $0$ as the radius $\radius
\rightarrow 0^+$. 

Next we consider the case when $\pargamma<0$, i.e., the
perturbation error $\enorm{\nop(\tvar)- \pop( \tvar)}$ blows up as
$\tvar$ gets close to $\tstar$. We refer to such operators as unstable operators.  Given radii $ \radius_1, \radius_2$
such that $\radius_2 > \radius_1\geq 0$, let \mbox{$\annulus(\tstar,
  \radius_1, \radius_2) = \ball(\tstar, \radius_2)\backslash
  \ball(\tstar, \radius_1)$} denote the annulus around $\tstar$ with
inner and outer radii $\radius_1$ and $\radius_2$ respectively.

\begin{definition}[$\instability$-Instability]
For
a given parameter $\pargamma < 0$ and radii $0 < \inradius <
\outradius$, we say that the operator~$\nop$ is $\instability$-unstable
over the annulus $\annulus(\tstar, \inradius, \outradius)$ with noise
function $\noise(\cdot)$ if
\begin{align}
  \label{eq:sample_instability}
  \Prob\brackets{\sup \limits_{\tvar \in \annulus(\tstar, \radius,
      \outradius)} \enorm{\nop(\tvar) - \pop( \tvar)} \leq
    \noise(\obs, \tol) \max \braces{
      \frac{1}{\radius^{\abss{\pargamma}}}, \outradius}} \geq 1 -
  \tol,
\end{align}
for any radius $\radius \in [\inradius, \outradius]$ and any
tolerance $\tol \in (0, 1)$. 
\end{definition}

Two remarks are in order: First, note that the main difference between \stability and \instability is how the error scales with the radius $r$ as it gets smaller. For stable operators, the error decreases with scaling $r^{\pargamma}$, while for unstable operators the error blows up as $r^{-\abss{\pargamma}}$ (where we use $\abss{\pargamma}$ for clarity). There is another subtle difference: the condition~\eqref{eq:sample_instability} defines the instability of the perturbation error $\enorm{\nop(\tvar)- \pop( \tvar)}$ in an annulus with the inner radius bounded below by $\inradius$, and does not characterize the behavior as the distance $\enorm{\tvar - \tstar}\to 0$. Let us now illustrate these definitions by following up on
Example~\ref{ExaPopulation}.

\begin{example}[Stable versus unstable updates]
  \label{ExaEmpirical}
  Consider an empirical function of the form
  \begin{align}
  \label{eq:simple_sample_problem}
\likelihood_\obs(\theta) & = \frac{1}{2p} \theta^{2p} + \frac{\sigma
  w}{2\sqrt{\obs}} \theta^2, \qquad \mbox{where $w \sim N(0,1)$}.
  \end{align}
  Here $p \geq 2$ is a positive integer. Note that $\Exs[\likelihood_\obs(\theta)] = \frac{1}{2p} \theta^{2p}$,
  which is equivalent to the population likelihood function
  considered in Example~\ref{ExaPopulation}.

A gradient update with stepsize $\stepsize > 0$ on the empirical
objective leads to the empirical gradient operator
\begin{align*}
  \Fgrad_\obs(\theta) & = \theta \left \{ 1 - \stepsize \theta^{2p-2} -
  \stepsize \frac{\sigma w}{\sqrt{\obs}} \right \}.
\end{align*}
Comparing with equation~\eqref{eq:simple_gd}, we obtain that 
$\left| \Fgrad_\obs (\theta) - \Fgrad (\theta) \right|\;
= \frac{\sigma}{\sqrt{\obs}} |w| \; |\theta|$.  
Since $|w| \leq 4 \sqrt{\log(1/\tol)}$ with probability at least $1 - \tol$,
we see for any $\lradius > 0$ and $\obs \geq 16 \sigma^2 \log(1/\delta)$,
the operator $\Fgrad_\obs$ is \mbox{$\stability$-stable} with parameter
$\pargamma = 1$, with respect to the noise function
\begin{align*}
\noise(\obs, \tol) & = 4 \sigma \sqrt{\frac{\log(1/\tol)}{\obs}}.
\end{align*}

As for the Newton update for the problem~\eqref{eq:simple_sample_problem}, 
we have 
\begin{align*}
\Fnewt_\obs(\theta) & 
= \; \theta - \frac{\theta^{2p-1}+{\sigma w\theta}/{\sqrt{\obs}}}
{
(2p-1) \theta^{2p-2}+{\sigma w}/{\sqrt{\obs}}},
\end{align*}
and hence
\begin{align*}
  \abss{\Fnewt(\theta)-\Fnewt_\obs(\theta)} 
  = \frac{(2p-2)}{(2p-1)} \cdot \frac{\sigma \abss{w}\abss{\theta}/\sqrt n}{(2p-1)\theta^
  {2p-2} + \sigma w/ \sqrt n}.
\end{align*}
Recall that $|w| \leq 4 \sqrt{\log(1/\tol)}$ with probability at least $1
- \tol$.
Plugging in $w > -4 \sqrt{\log(1/\tol)}$ in the denominator and $w < 4\sqrt{\log
1/\tol}$ of the RHS, and doing some algebra yields that
\begin{align*}
  \abss{\Fnewt(\theta)-\Fnewt_\obs(\theta)} 
  \leq 
  \frac{c_p}{\abss{\theta}} \sqrt{\frac{\log (1/\tol)}{n}}
  \quad
  \text{for}
  \abss{\theta} > \bigg(
  c_p' \sigma
  \sqrt{\frac{\log(1/\tol)}
    { n}}\bigg)^{\frac{1}{2p-2}},
\end{align*}
where $c_p = \frac{16(p-1)}{2p-1}$ and $c_p'=\frac{8}{2p-1}$ .
Thus, we conclude that the operator $\Fnewt_\obs$ is $\instability$-unstable
with parameter $\pargamma=-1$
over the annulus $\annulus(\tstar, \inradius, \outradius)$ with noise function $\noise$ where
\begin{align*}
  \inradius =\bigg(c_p'\sigma \sqrt{\frac{\log(1/\tol)}
    { n}}\bigg)^{\frac{1}{2p-2}}, 
    \quad \outradius = \infty, \quad\text{and}\quad
  \noise(n, \tol) = c_p\sigma\sqrt{\frac{\log (1/\tol)}{n}}.
\end{align*} 
\end{example}

\subsubsection{Comparison of our assumptions with empirical process literature}
 We note that while the definition of $\stability$ is reminiscent of the typical assumptions in empirical process literature, there is a subtle difference in our set-up. In a typical statistical learning problem, the following assumptions are commonly made on: (a) the local curvature of the population objective function (e.g., the expected negative log-likelihood $\likelihood$), and (b) bounds on the perturbation error between the population and sample objective functions (e.g., $\sup_{\tvar\in\ball(\tstar, r)}|\likelihood(\tvar)-\likelihood_{\obs}(\tvar)|$). With these assumptions, the statistical guarantees for the critical points (e.g., the maximum likelihood estimate (MLE)) are then established. See, e.g., Theorem 3.2.5~\citep{vanderVaart-98}.

Such a framework is oblivious about any computational aspect of the problem (e.g., how the MLE is computed), which is one of the key focus in our work. Our goal is to study the interplay of computational-statistical tradeoffs between various algorithms that are used to solve these learning problems. In particular, our aim is to identify the number of iterations taken by an algorithm, and the final statistical accuracy of the estimate returned by it. Consequently, our conditions are defined in terms of operators that correspond to the algorithm employed by the user to solve the problem at hand, rather than the landscape of the objective itself. In particular, in place of the curvature condition~(a) on $\likelihood$, we make assumptions on the convergence rate of the population operator $\pop$ (\slow/\fast). And, in place of the perturbation bounds on the objective functions ($\likelihood, \likelihood_n$), we make assumptions on the operator perturbation errors between $\pop$ and $\nop$ as in equations~\eqref{eq:sample_stability} and \eqref{eq:sample_instability} (\stability/\instability). 

\subsubsection{Further discussion on our definitions}
In several cases (also applicable to all examples in this paper), the user often knows (by design) the explicit relationship between the operators $\pop$ and $\nop$ and the corresponding objectives $\likelihood$ and $\likelihood_\obs$, e.g., when $\pop$ and $\nop$ correspond to gradient ascent (GA) or Newton's method (NM). In these situations, often it is possible to derive whether $\stability$ or $\instability$ conditions are satisfied given the assumptions on the curvature of $\likelihood$ and the perturbation error between $\likelihood$ and $\likelihood_{\obs}$ as in the empirical process literature. Our framework allows the user to simultaneously study the tradeoffs between the final statistical error and the computational budget needed between several algorithms at once. For example, we show in several settings that NM while being unstable provides computational benefits compared to its stable counterpart GA, since both NM and GA yield an estimate with comparable statistical error upon convergence while the former takes very few steps (although such a condition is not guaranteed always).

We remark that the property $\instability$ of the operators $\pop$ and $\nop$ as introduced above has not been commonly used in prior work, while $\stability$ has appeared often in prior works (albeit in slightly different forms, \citep{Siva_2017,chen2018stability,Raaz_Ho_Koulik_2018}).
The condition $\stabilitynotag(0)$ is perhaps the most common, which holds for most well-conditioned problems (when the Fisher information matrix is invertible). In such settings, the commonly used methods like GA and NM are also $\fast$ operators so that the final statistical error is of order $\noise(n, \delta)$ which is obtained in roughly $\log(1/\noise(n, \delta))$ steps~\citep{Siva_2017}. In simple words, the statistical-computational tradeoffs across several algorithms are fairly similar for such cases.

On the other hand, operators with $\slow$, and $\stability$ with $\gamma \geq 1$, would typically arise when the log-likelihood is not well-conditioned and one uses methods like GA, and $\instability$ with $\gamma<0$ would appear in such a setting when one uses a higher-order optimization scheme like NM to solve these ill-conditioned problems.  So far, there is a limited understanding of the statistical-computational tradeoff for slowly converging stable algorithms as well as any unstable algorithm that can arise in such settings. Our main results provide a comprehensive understanding towards this end.

Let us revisit Examples~1 and 2 which serve as motivating examples for the various ill conditioned settings: Suppose that the population negative log-likelihood is given by $\likelihood(\theta)= \theta^{2p}/(2p)$ (and the true parameter is $\tstar=0$), and the sample negative log-likelihood is given by $\likelihood_n(\theta)= \theta^{2p}/(2p)+\sigma w\theta^2/(2\sqrt n)$. For this setting, the population Fisher information (the second derivative of $\likelihood$) is given by $(2p-1)\theta^{2p-2}$, and the perturbation term $\sigma w\theta^2/(2\sqrt n)$ between the two objectives mimics the intuition that the finite sample Fisher information would typically have $1/\sqrt n$ fluctuations around its population-level objective with $n$ samples. With $p=1$, it is easy to establish that the operators $\pop$ and $\nop$ corresponding to both GD and NM are $\fast$ and $\stabilitynotag(0)$ operators. However, for $p\geq 2$, as our earlier computations illustrated, we observe a more interesting set of behaviors with GD and NM. In particular, the operators corresponding to GA are $\slow$ and $\stability$ with $\gamma>1$, and the NM operators exhibit a $\fast$ and $\instability$ with $\gamma<0$ behavior. The theory to follow provides a precise characterization of the statistical error achievable and the computational budged needed for convergence in such settings.

\section{General convergence results}
\label{sec:general_convergence}

With the definitions from the previous section in place, we are now
ready to state our main results.  In Section~\ref{sub:main_results},
we consider the case when $\nop$ is a stable perturbation of $\pop$,
and in
Section~\ref{sub:statistical_rates_for_unstable_but_fast_converging_operators},
we consider the case when it is an unstable perturbation of $\pop$. We summarize our findings in Table~\ref{tab:summary_rates}.


\subsection{Results for slowly converging but stable operators} 
\label{sub:main_results}

We first consider the setting in which the sample-based operator
$\nop$ is a stable perturbation of the population-level operator
$\pop$.  If, in addition, we assume that the operator $\pop$ has fast 
convergence (cf.\  equation~\eqref{eq:fast_convergence}), then past work
is applicable. In particular, Theorem 2 of~\cite{Siva_2017} provides a precise
characterization of the convergence behavior of iterates from the empirical operator $\nop$.  Here we instead
consider the more challenging setting in which the operator $\pop$
exhibits slow convergence to $\tstar$.  Analysis of this slow
convergence case requires rather different techniques than those used
to analyze the fast-convergent case.

Let us collect the assumptions needed to state our first result.  The
first two assumptions involve the Euclidean ball $\ballstar$ centered
at $\tstar$ of some fixed radius $\lradius > 0$.
\begin{enumerate}[label=(\Alph*)]
\item\label{item:local_lipschitz} The population operator $\pop$ is
  $1$-Lipschitz~\eqref{eq:lipschitz} and is
  \slow-convergent~\eqref{eq:slow_convergence} over the ball
  $\ballstar$.
\item\label{item:sample_stable} There is some \mbox{$\pargamma \in [0, \parbeta^{-1}]$} such
    that the empirical operator $\nop$ is
    \stability-stable~\eqref{eq:sample_stability} over $\ballstar$.

\item\label{item:sample_size} The tolerance parameters
  $\tol \in (0, 1)$ and $\smallthreshold \in (0,
  \frac{\parbeta}{1+\parbeta-\pargamma\parbeta})$ are fixed
  and the sample size is large enough such that
  \begin{align}
    \label{eq:sample_size}
    \specialeps \leq c \quad \text{where}
    \quad \tol^* := \tol
    \cdot\frac{\log(\frac{1+\parbeta}{\parbeta\pargamma})}
              {8\log(\frac{\parbeta}{\smallthreshold(1 + \parbeta
                  -\pargamma \parbeta}))},
  \end{align}
  and $c \in (0,1)$ is a sufficiently small constant.
\end{enumerate}
Assumptions~\ref{item:local_lipschitz} and \ref{item:sample_stable}
quantify, respectively, the convergence behavior of the operator
$\pop$ and the stability of the operator $\nop$;
Assumption~\ref{item:sample_size} is a book-keeping device needed to
state our results cleanly.
Given the above conditions, we now state our first main result.
\begin{theorem}
\label{theorem:slow_meta_theorem}
Under Assumptions~\ref{item:local_lipschitz},
\ref{item:sample_stable}, and \ref{item:sample_size}, consider the
sequence \mbox{$\tvar_n^{\iter+1}= \nop (\tvar_n^{\iter})$} generated
from an initialization $\tvar_n^0\in\ball(\tstar, \lradius/2)$.  Then
there is a universal constant $\unicon'$ such that for any fixed
$\smallthreshold \in (0, \frac{\parbeta}{1+\parbeta-\pargamma\parbeta})$ and uniformly for all iterations $\iter
\geq \unicon' \big(1/\specialeps)^{\frac{1}{1 + \parbeta -
    \pargamma\parbeta}} \log\frac{1}{\smallthreshold}$, we have
\begin{align}
  \label{eqn:slowsta_meta_result}
  \enorm{\tvar^\iter_n - \tstar} & \leq 2
        [\specialeps]^{\frac{\parbeta}{1 + \parbeta -\pargamma
            \parbeta} - \smallthreshold} \qquad \mbox{with probability
          at least $1-\tol$.}
\end{align}
\end{theorem}
\noindent Let us make some comments on this result 
(see Appendix~\ref{proof:theorem:slow_meta_theorem} for a detailed
proof). \\

\paragraph{Tightness of Theorem~\ref{theorem:slow_meta_theorem}:}
Disregarding the term $\smallthreshold$ and constants, the
bound~\eqref{eqn:slowsta_meta_result} guarantees that the sequence
\mbox{$\tvar_n^{\iter+1}= \nop (\tvar_n^{\iter})$} converges to a
statistical tolerance of order $[\specialeps]^{\frac{\parbeta}{1 +
    \parbeta -\pargamma \parbeta}} $ with respect to $\tstar$ in order $[\specialeps]^{-\frac{1}{1 +
    \parbeta -\pargamma \parbeta}}$ step.  This
guarantee turns out to be unimprovable under the given assumptions.
In Appendix~\ref{sec:lower_bounds} (see Proposition~\ref{prop:lower_poly}), we construct a family of examples with the operators $\pop$,
$\nop$ and noise functions $\noise_n$ (constant with respect to $\delta$) satisfying the assumptions for Theorem~\ref{theorem:slow_meta_theorem}, such that the following additional results hold:
\begin{align*}
  \enorm{\tvar_\obs^\iter-\tstar} &
      \begin{cases}
      \geq \noise_n^{\frac{\parbeta}{1 +
    \parbeta -\pargamma \parbeta}} \ \ \quad\text{for all} \quad t \geq 1,\\
      \geq 2\noise_n^{\frac{\parbeta}{1 +
    \parbeta -\pargamma \parbeta}}
      \quad\text{for all} \quad t\leq c'\noise_n^{-\frac{1}{1 +
    \parbeta -\pargamma \parbeta}}.
      \end{cases}
\end{align*}
As a result, we conclude that the results of Theorem~\ref{theorem:slow_meta_theorem} are tight for both statistical accuracy and the number of iterations needed for convergence.

\paragraph{Relation to prior work:} As noted earlier, prior work~\citep{Siva_2017} shows that when the operator is $\pop$ is $\fast$-convergent, and $\nop$ is a $\stabilitynotag(0)$ perturbation of $\nop$ with error $\noise(n, \delta)$, we have $\Vert\tvarn^T-\tstar\Vert\precsim\noise(n, \delta)$ for $T \succsim\log(1/\noise(n, \delta))$. On the other hand,~\cite{chen2018stability} argued that given the minimax error of a problem class, a fast converging algorithm can not be too stable. Neither of these works provided a precise characterization of what statistical errors are achievable when dealing with a slow converging stable operator, which is the focus of our Theorem~\ref{theorem:slow_meta_theorem}.\footnote{In several statistical settings, the error function satisfies $\noise(n, \delta)=c\sqrt{\log(1/\delta)/n}$. When the problems are well-conditioned, e.g., if the Fisher information matrix is invertible while estimating MLE, we typically have $\fast$ and $\stabilitynotag(0)$ condition for commonly used algorithms like gradient ascent, and Newton's method. In general, we do not expect a setting where the operators are $\fast$ and $\stability$ with $\pargamma \geq 1$. Although, one can construct pathological examples, in which case, our localization argument augmented with the earlier proofs by~\cite{Siva_2017} would yield that $\tvarn^t\stackrel{t\to\infty}{\to} \tstar$, i.e., the statistical error converges to zero as the number of iterations goes to $\infty$ (even with finite samples).} 

\paragraph{A direct sub-optimal proof argument:} 
Let us first illustrate how a naive argument that tries to directly tradeoff the perturbation error of $\nop$ with the convergence rate of $\pop$ leads to a sub-optimal guarantee. Let the assumptions in Theorem~\ref{theorem:slow_meta_theorem} remain in force. Roughly speaking, one can show that (cf. Lemma~\ref{lemma:t_perturbation_error}), the operator $\nop^t$ is also $\stability$ perturbation of $\pop^t$ with the noise function $t \cdot  \noise(n, \delta)$, so that we can bound the error at iteration $t$ as follows:
\begin{align}
  \Vert \tvarn^{t}\!-\!\tstar \Vert
  \!=\!  \Vert \nop^t(\tvarn^{0})\!-\!\tstar \Vert
    \!\leq\! \Vert \nop^t(\tvarn^{0})\! -\!\pop^t(\tvarn^{0}) \Vert
    \!+\! \Vert \pop^t(\tvarn^{0}) \!-\!\tstar \Vert
    \leq tC_{\rho} \cdot \noise(n, \!\delta)\! +\! \frac{1}{t^{\beta}},  \label{eq:sub_optimal}
\end{align}
where $C_{\rho} = \rho^\pargamma$ denotes a constant corresponding to the radius $\rho$ of initialization. Minimizing the last bound in the display above over the iteration index $t$, we find that the best possible error is of order $(\noise(n, \delta))^{\parbeta/(1+\parbeta)}$. This rate is clearly sub-optimal when compared to the statistical error of order $(\noise(n, \delta))^{\parbeta/(1+\parbeta-\pargamma\parbeta)}$ (unless $\pargamma=0$) guaranteed by display~\eqref{eqn:slowsta_meta_result} from Theorem~\ref{theorem:slow_meta_theorem}. The reason for sub-optimality of this bound is our failure to localize the argument with the perturbation error as the iterates $\tvarn^t$ converge closer to $\tstar$. \\

\paragraph{Outline of proof:} 
In order to derive the sharp guarantee, we need to establish a more refined tradeoff than that in equation~\eqref{eq:sub_optimal}. To this end, we generalize and refine the annulus-based localization argument introduced in our prior work on the EM algorithm~\citep{Raaz_Ho_Koulik_2018,
  Raaz_Ho_Koulik_2018_second}. In the past work~\citep{Raaz_Ho_Koulik_2018, Raaz_Ho_Koulik_2018_second}, we studied particular instantiations of the EM algorithm, for which the operators $\pop$ and $\nop$ had closed-form solutions. Here in the absence of closed-form expressions, the argument is necessarily more abstract to establish a sharp guarantee under the more general Assumptions~\ref{item:local_lipschitz}, \ref{item:sample_stable}, and \ref{item:sample_size}, which also handles the previous analysis as a special case (as illustrated in Section~\ref{sub:over_specified_gaussian_mixture_models}).

\begin{figure}[t]
\begin{center}
  \begin{tabular}{c}
    \widgraph{0.9\textwidth}{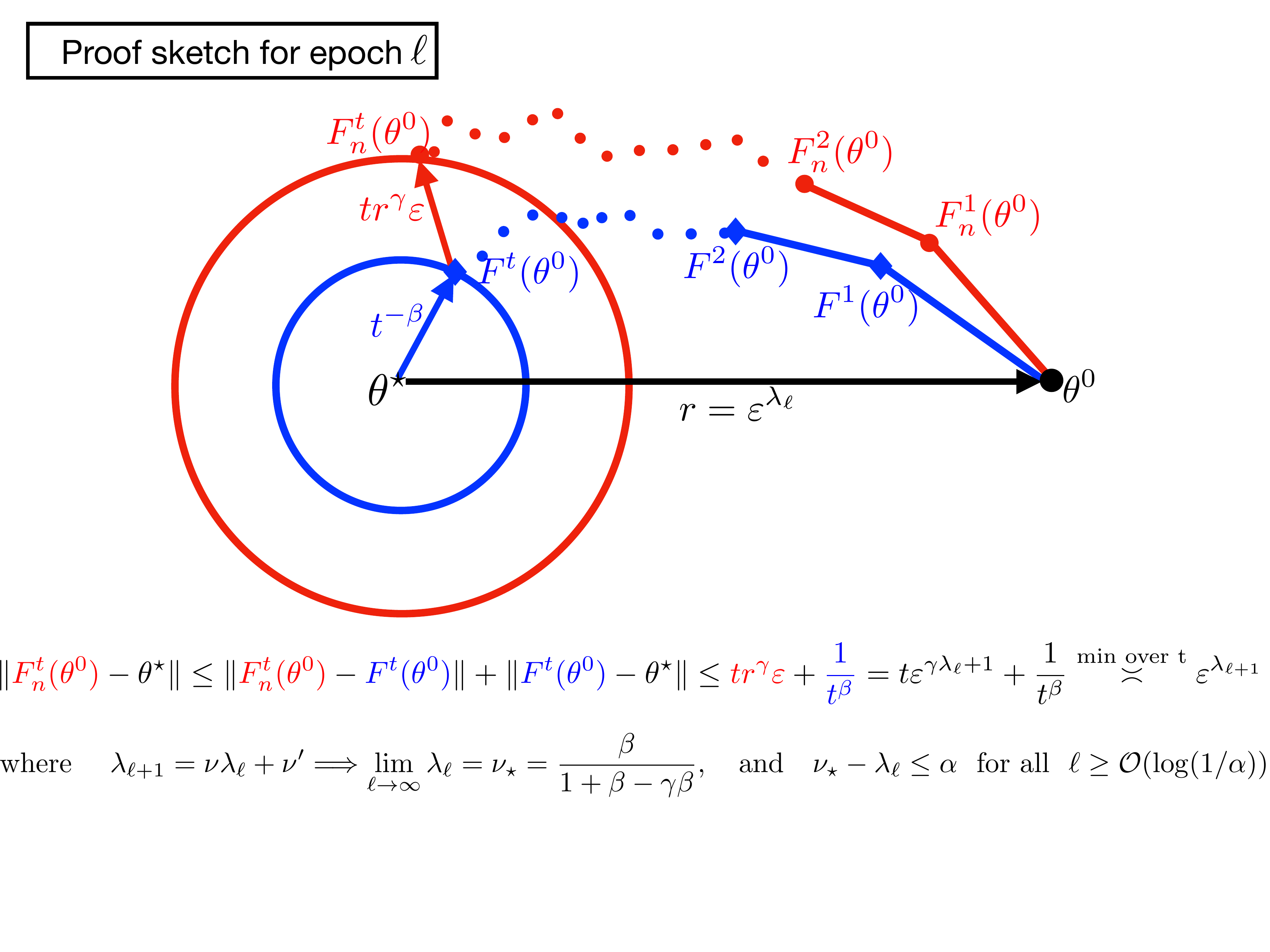}
  \end{tabular}
\end{center}
\vspace{-15mm}
  \caption{An illustration of the epoch-based argument when the population
 operator $\pop$ is $\slow$-convergent, and the noisy operator is $\stability$-stable
  (Theorem~\ref{theorem:slow_meta_theorem}).
  In order to simplify the visualization, we use the shorthand $\noise = \noise(\obs,
  \tol^*)$. Moreover, here $\theta^0$ denotes the starting
  point for a given epoch $\ell$ (assumed to be at distance $r=\noise^{
  \seqalpha{\ind}}$ from $\tstar$), and the iterations $1, 2, \ldots, t$ denote
  the iteration count in that epoch. 
  The population iterates $F^{1}(\theta^0), F^{2}(\theta^0), \ldots$ converge
  towards to $\tstar$ at the rate $t^
  {-\beta}$ (shown in blue), and their distance from the noisy iterates
  $F_n^{1}(\theta^0),
  F_n^{2}(\theta^0), \ldots$ grows at the rate at a distance of $t r^{\gamma}
  \noise$. Trading-off the two errors, we can show that at the end of epoch $\ell$
  (denoted by a suitable choice of $t$), the distance $\enorm{\nop^{t}(\theta^0)-\tstar}
  \precsim \noise^{\seqalpha{\ind+1}}$. By establishing that $\seqalpha{\ind}$
  converges to $\nu_\star$ exponentially fast, and that similar arguments
  can be made for sufficiently many epochs,  we obtain the result
  in Theorem~\ref{theorem:slow_meta_theorem}.
  See Appendix~\ref{proof:theorem:slow_meta_theorem} for a formal argument.
  }
  \label{Fig:annulus_argument}
\end{figure}

At a high-level, the proof proceeds by decomposing the total
collection of iterations $\{1, 2, \ldots, \iter\}$ into a disjoint
partition of subsets $\{ \numsteps_{\ind }\}_{\ind \geq 0}$, referred
to as epochs, where the nonnegative integers
$\ind$ and $\numsteps_{\ind}$ respectively denote the index of a given
epoch and the number of iterations in that epoch. We use
$\totalnumsteps_{\ind} \defn \sum_{i = 0}^{\ind} \numsteps_{i}$ to
denote the total number of iterations up to epoch $\ind$.  By
carefully choosing the sequence $\{\numsteps_{\ind}\}_{\ind \geq 0}$,
we ensure that at the end of a given epoch $\ind$, the error
$\enorm{\tvarn^{\totalnumsteps_{\ind}} - \tstar}$ has decreased to a
prescribed threshold.  More precisely, using an inductive argument on $\ell$, we
show that
\begin{align}
\label{eq:induction_step}
    \|\tvarn^{\totalnumsteps_{\ind}} - \tstar\| \leq
    \mydndelta^{\seqalpha{\ind}} \quad\text{for all epoch }\quad \ind
    \geq 1,
  \end{align}
  where the sequence $\{\seqalpha{\ind}\}_{\ind \geq 0}$ is defined via
  the recursion
  \begin{align}
  \label{eq:recursion}
    \seqalpha{0} = 0 \quad \text{and} \quad \seqalpha{\ind + 1} =
    \parnuone \seqalpha{\ind} + \parnutwo, \quad \text{for all} \quad
    \ind \geq 1,
  \end{align}
  with the scalars $\parnuone \in (0, 1)$ and $\parnutwo > 0$
  determined by the problem parameters $\parbeta$ and $\pargamma$.  We
  show that the sequence $\{\seqalpha{\ind}\}_{\ind \geq 0}$ converges
  to $\parfinal_\star\mydefn \frac{\parbeta}{1 + \parbeta - \pargamma
    \parbeta}$ fast enough and we have $|\seqalpha{\ind} -
  \parfinal_\star| \leq \smallthreshold$ for all $\ind \geq
  \order{\log (1/\smallthreshold)}$.  Deriving a suitable upper bound
  on $\numsteps_{\max}$ on the epoch size $\numsteps_i$, we then put
  the pieces together to (roughly) conclude that
  \begin{align*}
      \|\tvarn^{\iter} - \tstar\| \leq c\mydndelta^{\parfinal_\star-\smallthreshold}
      \quad\text{for}\quad
      \iter \geq c' \numsteps_{\max} \cdot \log \frac{1}{\smallthreshold}.
  \end{align*}
  As expected, much of the technical work is required to establish the inductive step.  The full proof of the theorem is given
  in Appendix~\ref{proof:theorem:slow_meta_theorem}.
  We also illustrate the high-level ideas of the epoch-based
  localization argument in Figure~\ref{Fig:annulus_argument}.

\subsection{Results for unstable operators}
\label{sub:statistical_rates_for_unstable_but_fast_converging_operators}

We now turn to our next main result which characterizes the
convergence when the operator $\nop$ is an unstable perturbation of
the operator $\pop$.  We consider two distinct cases depending on
whether the operator $\pop$ is (a) $\fast$-convergent or (b)
$\slow$-convergent.
\begin{theorem} 
\label{theorem:fast_unstab_meta_theorem}
For a given parameter $\tol \in (0, 1)$, consider the sequence
\mbox{$\tvar_n^{\iter+1}=\nop (\tvar_n^{\iter})$} for some initial
point $\tvar_n^0$ in the ball~$\ball(\tstar, \lradius/2)$.  Suppose
that for some $\pargamma < 0$, the empirical operator $\nop$ is
$\instabilitynotag(\pargamma)$-unstable over the annulus $\Ann(\tstar,
\innerradius_\obs, \lradius)$ with respect to the noise function
$\noise$.
\begin{enumerate}[label=(\alph*)]
\item\label{item:fast} Suppose that the operator $\pop$ is
  $\fast$-convergent over the ball $\ballstar$, and the sample size
  $\obs$ is sufficiently large so as to ensure that
    \begin{subequations}
    \begin{align}
    \label{eq:fast_unstable_thm_condition}
          [\noise(\obs, \tol)]^{\frac{1}{1+\abss{\pargamma}}} \leq
                (1-\fastcont) \lradius.
    \end{align}
Then with probability at least $1-\tol$, for any iteration
$\iter \geq \frac{\log(\frac{\lradius}{\noise(\obs, \tol)})}
{(1+\abss{\pargamma})\log\frac{1}{\fastcont}}$, we have
\begin{align}
  \label{eq:fast_unstable_thm_statement}
  \min_{\counter \in \braces{0, 1, \ldots,
      \iter}}\enorm{\tvar_\obs^\counter-\tstar} \leq \max \left \{
  \small{\frac{(2-\fastcont)}{(1-\fastcont)}}\cdot [\noise(\obs,
    \tol)]^{\frac{1}{1+\abss{\pargamma}}}, \; \innerradius_\obs \right \}.
\end{align}
\end{subequations}
\item\label{item:slow} Suppose that the operator $\pop$ is
  $1$-Lipschitz and \mbox{$\slow$-convergent} for some
  $\parbeta>0$, and that the sample size $\obs$ is large enough to
  ensure that
    \begin{subequations}
    \begin{align}
    \label{eq:slow_unstable_thm_condition}
                     [\noise(\obs,
                       \tol)]^{\frac{\parbeta}{1+\parbeta-\pargamma\parbeta}}
                     \leq \lradius.
    \end{align}
Then with probability at least $1-\tol$, for any iteration $\iter \geq
{\frac{1}{[\noise(\obs, \tol)]^{\frac{1} {1 + \parbeta}}}}$, we have
    \begin{align}
    \label{eq:slow_unstable_thm_statement}
    \min_{\counter \in \braces{0, 1, \ldots, \iter}}
        \enorm{\tvar_\obs^\counter-\tstar} \leq \max \left \{
              [\noise(\obs,
                \tol)]^{\frac{\parbeta}{1+\parbeta-\pargamma\parbeta}},
              \innerradius_\obs \right \}.
    \end{align}
    \end{subequations}
\end{enumerate}
\end{theorem}  
\noindent Let us make a few comments about these bounds. (See Appendix~\ref{proof:theorem:fast_unstab_meta_theorem} for a detailed
proof.)\\ 

 \paragraph{Choice of the inner radius $\innerradius_\obs$:}
In order to obtain sharp upper bounds---ones that depend purely on the
noise function $\noise$---the inner radius $\innerradius_\obs$ must be
chosen suitably. Focusing on part (a), if we ensure that
$\innerradius_\obs \leq [\noise(\obs, \tol)]^{\frac{1}{1+\pargamma}}$,
then we obtain an upper bound on the error that involves only the
noise function. We show how to make such choices in our
applications of this general theorem.  A similar statement applies to
part (b) of the theorem.

\paragraph{Tightness of Theorem~\ref{theorem:fast_unstab_meta_theorem}:}
In Appendix~\ref{sec:lower_bounds}, we construct examples of the
operators $\pop$ and $\nop$ which satisfy the assumptions of
Theorem~\ref{theorem:fast_unstab_meta_theorem}, and with the inner
radius satisfying the bound $\innerradius_\obs \leq [\noise(\obs,
  \tol)]^{\tau}$, $\tau = \frac{1}{1 + \pargamma}$ for part (a) or
$\tau = \frac{\parbeta}{1+\parbeta-\pargamma\parbeta}$ for part (b).
For each of these examples, we show that the sequence
\mbox{$\tvar_n^{\iter+1}= \nop (\tvar_n^{\iter})$} satisfies the lower
bound
\begin{align*}
        \enorm{\tvar_\obs^\iter-\tstar} \geq [\noise(\obs, \tol)]^{\tau}
        \quad
        \text{for all }\quad\iter \geq 0,
\end{align*}
with constant probability.  Thus, we conclude that the results of
Theorem~\ref{theorem:fast_unstab_meta_theorem} are tight and not
improvable in general. 

 \paragraph{Necessity of the minimum:}
Note that both of the bounds~\eqref{eq:fast_unstable_thm_statement}
and \eqref{eq:slow_unstable_thm_statement} apply to the minimum over
all iterates $k \in \{1, 2, \ldots, t \}$, as opposed to the final
iterate $t$.  For this reason, our results only guarantee that the
iterates produced by an unstable operator $\nop$ converge at least
once to a vicinity of the parameter $\tstar$, but \emph{not} that they
necessarily stay there for all the future iterations.  In fact, such
``escape'' behavior for an unstable algorithm is unavoidable in the
absence of any additional regularity assumptions.  In particular, we
provide a simple example in Appendix~\ref{AppCounterExample} that
illustrates this unavoidability. 

\paragraph{Additional regularity condition:}  If we impose an additional
regularity condition, then we can remove the minimum from the
guarantee. In particular, consider the condition:
  \begin{enumerate}[label=(\Alph*)]
    \setcounter{enumi}{3}
    \item\label{item:unstab_control} There exists a universal constant $C$
    such that for a given initialization $\tvar_\obs^0$,
    the sequence $\tvar_{n}^{t} = \nop^{t}( \tvar_{\obs}^{0})$ 
    has the following property:
    \begin{align}
    \label{eq:unstab_control}
    \enorm{ \tvar_{\obs}^{\iter + 1} - \tstar} &\leq C \innerradius
    \quad \text{whenever} \quad \enorm{\theta_{n}^{\iter} - \tstar} \leq
    \innerradius,
    \end{align}
    where the radius $\innerradius$ corresponds to 
    equation~\eqref{eq:fast_unstable_thm_statement} or~\eqref{eq:slow_unstable_thm_statement}
    as relevant to $\pop$.
  \end{enumerate}
Under this condition, it is straightforward to modify the proof of
Theorem~\ref{theorem:fast_unstab_meta_theorem} to show that the bounds
in both parts (a) and (b) can be sharpened by replacing the term
$\min_{\counter\in \braces{0, 1, \ldots,
    \iter}}\enorm{\tvar_n^\counter-\tstar}$ with
$\enorm{\tvar_n^\iter-\tstar}$.  In Section~\ref{sec:specific_models}
to follow, we provide a number of examples for which
Assumption~\ref{item:unstab_control} is satisfied.


\section{Some concrete results for specific models}
\label{sec:specific_models}

In this section, we study three interesting classes of statistical
problems that fall within the framework of the paper. We also discuss
various consequences of Theorems~\ref{theorem:slow_meta_theorem} and
Theorem~\ref{theorem:fast_unstab_meta_theorem} when applied to these
problems.


\subsection{Informative non-response model}
\label{subsec:infor_resp} 

In our first example, let us consider the problem of biased or
informative non-response in sample surveys. In certain settings, the
chance of a response to not be observed depends on the value of the
response.  This form of non-response introduces systematic biases in
the survey and associated conclusions~\citep{Heckman_1976}.  Some
examples where this issue arises include longitudinal
data~\citep{Diggle_1994}, housing surveys
and election polls~\citep{shaiko1991pre}.  In such settings, it is
common practice to estimate the non-responsive behavior in order to
correct for the bias. We now describe one simple formulation of such a
setting.
 
Suppose that we have $\obs$ i.i.d.\ values $Y_{1}, \ldots, Y_{n}$ for the
response variable $Y \sim \NORMAL(\mu, \sigma^2)$, where for each $Y_i$
there is a chance that the value is not observed.
To account for such a possibility, 
we define $\braces{0,1}$-valued random variables $R_i$ for $i=1, \ldots,
\obs$ as follows:
\begin{subequations}
\label{eq:informative_model}
\begin{align}
    R_i = 1 \quad\text{if $Y_i$ is observed},
    \qquad \text{and}\qquad R_i = 0 \quad\text{otherwise}.
\end{align}
We assume that the conditional distribution $R_i | Y_i$ takes
the form
\begin{align}
    \Prob_\tvar (R_{i} = 1| Y_{i} = y)    
    = \exp \parenth{H\big(\tvar{(y-\mu)}/{\sigma}\big)},
\end{align}
where $H$ is a known function and $\tvar$ is an unknown parameter which
controls the dependence of the probability of non-response on the observation
$Y=y$. In a general setting, all the parameters $\mu, \sigma$ and $\theta$
are unknown and are estimated jointly from the data. However, to simplify
our presentation, we assume that the parameters $(\mu, \sigma)$ are known
and only $\theta$ needs to estimated.
In particular, we consider the case when the response variable 
$Y \sim \Ncal(\mu, \sigma^2) \equiv \Ncal(0, 1)$ and $H(x) = -x^2-\log 2$.
Under these assumptions, simple algebra yields that
\begin{align}
    \Prob_\tvar (R_{i} = 1| Y_{i} = y) 
        = \exp \parenth{- \frac{\tvar^2 y^2}{2} - \log 2}
        \quad \text{and} \quad
        \Prob_\tvar (R_{i} = 1) = \frac{1}
    {2 \sqrt{\tvar^2 + 1}}.
\end{align}
\end{subequations}
Given $n$ i.i.d. samples $\{R_i, Y_i\}_{i = 1}^n$, where we note that $Y_i$
is not observed when $R_i=0$, the log-likelihood is given by
\begin{align}
\label{eq:sam_likeli_infor}
    \likelihoodres_{n}( \tvar) 
        \mydefn \frac{1}{n} \sum_{i = 1}^{n} - \frac{ R_{i} 
        \parenth{ Y_{i}^2 (\tvar^2 + 1) + 2 \log 2}}{2} 
        + (1 - R_{i}) \log \parenth{1 - \frac{1}{2 \sqrt{\tvar^2 + 1}}}.
\end{align}
Note that the likelihood above does not depend on the unobserved $Y_i$ 
since $R_i=0$ makes the contribution of the corresponding term $0$.

In the remainder of this section, we focus on the singular regime,
i.e., when the true parameter $\tstar = 0$ and consequently the
probability of observing any sample $Y_i=y$ is always $1/2$
(independent of the value $y$).  For such a setting, the results of~\cite{Rotnitzky_2000} imply that the statistical error
of the MLE is larger than the parametric rate $n^{-\frac12}$. In
particular, they showed that $\vert{\widehat{\tvar}_ {\obs,
    \mathrm{MLE}}-\tstar}\vert = \mathcal{O}({\obs^{-\frac{1}{4}}})$.
However, with high probability, the log-likelihood
$\likelihoodres_{n}$ is non-concave\footnote{For instance, when
  $\sum_{i = 1}^{n} R_{i} (Y_{i}^2 + 1) < n$, the sample
  log-likelihood function is bimodal and symmetric around 0.} and
thereby a closed-form for the maximum-likelihood estimate is not
available.  Thus a theoretical analysis of the estimates obtained via
different optimization algorithms (that can be used to maximize the
log-likelihood $\likelihoodres_{n}$) can be of significant interest.
We now apply our general theory to analyze two optimization methods:
(i) gradient ascent, and (ii) Newton's method.

\subsubsection{Theoretical guarantees}
We now state a theoretical guarantee on the behavior of the
optimization algorithms in practice with the informative non-response
model~\eqref{eq:informative_model}---that is, when applied
to the sample log likelihood~\eqref{eq:sam_likeli_infor}. 
We analyze the gradient ascent updates for a step-size $\learnrate
\in (0, \tfrac{8}{3})$, and the pure Newton updates. 
We use $\samresga$ and $\samresnm$ respectively to denote the sample-based
operators for gradient ascent and Newton's method   (see Appendix~\ref{sub:proof_of_corollary_ref_grad_info}
for the precise form of these operators). The following statement also involves
other universal constants $c, c_i, c_i', c_i''$ etc.
\begin{corollary} 
\label{cor:grad_infor_response}
For the singular setting of informative non-response model ($\tstar = 0$) and given some $\tol \in
(0,1)$, the following properties hold
with probability at least $1 - \tol$:
\begin{enumerate}[label=(\alph*)]
  \begin{subequations}
\item\label{item:cor_infor_gd} For any fixed $\smallthreshold \in (0, 1/4)$ and initialization $\tvar^0 \in \ballnotag (\tstar, 1/ 2)$,
the sequence $\tvar^\iter \defn (\samresga)^\iter(\tvar^0)$ of gradient
iterates satisfies the bound
\begin{align}
  \abss{\tvar^\iter - \tstar } & \leq c_{1} \parenth{\frac{\log(
  \frac{\log(1/\smallthreshold)}{\tol})}
    {\obs}}^{\frac{1}{4}-\smallthreshold}
  \qquad \mbox{for all iterates $\iter \geq c_{1}' \sqrt{\obs} \log\frac{1}
  {\smallthreshold}$,}
\end{align}
as long as $n\geq c_1'' \log\frac{\log(1/\smallthreshold)}{\tol}$.
\item\label{item:cor_infor_newton} 
For any initialization $\tvar^0 \in \annulus ( \tstar, \sqrt{2 \unicon}
\parenth{\log( 1/ \delta)/ n}^{1/ 4}, 1/ 2)$, the sequence of Newton iterates $\tvar^\iter \defn
  (\samresnm)^\iter(\tvar^0)$ satisfies the bound
  \begin{align}
  \abss{\tvar^\iter - \tstar } & \leq c_{2} \parenth{\frac{\log(1/\tol)}
  {\obs}}^{\frac{1}{4}}
  \qquad \mbox{for all iterates $\iter \geq c_{2}' \log \obs$,}
  \end{align}
  as long as $n\geq c_2'' \log(1/\tol)$.
  \end{subequations}
\end{enumerate}
\end{corollary}
\noindent See Appendix~\ref{sub:proof_of_corollary_ref_grad_info} for
the proof of this corollary (and below for the proof sketch).\\

Corollary~\ref{cor:grad_infor_response} shows that given $n$ samples,
 (i) the final statistical errors achieved by the iterates generated by the
gradient descent
and the Newton's method are similar (of order $n^{-\frac {1}{4}}$), 
and (ii) the Newton's method takes a considerably smaller number (of order $\log n$) of steps in comparison to
that taken by gradient ascent (of order $\sqrt n$). Finally, in
Appendix~\ref{sub:proof_of_corollary_ref_grad_info}, we show that all
the non-zero fixed points of the considered operators have a magnitude
of the order $n^{-\frac{1}{4}}$ with constant probability. Therefore, the statistical radius
achieved by the given optimization methods are optimal.

\subsubsection{Proof sketch for Corollary~\ref{cor:grad_infor_response}} 
\label{ssub:proof_sketch_for_cor_grad_infor_response}
Our proof of Corollary~\ref{cor:grad_infor_response} 
starts with an analysis of the gradient ascent and Newton
iterates on the population-level analog of the problem.
In particular, taking expectations in equation~\eqref{eq:sam_likeli_infor},
we obtain the following population-level optimization problem
\begin{align}
\label{eq:pop_likeli_infor_expli}
\max_{\tvar \in \Rspace} \likelihoodres( \tvar) \quad\text{where}\quad
\likelihoodres( \tvar) = \frac{1}{2} \log \parenth{1 - \frac{1}{2
    \sqrt{\tvar^2 + 1}}} - \frac{\tvar^2 + 1}{4}.
\end{align}
Let $\resga$ denote the gradient update operator applied to this
objective with a given step-size $\learnrate$, and let $\resnm$ denote
the Newton update.
In Appendix~\ref{sub:proof_of_corollary_ref_grad_info} (where we also provide
explicit forms of these operators), we show that
with $\tstar=0$,  the population-level operators have the following properties:
\begin{enumerate}[label=(P\arabic*)]
    \item\label{item:pop_gd} 
    The gradient operator $\resga$ is $\slow$-convergent with
    parameter $\parbeta=\frac12$ over the  Euclidean ball $\ball(\tstar,
    \frac{1} {2})$, i.e., for the sequence $\tvar^\iter = (\resga)^\iter(\tvar^0)$
    with $\tvar^0 \in \ball(\tstar, \frac{1} {2})$,
    we have $\abss{\tvar^\iter-\tstar} \leq \frac{c}{\iter^{1/2}}$.
    \item\label{item:pop_newton} The Newton operator $\resnm$ is $\fast$-convergent
    with parameter $\fastcont=\frac{4}{5}$ over the Euclidean ball $\ball(\tstar,
    \frac{1}{2})$, i.e., for the sequence $\tvar^\iter = (\resnm)^\iter(\tvar^0)$
    with $\tvar^0 \in \ball(\tstar, \frac{1} {2})$,
    we have $\abss{\tvar^\iter-\tstar} \leq {c}\; e^{-\fastcont \iter}$.
\end{enumerate}
Moreover in the same Appendix~\ref{sub:proof_of_corollary_ref_grad_info}, 
we show that with the noise function $\noise(\obs, \tol) = \sqrt{
\frac{\log(1/\tol)}{\obs}}$, the sample-level operators satisfy the following
properties:
\begin{enumerate}[label=(S\arabic*)]
  \item\label{item:sample_gd} The sample-based gradient ascent operator
  $\samresga$ is $\stability$-stable
  with parameter
  $\pargamma=1$  over the ball $\ball(\tstar, \frac{1}{2})$, and 
  \item\label{item:sample_newton} the operator $\samresnm$ is $\instability$-unstable
  with parameter $\pargamma=-1$
  over the annulus $\Ann(\tstar, \innerradius_\obs, \lradius)$
  with $\innerradius_\obs = c[\noise(\obs,\tol)]^{\frac{1}{2}}$ and $\lradius
  = \frac{1}{2}$ where $c$ denotes some universal positive constant.
\end{enumerate}
Given these properties, we now show how 
our general theory yields the results stated in Corollary~\ref{cor:grad_infor_response}.
To simplify the following discussion, we omit the universal constants and
a few-logarithmic terms, and track the dependency only on the sample size
$\obs$.

\paragraph{Results for gradient ascent:} 
The items~\ref{item:pop_gd} and \ref{item:sample_gd} establish that
the gradient operators are slow-convergent and stable, and thus we can apply
our general result from Theorem~\ref{theorem:slow_meta_theorem}.
In particular, plugging $\parbeta=\frac{1}{2}$, and $\pargamma=1$ in 
Theorem~\ref{theorem:slow_meta_theorem}, we find that the statistical error
for the gradient iterates $\tvar^{\iter} = (\samresga)^\iter(\tvar^0)$
satisfies
\begin{subequations}
\label{eq:sketch_gd_infor}
\begin{align}
\label{eq:rate_gd_infor}
  \abss{\tvar^\iter - \tstar } &\precsim [\noise(\obs, \tol)]^{\frac{\parbeta}
  {1+\parbeta-\pargamma\parbeta}}
  \asymp [n^{-\frac12}]^{\frac{1/2}{1+1/2-1/2}} = n^{-\frac 14},\\
  \text{for}\quad
\label{eq:steps_gd_infor}
  \iter &\succsim [\noise(\obs, \tol)]^{-\frac{1}
  {1+\parbeta-\pargamma\parbeta}} \asymp [n^{-\frac12}]^{-\frac{1}{1+1/2-1/2}}
  = n^{\frac12}.
\end{align}
\end{subequations}

\paragraph{Results for Newton's method:} 
The items~\ref{item:pop_newton} and \ref{item:sample_newton} establish
that the Newton operators are fast-convergent but unstable, and as a
consequence our general result from
Theorem~\ref{theorem:fast_unstab_meta_theorem}\ref{item:fast} can be
applied. In particular, plugging $\pargamma=-1$ in
Theorem~\ref{theorem:fast_unstab_meta_theorem}\ref{item:fast}, we find
that the Newton iterates $\tvar^{\iter} = (\samresnm)^\iter(\tvar^0)$
satisfy
\begin{align}
\label{eq:sketch_nm_infor}
  \abss{\tvar^\iter - \tstar } & \precsim \max\braces{[\noise(\obs, \tol)]^
  {\frac{1}
    {1+\abss{\pargamma}}}, \innerradius_\obs} \nonumber \\
  & \asymp [n^{-\frac12}]^{\frac{1}{1+1}} = \obs^{-\frac 14}
   \ \ \text{for}\  \
  \iter \succsim \log(1/\noise(\obs, \tol)) \asymp \log \obs.
\end{align}
Moreover, we show that (see the discussion around 
equation~\eqref{eq:lower_infor_resp}) Assumption~\ref{item:unstab_control} holds for the Newton iterates
with an initialization outside the ball $\ball(\tstar, \innerradius_\obs)$,
and hence part~\ref{item:cor_infor_newton} of the Corollary~\ref{cor:grad_infor_response}
states that the Newton iterates stay in a close vicinity of $\tstar$ for
all future iterations. 

\subsection{Over-specified Gaussian mixture models} 
\label{sub:over_specified_gaussian_mixture_models}
We now consider the problem of parameter estimation in Gaussian
mixture models; and analyze the behavior of two popular algorithms
namely (a) Expectation-Maximization (EM) algorithm~\citep{Rubin-1977},
and (b) Newton's method.  We note that EM is arguably the most widely
used algorithm for parameter estimation in mixture models and other
missing data problems~\citep{Rubin-1977}.  Here we study the problem of
estimating the parameters of a Gaussian mixture model given $\obs$
i.i.d.\ samples from the model. When the number of components in the
mixture is known, prior
works~\citep{Siva_2017,Daskalakis_colt2017,Cai_2018} have shown that
(i) the mixture parameters can be estimated at the parametric rate
$\obs^{-\frac{1}{2}}$ with the EM algorithm and (ii) the algorithm
takes at most $\log \obs$ steps to converge.  In the over-specified
setting, i.e., when the fitted model has more components than the true
model, recent
works~\citep{Raaz_Ho_Koulik_2018,Raaz_Ho_Koulik_2018_second,wu2019randomly}
have established the slow convergence of EM on both the statistical
and algorithmic fronts. For example, for over-specified
Gaussian-location mixtures EM takes $\obs^{\frac{1}{2}} \gg \log \obs$
steps (where $\gg$ denotes much greater than) to converge and produces
an estimate for the mean parameter that has a statistical error of
order $\obs^{-\frac{1}{4}} \gg \obs^{-\frac{1}{2}}$.  

In the sequel,
we apply our general theory to study the behavior of EM and Newton's
method for parameter estimation in over-specified Gaussian-location
mixtures.  First, we recover the slow convergence of EM as derived in
prior works~\citep{Raaz_Ho_Koulik_2018}.  Second, we prove that the
Newton's method---although an unstable algorithm in this
setting---achieves a similar statistical accuracy as EM albeit in an
exponentially fewer number of steps.  We now formalize the details.
\begin{subequations}
\label{eq:over_specified_gmm}
Let $\normDensity(\cdot; \tvar, \sd^2)$ denote the density of $\NORMAL(\tvar,
\sd^2)$ random variable, i.e., 
\begin{align}
\normDensity(x; \tvar, \sd^2) = (2 \pi \sd^2)^{- 1/ 2}e^{-
  \frac{(x - \tvar)^2}{2 \sd^2}}  
\end{align}
and let $X_{1},\ldots,X_ {n}$ be $n$ i.i.d. draws from the standard normal
distribution (density $\normDensity
(\cdot;0,1)$).  Given this data, we fit an over-specified mixture
model namely, a two-component symmetric Gaussian mixture with equal fixed
weights whose density is given by
\begin{align}
\label{Eqnsingular_location_scale}
    \FitDensity(x) 
    = \frac{1}{2}\normDensity(x;-\tvar, 1) + \frac{1}{2}\normDensity(x;\tvar,
    1),
\end{align}
where $\tvar$ is the parameter to be estimated. In such a setting, 
the true parameter is unique and given by $\thetastar=0$ since $f_0(\cdot)
= \normDensity(\cdot;0,1)$. However, the fact that we fit a mixture that
has one extra component than the true model (which has just one component)
leads to interesting consequences as we now elaborate. 
Using $\samlikelihood$ to denote the log-likelihood function, the MLE estimate
is given by
\begin{align}
\label{eq: sample_loglikelihood}
  \localmle \in \mathop {\arg \max}_{ \tvar \in
  \Rspace} \samlikelihood ( \tvar)
  \quad\text{where}\quad \samlikelihood ( \tvar) \mydefn \frac{1}{n} \sum_
  {i = 1}^{n} \log
  \FitDensity( X_{i}).
\end{align}
On one hand, it is known~\citep{Chen1992} that the over-specification in
such a setting leads to a slower than $n^{-\frac{1}{2}}$ statistical rate
for the MLE,
i.e.,  $\vert{\localmle-\tstar}\vert = \mathcal{O}(n^{-\frac{1}{4} })$.
On the other hand, MLE does not admit a closed-form expression and thus
it is of significant interest to understand the behavior of iterative algorithms
that are used to estimate the MLE. Next, we use our general framework to
provide a precise characterization of two algorithms namely, EM, and 
Newton's method on maximizing the log-likelihood 
$\samlikelihood$~\eqref{eq: sample_loglikelihood}.
\end{subequations}  


\subsubsection{Theoretical guarantees}

The next corollary provides a precise characterization of EM and Newton's
method for the over-specified setting described in the previous section.
We analyze the EM updates and the pure Newton updates. 
Moreover, we use $\nopem$ and $\samnewem$ respectively to denote
the sample-based operators for EM and Newton's method  
(see Appendix~\ref{ssub:proof_of_cor:Newton_mixture_model} for the precise form 
of these operators). Finally, the scalars $c, c_i, c_i', c_i''$ denote some
positive universal constants.

\begin{corollary} 
\label{corollary:Newton_mixture}
For the over-specified Gaussian mixture model~\eqref{eq:over_specified_gmm}
with $\tstar = 0$, given some $\tol \in (0,1)$, the following properties
hold with probability at least $1 - \tol$:
\begin{enumerate}[label=(\alph*)]
  \begin{subequations}
\item\label{item:cor_mix_em} For any fixed $\smallthreshold \in (0,
1/4)$ and initialization $\tvar^0 \in \ballnotag (\tstar, 1 )$,
the sequence $\tvar^\iter \defn (\nopem)^\iter(\tvar^0)$ of EM
iterates satisfies the bound
\begin{align}
  \abss{\tvar^\iter - \tstar } & \leq c_{1} \parenth{\frac{\log(
  \frac{\log(1/\smallthreshold)}{\tol})}
    {\obs}}^{\frac{1}{4}-\smallthreshold}
  \qquad \mbox{for all iterates $\iter \geq c_{1}' \sqrt{\obs} \log\frac{1}
  {\smallthreshold}$,}
\end{align}
as long as $n\geq c_1'' \log\frac{\log(1/\smallthreshold)}{\tol}$.
\item\label{item:cor_mix_nm} 
For any initialization $\tvar^0 \in \annulus ( \tstar, \frac{\sqrt{2 c} \log^2(3 n / \tol)}{n^{1/ 4}}, 1/ 3)$, the sequence of Newton iterates
$\tvar^\iter \defn
  (\samnewem)^\iter(\tvar^0)$ satisfies the bound
  \begin{align}
  \abss{\tvar^\iter - \tstar } & \leq c_{2} \parenth{\frac{\log(n/\tol)}
  {\obs}}^{\frac{1}{4}}
  \qquad \mbox{for all iterates $\iter \geq c_{2}' \log \obs$,}
  \end{align}
  as long as $n\geq c_2'' \log(1/\tol)$.
  \end{subequations}
\end{enumerate}
\end{corollary}
\noindent See Appendix~\ref{ssub:proof_of_cor:Newton_mixture_model} for
the proof (and below for the proof sketch).\\

Corollary~\ref{corollary:Newton_mixture} establishes
that the Newton EM is significantly faster than EM for the model setup~\eqref{eq:over_specified_gmm}.
More precisely, it reaches ball around $\tstar$ with a statistical radius of order $\obs^{-\frac14}$ within $\log \obs$ steps, which is much smaller than the
number of steps taken by EM. 
Moreover, the updates from Newton's method do not
escape this ball for future iterations.
This behavior is a consequence of the fact that under the assumed 
initialization condition, the (cubic-regularized)
Newton EM sequence satisfies assumption~\ref{item:unstab_control}. 

\begin{figure}[t!]
\begin{adjustbox}{width=0.95\textwidth,center=\textwidth} 
  \begin{tabular}{cc}
    \widgraph{0.5\textwidth, trim={0.2cm, 0, 0, 0}, clip}{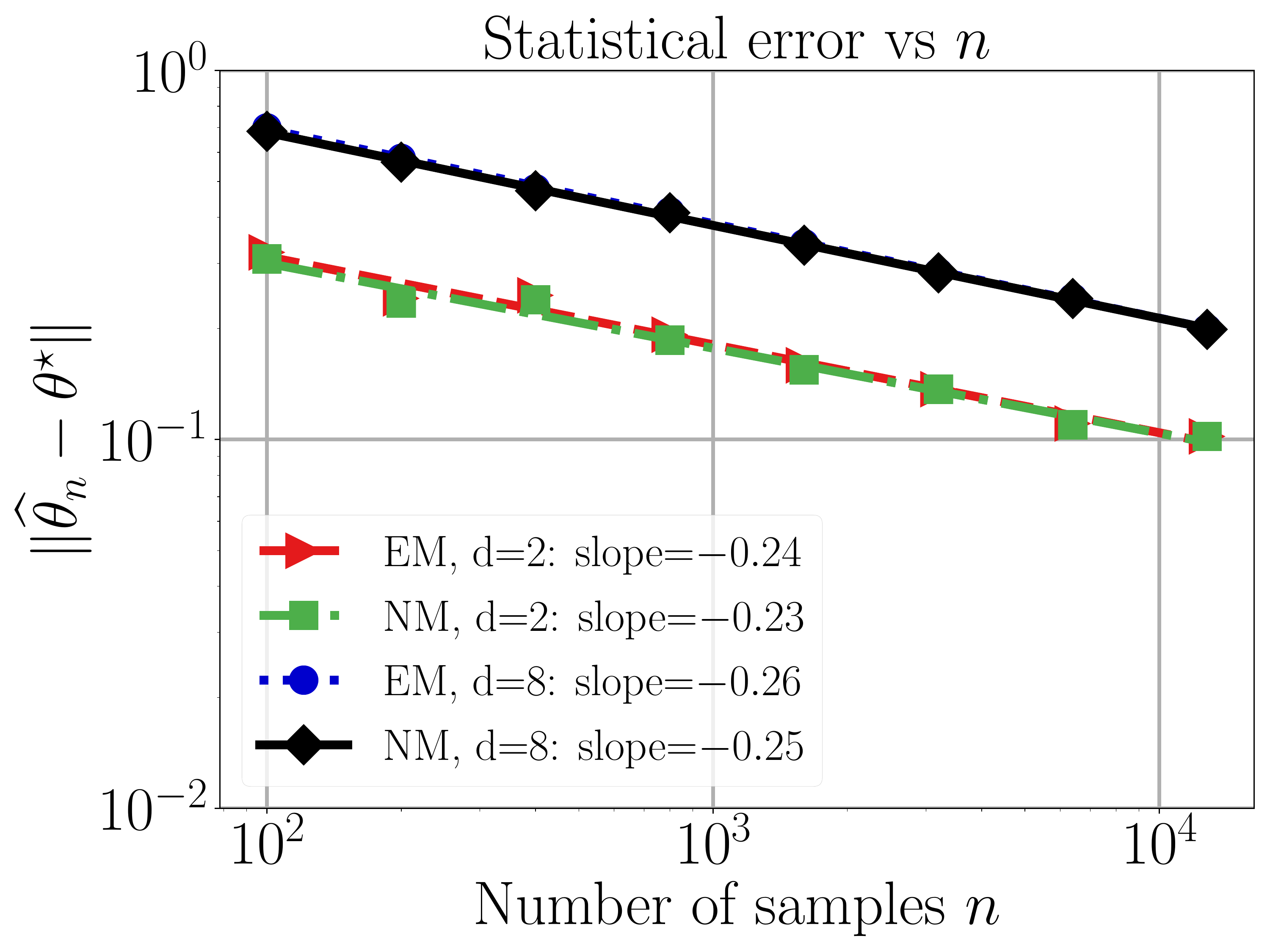}
    &
    \widgraph{0.5\textwidth, trim={0.2cm, 0, 0, 0}, clip}{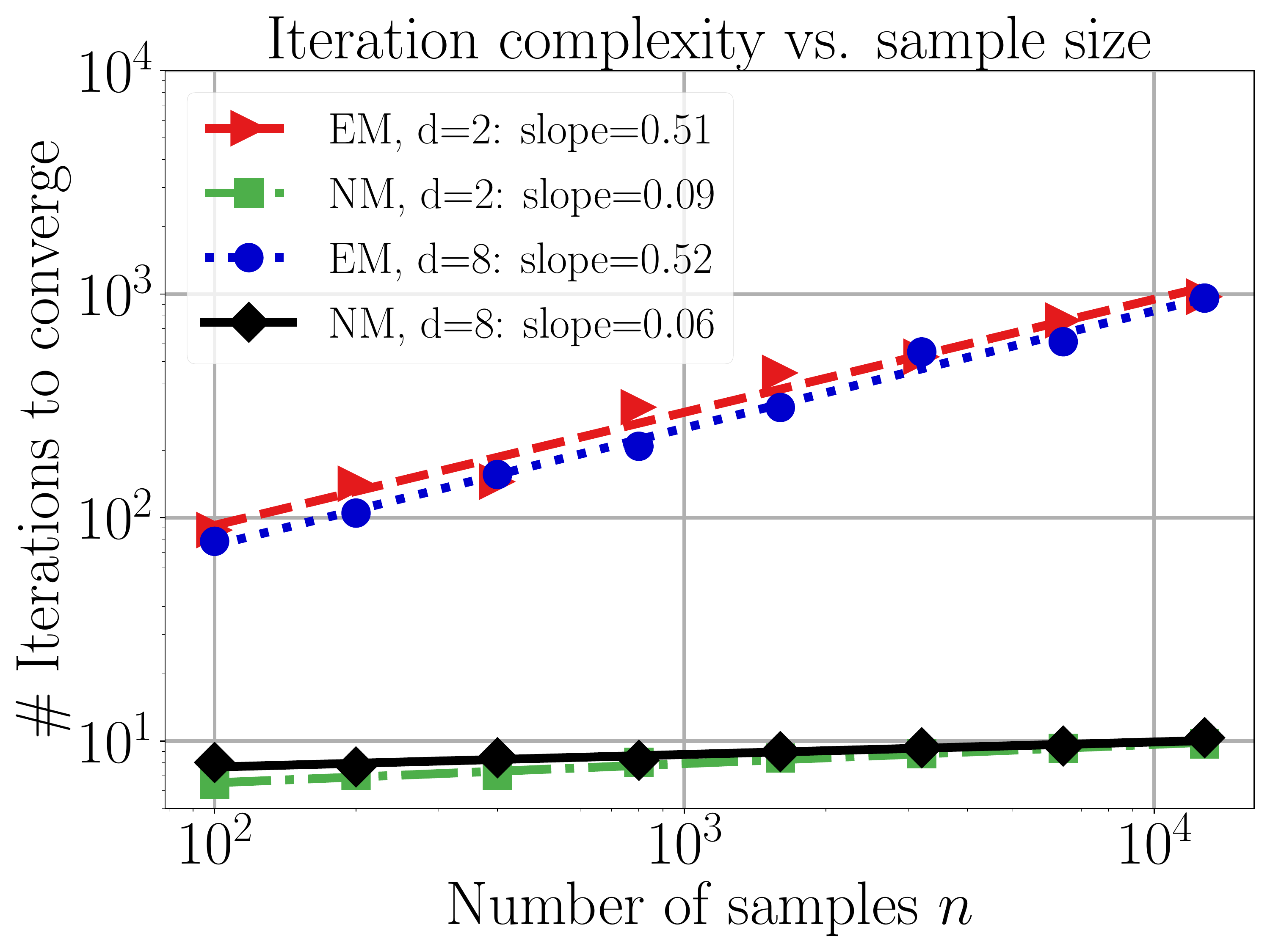} \\ (a) &
    (b)
  \end{tabular}
  \end{adjustbox}
  \caption{Plots characterizing the behavior of Expectation-Maximization (EM) and Newton's method (NM) for two Gaussian mixture models in $d=2$ and $d=8$ dimensions. (a) Log-log plots of the Euclidean
    distance $\|\widehat{\theta}_\obs - \tstar \|_2$ versus the sample
    size.  It shows that all the algorithms converge to an estimate at
    Euclidean distance of the order $n^{-1/4}$ from the true parameter
    $\tstar$.  (b) Log-log plots for the number of iterations taken by
    different algorithms to converge to the final estimate. While EM takes roughtly $\sqrt{n}$ iterations, the scaling of iterations taken by Newton's method is significantly slower.}
  \label{FigEMEmpirical}
\end{figure}

\paragraph{Multivariate settings:} In Figure~\ref{FigEMEmpirical}, we discuss the performance of EM and Newton's method under the multivariate setting of the over-specified Gaussian mixture model~\eqref{Eqnsingular_location_scale}. Similar to the univariate setting, both algorithms converge to a statistical error of order $(d/ n)^{1/4}$ around the true parameter $\tstar$.  Furthermore, the EM algorithm takes $\sqrt{n/ d}$ number of iterations to converge to the final estimate (see Appendix~\ref{sec:multivariate_mixture} for a formal result) while the Newton's method takes much fewer number of iterations (which seems in agreement with the $\log n$ scaling suggested by our theory). Given that each iteration of the EM algorithm takes order $n \cdot d$ arithmetic operations, the computational complexity for the EM algorithm to reach the final estimate is of order $n^{3/2} d^{1/2}$. On the other hand, each iteration of the Newton's method takes an order of $n \cdot d + d^3$ arithmetic operations where $d^3$ is computational complexity of computing inverse of an $d \times d$ matrix via Gauss-Jordan elimination approach. It leads to the computational complexity at the order $(n d+ d^3) \log n$ for the Newton's method to reach to the final estimate. Thus, when $d^{5/3} \ll n$, Newton's method is computationally more efficient than the EM algorithm.

\subsubsection{Proof sketch for Corollary~\ref{corollary:Newton_mixture}} 
\label{ssub:proof_sketch_for_corollary_corollary:newton_mixture}
The proof strategy for this case is similar to that laid out in 
Section~\ref{ssub:proof_sketch_for_cor_grad_infor_response} for informative non-response model.
First, to study this problem in our framework, we consider the population
level objective $ \likelihood$ by replacing the sum over samples in equation~\eqref{eq:
sample_loglikelihood} with the corresponding expectation:
\begin{align}
\label{eq:population_loglikelihood}
  \likelihood( \tvar) \mydefn \Exs_{X \sim \NORMAL(0, 1)} \brackets{ \log
    \FitDensity(X) } = \Exs_{X} \brackets{\frac{1}{2}
    \normDensity(X; - \tvar, 1) + \frac{1}{2} \normDensity(X;\tvar, 1) }.
\end{align}
Second, we use $\popem$ and $\newem$ respectively to denote the corresponding
population-level EM and Newton's method operators (see Appendix~\ref{ssub:proof_of_cor:Newton_mixture_model}
for the precise expressions). \\

\paragraph{Results for EM:}
For the case of $\tstar=0$, Theorem 2 and Lemma 1 of our prior works~\citep{Raaz_Ho_Koulik_2018}
show that, for any initialization $\tvar^0$, the EM operators $\popem$
and $\nopem$ satisfy
\begin{align}
\abss{(\popem)^\iter(\tvar^0) - \tstar}
& \leq \frac{c}{\iter^\frac{1}{2}}
\quad \text{and}, \nonumber \\
\sup_{\tvar \in \ball(\tstar, \radius)} \abss{\popem(\tvar) - \nopem(\tvar)}
& \leq c_{1} \radius \cdot \sqrt{\frac{\log(1/ \tol)}{n}},
 \label{eq:gradient_ascent_mixture_bounds}
\end{align}
where the second bound holds with probability at least $1- \delta$ for
any fixed radius $r > 0$. In the framework of our current work, 
the bounds~\eqref{eq:gradient_ascent_mixture_bounds} imply that the operator
$\popem$ exhibits $\slownotag(\frac{1}{2})$-convergence, and the operator
$\nopem$ is $\stabilitynotag(1)$-stable with the noise function $\sqrt
\frac{\log ( 1/ \tol)}{\obs}$.
Thus a direct application of Theorem~\ref{theorem:slow_meta_theorem}
of this paper (in a fashion similar to that of equations~\eqref{eq:rate_gd_infor} and~\eqref{eq:steps_gd_infor}),
recovers the main result of our prior work~\citep{Raaz_Ho_Koulik_2018}
(Theorem 3). That is, with high probability, the sequence $\tvar_\obs^{\iter+1}
= \nopem(\tvar_\obs^{\iter})$ satisfies
\begin{align}
\label{eq:rate_em}
  \abss{\tvar^\iter - \tstar } \precsim [n^{-\frac12}]^{\frac{1/2}{1+1/2-1/2}} = n^{-\frac 14}
  \qtext{for}
  \iter \succsim [n^{-\frac12}]^{-\frac{1}{1+1/2-1/2}} = n^{\frac12}.
\end{align}

\paragraph{Results for Newton's method:}
In Appendix~\ref{ssub:proof_of_cor:Newton_mixture_model}, we demonstrate the following properties of Newton's method operators:
\begin{enumerate}[label=(M\arabic*)]
\item\label{item:pop_newton_mixture} the Newton operator $\newem$ is $\fastnotag (\tfrac{7}{9})$-convergent over the ball
$\ball(\tstar, \tfrac{1}{3})$, and
\item\label{item:sam_newton_mixture} the operator $\samnewem$ is $\instabilitynotag(-
1)$-unstable over the annulus $\annulus ( \tstar, \innerradius_\obs, 1/ 3)$ with noise function $\noise( \obs,
\delta)~=~\frac{\log(n/ \delta)}{\sqrt{n}}$ where $\innerradius_\obs = \frac{c \log^2(3 n / \tol)}{n^{1/ 4}}$ .
\end{enumerate}
Based on the results of Theorem~\ref{theorem:fast_unstab_meta_theorem}\ref{item:fast} with $\kappa = \frac{7}{9}$ and $\pargamma = - 1$, 
the items~\ref{item:pop_newton_mixture} and~\ref{item:sam_newton_mixture} 
suggest that the Newton updates $\tvar^{\iter} = (\samresnm)^\iter(\tvar^0)$
satisfy
\begin{align}
\label{eq:sketch_nm_mixture}
  \abss{\tvar^\iter - \tstar } \precsim \max\braces{[\noise(\obs, \tol)]^
  {\frac{1}
    {1 + 1}}, \innerradius_\obs}
  \precsim \obs^{-\frac 14}
   \ \ \text{for}\  \
  \iter \succsim \log(1/\noise(\obs, \tol)) \asymp \log \obs.
\end{align}
Furthermore, we prove that the Newton iterates satisfy 
Assumption~\ref{item:unstab_control} (see the argument with equation~\eqref{eq:lower_mixture_model}). 
Therefore, the Newton iterates stay in a close vicinity of $\tstar$ for
all future iterations.  


\subsection{Non-linear regression model}
\label{subsec:single_index}
In our third example, we consider a non-linear regression
model~\citep{Carroll-1997} with a known link function $g$.  Models of
this type have proven useful for applications in signal processing,
econometrics, statistics, and machine learning~\citep{Ich93,
  Horowitz_direct}.  For simplicity, we briefly summarize the
one-dimensional version of this problem. The multivariate setting of the problem is considered in Appendix~\ref{sec:multivariate_nonlinear_regression}. We observe the pairs of data
\mbox{$(X_{i}, Y_{i}) \in \real^2$} that are generated from the model
\begin{subequations}
  \label{eq:single_index_model}
\begin{align}
\label{eq:single_index}
Y_{i} = g \parenth{X_{i} \tvar^{*}} + \newnoise_{i}\quad\text{for
}\quad \quad i = {1, \ldots, \obs}.
\end{align}
Here $Y_{i}$ denotes the response variable, $X_{i}$ corresponds to the
covariate and $\newnoise_{i}$ denotes the additive noise assumed to have
a standard Gaussian distribution, i.e., $\newnoise_i
\stackrel{\mathrm{i.i.d.}}{\sim}\NORMAL(0, 1)$. Note that, the Gaussianity of the additive noise is for the simplicity of the proof, and the results can be extended to sub-Gaussian errors.

In this example, we
consider the case of random design for the covariates, i.e., the
covariates $\{X_{i}\}_{i = 1}^n$ are independent and $X_i \sim
\NORMAL(0, 1)$.  Given the samples $\braces{(X_i, Y_i), i \in
  [\obs]}$, we want to estimate the unknown parameter $\tvar^{*}$.  A
popular choice is the maximum-likelihood estimate (MLE):
\begin{align}
\label{eq:sam_likeli_single_index}
\estimate^{\textrm{mle}} \in \arg\min_{\tvar \in \Rspace}
\likelihoodsing_{n}( \tvar)
\quad\text{where} \quad
\likelihoodsing_{n} \mydefn \frac{1}{2 \obs} \sum_{i = 1}^{\obs} 
\parenth{Y_{i} -  g \parenth{X_{i} \tvar}}^2.
\end{align}
Generally, the loss-function $\likelihoodsing_{n}$ is non-convex and hence
the MLE does not admit a closed-form expression. Consequently, one needs
to make use of certain optimization algorithms to compute an estimate~$\estimate$, which need not be the same as $\estimate^{\textrm{mle}}$.

In the remainder of this section, we study the case when the SNR degenerates
to zero. Specifically, we consider $\tvar^{*} = 0$ and a link function of
the form  $g(x) = x^{2p}$ with  $p \geq 1$.
For such a setting, the optimization problem~\eqref{eq:sam_likeli_single_index}
takes the following form:
\begin{align}
\label{eq:sam_likeli_single_index_new}
\estimate \in \arg\min_{\tvar \in \Rspace}
\likelihoodsing_{n}( \tvar)
\quad\text{where} \quad
\likelihoodsing_{n} \mydefn \frac{1}{2 \obs} \sum_{i = 1}^{\obs} 
\parenth{Y_{i} - \parenth{X_{i} \tvar}^{2 p}}^2.
\end{align}
\end{subequations}

\subsubsection{Theoretical guarantees}

For the non-linear regression model described above
with the link function $g(x) = x^{2p} $, we consider three iterative 
optimization methods: (a) gradient descent with a step
size $\learnrate \in (0, \frac{1}{(4 p - 1)!! (2 p)}]$, (b) (pure) Newton's
method, and (c) cubic-regularized Newton's
method with Lipschitz constant $\regular \mydefn (4 p - 1)!! (4 p - 1) p/ 3$. 
We denote the updates for these three methods via the
operators $\nopgd$, $\nopnm$, and $\nopcnm$ respectively (see 
Appendix~\ref{subsec:proof:corollary:grad_single_index} for the precise
expressions of these operators).
The next result characterizes the behavior of these three methods:

\begin{corollary} 
\label{cor:grad_single_index}
For the non-linear regression model~\eqref{eq:single_index_model} with link function
$g(x) = x^{2p}$ for $p\geq 1$ and true parameter $\tstar = 0$, given some $\tol
\in (0,1)$, the following properties hold with probability at least $1 -
\tol$:
\begin{enumerate}[label=(\alph*)]
  \begin{subequations}
\item\label{item:cor_single_gd} For any fixed $\smallthreshold \in (0,
1/4)$ and initialization $\tvar^0 \in \ballnotag (\tstar, 1 )$,
the sequence $\tvar^\iter \defn (\nopgd)^\iter(\tvar^0)$ of gradient
iterates satisfies the bound
\begin{align}
  \abss{\tvar^\iter - \tstar } & \leq c_{1} \parenth{\frac{\log^{4p}(
  n\frac{\log(1/\smallthreshold)}{\tol})}
    {\obs}}^{\frac{1}{4p}-\smallthreshold}
  \mbox{for all iterates $\iter \geq c_{1}' \obs^{\frac{2p-1}{2p}}
  \log\frac{1}
  {\smallthreshold}$,}
\end{align}
as long as $n\geq c_1'' \log\frac{\log(1/\smallthreshold)}{\tol}$.
\item\label{item:cor_single_nm} 
For any initialization $\tvar^0 \in \annulus( \tstar, c
\frac{\log^{p/ (2 p - 1)} (n/ \delta)}{n^{1/ 4 (2 p - 1)}},
1)$, the sequence of Newton iterates $\tvar^\iter \defn
  (\nopnm)^\iter(\tvar^0)$ satisfies the bound
  \begin{align}
  \abss{\tvar^\iter - \tstar } & \leq c_{2} \parenth{\frac{\log^{4p}(n/\tol)}
  {\obs}}^{\frac{1}{4p}}
  \qquad \mbox{for all iterates $\iter \geq c_{2}' \log \obs$,}
  \end{align}
  as long as $n\geq c_2'' \log(1/\tol)$.
\item\label{item:cor_single_cnm} 
The sequence of cubic-regularized Newton iterates 
$\tvar^\iter \defn (\nopcnm)^\iter(\tvar^0)$ with initialization $\tvar^0 \in \annulus( \tstar, c
\frac{\log^{p/ (2 p - 1)} (n/ \delta)}{n^{1/ 4 (2 p - 1)}},
1)$ satisfies the bound
  \begin{align}
  \abss{\tvar^\iter - \tstar } & \leq c_{3} \parenth{\frac{\log^{4p}(n/\tol)}
  {\obs}}^{\frac{1}{4p}}
  \qquad \mbox{for all iterates $\iter \geq c_{3}' \obs^{\frac{4p-3}{2(4p-1)}}$,}
  \end{align}
  as long as $n\geq c_3'' \log(1/\tol)$.
  \end{subequations}
\end{enumerate}
\end{corollary}
\noindent See Appendix~\ref{subsec:proof:corollary:grad_single_index} for
the proof (and below for the proof sketch).\\

This corollary shows that the final statistical errors achieved by gradient descent and the (cubic-regularized) Newton's method have the same scaling.  Moreover,
Newton's method, while unstable, converges to the correct statistical
radius in a significantly smaller $\log \obs$ number of steps when
compared to gradient descent, which takes $\obs^{\frac{2 p - 1}{2 p}}$
steps and cubic-regularized Newton's method, which takes $\obs^{\frac{4 p -
    3}{2 (4 p - 1)}}$ steps. 
Moreover, we also show that assumption~\ref{item:unstab_control} holds for
the iterates from the (cubic-regularized) Newton method's\footnote{See the proofs of
  equations~\eqref{eq:lower_single_index} 
  and~\eqref{eq:lower_cubic_single_index} in
  Appendix~\ref{subsec:proof:corollary:grad_single_index} for more
  details.} and hence we obtain
that these iterates not only converge to a ball of radius $\obs^{-\frac{1}
{4 p}}$ around $\tstar$, but also that they stay there for all the future
iterations. Finally, in
Appendix~\ref{subsec:proof:corollary:grad_single_index} (see
equation~\eqref{eq:global_min_single_index}) we also establish that
the statistical radius $n^{-1/ (4 p)}$ achieved by the considered optimization methods is tight.

When $g(x) = x^2$, the model~\eqref{eq:single_index} corresponds to a
phase retrieval problem.  In the regime of large signal-to-noise ratio
(SNR), i.e., $\abss{\tstar} \gg 1$, and with the link function $g(x) =
x^2$, there are efficient algorithms which produce an estimate
$\estimate$ satisfying a bound \mbox{$\vert \estimate-\tstar\vert
  \precsim \obs^{-\frac{1}{2}}$}~\citep{Eldar-2013, Candes-2015,
  Yanshuo-2018}.  However, as the SNR approaches zero these parametric
rates do not apply and precise statistical behavior of these estimates
are not known.

\subsubsection{Proof sketch for Corollary~\ref{cor:grad_single_index}}
\label{ssub:proof_sketch_for_corollary_cor:grad_single_index}

In order to study these updates using our framework, we need to
consider the population-level version of the optimization
problem~\eqref{eq:sam_likeli_single_index_new}, which is given by
\begin{align}
\notag
\min_{\tvar \in \Rspace}
\likelihoodsing( \tvar) \quad\text{where}\quad
\likelihoodsing( \tvar) \mydefn \frac{1}{2} \Exs_{(X, Y)} \brackets{
\parenth{Y
-  \parenth{X \tvar}^{2 p}}^2},
\end{align}
where the expectation is taken with respect to $X \sim \NORMAL(0, 1)$,
 $Y \sim \NORMAL(0, 1)$ as $\tstar = 0$.  Direct
computation yields that 
\begin{align}
\label{eq:pop_likelihood_single_index}
    \likelihoodsing( \tvar) = \frac{1}{2} + \frac{(4 p - 1)!!  \tvar^{4
 p}}{2}
 \qquad \text{and} 
 \qquad 
 \arg\min_\tvar\;\; \likelihoodsing(\tvar) = 0 = \tstar.
\end{align}
Like the previous proof sketches, we let $\popgd, \popnm$ and $\popcnm$ denote the population operators corresponding to the algorithms, gradient descent, Newton's method and cubic-regularized Newton's method, for the problem~\eqref{eq:pop_likelihood_single_index} 
(for a given $p$).
See Appendix~\ref{subsec:proof:corollary:grad_single_index} for the precise
definitions of these operators.
In Appendix~\ref{subsec:proof:corollary:grad_single_index}, we show that
with $\tstar=0$, these population-level operators satisfy
the following properties over the ball $\ball(\tstar, 1)$:
\begin{enumerate}[label=($\widetilde{\mathrm P}$\arabic*)]
\item\label{item:pop_single_index_gd} the gradient operator $\popgd$ is
$\slownotag(\frac{1}{4p-2})$-convergent for step size $\learnrate \in (0, \frac{1}{(4 p - 1)!!(2 p)}]$,
\item\label{item:pop_single_index_newton} the Newton operator $\popnm$ is
$\fastnotag(\frac{4p-2}{4p-1})$-convergent, and
\item\label{item:pop_single_index_cubic_newton} the cubic-regularized Newton
operator $\popcnm$ is $\slownotag(\frac{2}
{4p-3})$-convergent.
\end{enumerate}
Moreover in the Appendix~\ref{subsec:proof:corollary:grad_single_index},
we show that with the noise function
\mbox{$\noise(\obs, \tol) = \sqrt{ \frac{\log^{4p}(n/\tol)}{\obs}}$}, the
sample-level operators satisfy the following
properties:
\begin{enumerate}[label=($\widetilde{\mathrm S}$\arabic*)]
  \item\label{item:sample_single_index_gd} the operator $\nopgd$ is $\stabilitynotag
  (2 p - 1)$-stable over the ball $\ball(\tstar, 1)$, 
  \item\label{item:sample_single_index_newton} the operator $\nopnm$ is $\instabilitynotag(-(2p-1))$-unstable
  over the annulus $\Ann(\tstar, \innerradius_\obs, 1)$
  with inner radius \mbox{$\innerradius_\obs = c \; \log^{p/ (2 p - 1)}
  (n/ \tol)/ n^{1/4
    (2 p - 1)}$}, and 
  \item\label{item:sample_single_index_cubic_newton}
  the operator $\nopcnm$ is $\instabilitynotag(-\frac12)$-unstable over
  the annulus $\Ann(\tstar, \innerradius_\obs, 1)$.
\end{enumerate}
These properties show that the gradient descent is a slow-converging stable
method and we can apply Theorem~\ref{theorem:slow_meta_theorem}. On the
other hand, Newton's method is a fast-converging unstable method, and 
Theorem~\ref{theorem:fast_unstab_meta_theorem}\ref{item:fast}
can be applied. 
Finally, cubic-regularized Newton's method is a slow-converging unstable
method and Theorem~\ref{theorem:fast_unstab_meta_theorem}\ref{item:slow}
can be applied.
In the subsequent proof-sketch, we track the dependency only on the sample
size
$\obs$ and ignore logarithmic factors and universal constants. Moreover, since the computations here mimic the discussion from
Section~\ref{ssub:proof_sketch_for_cor_grad_infor_response}, we keep the
discussion briefer. 

\paragraph{Results for gradient descent:}
Applying Theorem~\ref{theorem:slow_meta_theorem} 
with $\parbeta= \frac{1}{4p-2}$, and $\pargamma= 2 p - 1$ 
(items~\ref{item:pop_single_index_gd} and \ref{item:sample_single_index_gd}
respectively), we find that the statistical error
for the gradient iterates $\tvar^{\iter} = (\nopgd)^\iter(\tvar^0)$
satisfy
\begin{align}
\label{eq:rate_gd_index}
  \abss{\tvar^\iter - \tstar } \precsim [\noise(\obs, \tol)]^{\frac{\parbeta}
  {1+\parbeta-\pargamma\parbeta}}
  \precsim n^{-\frac{1}{2p}} \ \ \text{for}\  \
  \iter \succsim [\noise(\obs, \tol)]^{-\frac{1}
  {1+\parbeta-\pargamma\parbeta}} \asymp n^{\frac{2 p - 1}{2 p}}.
\end{align}

\paragraph{Results for Newton's method:}
Next applying Theorem~\ref{theorem:fast_unstab_meta_theorem}\ref{item:fast}
for the Newton's method with $\kappa = \frac{4p - 2}{4p - 1}$, and $\pargamma = - (2 p - 1)$ (see items~\ref{item:pop_single_index_newton} and~\ref{item:sample_single_index_newton}), we conclude that the updates $\tvar^\iter =
  (\nopnm)^\iter(\tvar^0)$ from the Newton's method have the following property:
\begin{align}
\label{eq:sketch_nm_index}
  \abss{\tvar^\iter - \tstar } \precsim \max\braces{[\noise(\obs, \tol)]^
  {\frac{1}
    {1 + \abss{ \pargamma}}}, \innerradius_\obs}
  \precsim \obs^{-\frac {1}{2p}}
   \ \ \text{for}\  \
  \iter \succsim \log(1/\noise(\obs, \tol)) \asymp \log \obs.
\end{align}

\paragraph{Results for cubic-regularized Newton's method:}
Finally by using Theorem~\ref{theorem:fast_unstab_meta_theorem}\ref{item:slow}
for the cubic-regularized Newton's method with $\parbeta= \frac{2}{4p-3}$, and $\pargamma= - \frac{1}{2}$ (see items~\ref{item:pop_single_index_cubic_newton} and~\ref{item:sample_single_index_cubic_newton}), the following results hold for the cubic-regularized Newton iterates 
$\tvar^\iter = (\nopcnm)^\iter(\tvar^0)$: 
\begin{align}
\label{eq:sketch_cnm_index}
  \abss{\tvar^\iter - \tstar } & \precsim \max\braces{[\noise(\obs, \tol)]^
  {\frac{\parbeta}
    {1 + \parbeta - \pargamma \parbeta}}, \innerradius_\obs} \nonumber \\
  & \precsim \obs^{-\frac {1}{2p}}
   \ \ \text{for}\  \
  \iter \succsim [\noise(\obs, \tol)]^{-\frac{1}{1 + \parbeta}} \asymp \obs^{\frac{4p-3}{2(4p-1)}}.
\end{align}
\section{Discussion}
\label{sec:discussion}

In this paper, we established several results characterizing the
statistical radius achieved by a sequence of updates
$\{\nop^{t} (\tvar_{n}^{0})\}_{t \geq 0}$,
induced by an operator $\nop$ and a given initial point $\tvar_{n}^{0}$.
We established these results by analyzing the interplay between (in)-stability
of the operator $\nop$ for its population operator $\pop$ and
the local convergence of $\pop$ around its fixed point $\tstar$. 
We then applied our general theory to derive sharp algorithmic and statistical
guarantees for several iterative algorithms by analyzing
the corresponding sample and population operators, in three different 
statistical settings. In particular, we studied the behavior of
gradient methods and higher-order (cubic-regularized) Newton's method
for parameter estimation---in the weak signal-to-noise ratio regime---in
Gaussian mixture models, non-linear regression models, and informative non-response
models. We showed that for such models, despite instability, fast algorithms
like Newton's method may still be preferred over a stable one like gradient
descent since they achieve the same statistical accuracy as that of the stable counterpart in exponentially fewer steps.

We now discuss a few questions that arise naturally from our work.
First, our results, as stated, are not directly applicable to the
settings of accelerated optimization methods or quasi-Newton methods,
e.g., accelerated gradient descent~\citep{Nesterov-2013-Introductory}
and L-BFGS~\citep{Fletcher-1987-Practical}.  On the one hand, the
updates from an accelerated gradient descent method require that the
operators $\nop$ and $\pop$ to change with each iteration. On the
other hand, the updates from the L-BFGS method would require
additional machinery to deal with the preconditioning matrices in each
step.  Developing a general theory to characterize the statistical
performance of algorithms associated with a time-varying operator
$\nop$ is an interesting direction for future research.

Secondly, it is desirable to understand the behavior of optimization
methods to a wider range of statistical problems. In the context of
mixture models, recent work by Dwivedi et
al.~\citep{Raaz_Ho_Koulik_2018_second} established that for
over-specified mixtures with both location and scale parameter
unknown, EM takes an $\mathcal{O}(n^{\frac34})$ steps to return
estimates with minimax statistical error of order $n^{-\frac18}$ and
$n^{-\frac14}$ for the location and scale parameter, respectively.
Whether an unstable method like (cubic-regularized) Newton's EM proves
computationally advantageous (without losing statistical accuracy) in
such more challenging non-convex landscapes remains an open problem.

Finally, our theory does not easily extend to the settings with
dependent data, such as time series. When the samples are (time)
dependent, taking the limit of infinite sample size does not yield a
natural population-level operator. One possible fix is to borrow the
technique of truncating the sample operator from the analysis of the
Baum-Welch algorithm for hidden Markov
models~\citep{Fanny-2017}. However, even with the help of such a
technique, ample technical challenges remain towards developing a
general theory for such non-i.i.d. settings.


\appendix

\newcommand{\NM}{Q^{\textrm{NM}}}
\newcommand{\GD}{Q^{\textrm{GD}}}
\renewcommand{\stepsize}{\ensuremath{\eta}} 

In this supplementary material, we provide the details of proofs and
results that were deferred from the main
paper. Appendices~\ref{sec:Proof} and~\ref{label:tehcnical_lemmas}
contain the proofs of Theorems~\ref{theorem:slow_meta_theorem}
and~\ref{theorem:fast_unstab_meta_theorem}, respectively, including
all the details of the localization argument and the proofs of all
auxiliary technical lemmas. In Appendix~\ref{sec:lower_bounds}, we
construct a simple class of problems to demonstrate that the
guarantees Theorems~\ref{theorem:slow_meta_theorem}
and~\ref{theorem:fast_unstab_meta_theorem} are unimprovable in
general. Finally, in Appendix~\ref{sec:proofs_of_corollaries}, we
collect the proofs of several corollaries stated in the paper.
Finally, we discuss an extension of the theoretical results in the
main text to multivariate settings in
Appendix~\ref{sec:extension_multivariate}.


\appendix

\section{Proofs of main results}
\label{sec:Proof}

In this section, we provide the proofs of our main results, namely
Theorems~\ref{theorem:slow_meta_theorem}
and~\ref{theorem:fast_unstab_meta_theorem}.


\subsection{Proof of Theorem~\ref{theorem:slow_meta_theorem}}
\label{proof:theorem:slow_meta_theorem}

The reader should recall the proof outline provided following the
statement of the theorem.  Our proof here follows this outline, making
each step precise.  For the remainder of the proof, we assume without
loss of generality that $\tstar = 0$ and $r_0 = 1$. Proofs for the
cases $\tstar \neq 0$ or $r_0 > 1$ can be reduced to this case in a
straightforward fashion and are thereby omitted.


\subsubsection{Notation for stable case}
\begin{subequations}

For each positive integer $\ind = 1, 2, \ldots$, let
$\numsteps_{\ind}$ denote the number of iterations during the
$\ind$-th epoch, and let $\totalnumsteps_{\ind}$ denote the total
number of iterations taken up to the completion of epoch $\ind$.  In
order to describe some recursions satisfied by these quantities, we
define
\begin{align}
\begin{gathered}
\label{eq:time_one_and_two}   
  \numsteps_{\ind}^{(1)} \defn \bigc \mydndelta^{-\frac{
      \seqalpha{\ind-1}(\pargamma) + 1}{1 + \parbeta}} \quad \text{
    and }\quad \numsteps_{\ind}^{(2)} \defn \bigc' \mydndelta^{-\frac{
      \seqalpha{\ind}(\pargamma) + 1}{1 + \parbeta}}, \\
\quad \text{for} \quad \bigc \defn {(\unicontwo
  2^\pargamma)^{-\frac{1}{(1+\parbeta)}}} \quad \quad \text{ and }
\qquad \quad \bigc' \defn \bigc
(\constfn)^{\frac{\pargamma}{1+\parbeta}},
\end{gathered}
\end{align}
where $\constfn \defn (\unicontwo2^\pargamma)^{\frac{\parbeta}{1 +
    \parbeta}} = \bigc^{-\parbeta}$ and hence we have $\bigc' =
\bigc^{\frac{1 + \parbeta + \parbeta \pargamma}{1 + \parbeta}}$.  Here
the constant $\unicontwo$ is the constant from the the stability
definition~\eqref{eq:sample_stability}.  The sequences
$\{\numsteps_\ind \}$ and $\{\totalnumsteps_\ind \}$ have the
following properties: with the initialzation $\numsteps_{0} \defn 0$,
we have
\begin{align}
\label{eq:time_sequence}  
\numsteps_{\ind} \defn \ceil{\numsteps_{\ind}^{(1)} +
  \numsteps_{\ind}^{(2)}} \quad \mbox{and} \quad \totalnumsteps_{\ind}
\defn \sum_{j = 0}^\ind \numsteps_j \quad \mbox{for $\ind = 1, 2,
  \ldots$.}
\end{align}
\noindent Our proof is based on studying the sequence of real-numbers $\{\seqalpha{\ind}\}_
{\ind \geq 0}$ given by
\begin{align}
\label{eq:alpha_seq}
\seqalpha{0} = 0 \quad \text{and} \quad \seqalpha{\ind + 1} =
\seqalpha{\ind} \parnuone + \parnutwo, \quad \mbox{where $\parnuone=
  \frac{\parbeta \pargamma}{1 + \parbeta}$ and $\parnutwo =
  \frac{\parbeta}{ 1 + \parbeta}$.}
\end{align}
Note that Assumption~\ref{item:sample_stable} 
implies that $\parnuone \in (0, 1)$ and hence
\begin{align}
    \label{eq:alpha_seq_defn}
    \seqalpha{\ind} = \parfinal_\star(1-\parnuone^{\ind})
    \uparrow \parfinal_\star\quad \text{where}\quad 
    \parfinal_\star \mydefn
    \frac{\parbeta}{1 + \parbeta - \pargamma \parbeta}.
\end{align}
\end{subequations}
\noindent In the epoch-based argument, we need to control the deviation 
$\sup_{\enorm{\tvar} \leq \radius} \enorm{\pop(\tvar) - \nop(\tvar)}$
uniformly for each radii $r \in \radii'$. To this end, for any 
tolerance $\tol \in (0,1)$, we define the event $\event$ by
\begin{align}
\label{eqn:event_definition}
\event \mydefn \left\lbrace \sup_{\tvar\in\ball(\tstar, \radius)}\enorm{ \pop(\tvar) - \nop(\tvar)}
  \leq \unicontwo  \radius^\pargamma \mydndelta
    \quad \text{ uniformly for all } \radius \in \radii'
    \right\rbrace, 
\end{align}  
where $ \tol^* = \tol
    \cdot\frac{\log(\frac{1+\parbeta}{\parbeta\pargamma})}
              {8\log(\frac{\parbeta}{\smallthreshold(1 + \parbeta
                  -\pargamma \parbeta}))}$ 
was defined in equation~\eqref{eq:sample_size} and the radii-set $\radii'$
is defined as
\begin{align}
\begin{gathered}
    \label{eq:full_r}
    \radii' \defn  \radii \cup  2\radii,
    \quad\text{with}\\
    \radii \defn \braces{ \mydndelta^{\seqalpha{0}}, \ldots, 
                \mydndelta^{\seqalpha{\imax}},
                \constfn\mydndelta^{\seqalpha{0}},
                \ldots,
                \constfn\mydndelta^{\seqalpha{\imax}}
                }, \\
\imax = \ceil{\log(1/\paralpha)} \quad  \text{ and }
 \quad  \constfn = (\unicontwo2^\pargamma)^{\frac{\parbeta}{1+\parbeta}}.
 \end{gathered}
\end{align}
Combining the \stability-stability assumption~\eqref{eq:sample_stability}
with a standard application of union bound we conclude that
\begin{align}
    \label{eqn:event_prob_lb}
    \Prob (\event) \geq 1 - \tol.
\end{align} 
Before we start the main argument, we state a lemma useful in the
proof of our theorem:
\begin{lemma}
\label{lemma:t_perturbation_error}
Assume that the assumptions of Theorem~\ref{theorem:slow_meta_theorem} are
in force. Then conditioned on the event $\event$~(\ref{eqn:event_definition}.
\ref{eqn:event_prob_lb}), for all radius $\radius$
in the set $\radii$~\eqref{eq:full_r}, we have
\begin{align}
  \label{eq:t_error}
  \sup_{\tvar\in\ball(\tstar,\radius)}\enorm{ \pop^\iter(\tvar) -
    \nop^\iter (\tvar)} \leq \unicontwo (2\radius)^\pargamma
  \mydndelta \cdot \iter \quad \mbox{ for all $\iter \leq
    \tfun(\radius)$},
\end{align}
where $\tfun(\radius) \defn \frac{\radius^{1-\pargamma}}{2^\pargamma
    \unicontwo\mydndelta}$. Furthermore, for all  $\ind \leq \imax$
we have
\begin{align}   
\label{eq:time_bounds_validity2} 
 \numsteps^{(1)}_{\ind+1} \leq \tfun(\mydndelta^{\seqalpha{\ind}})
\quad\text{and}\quad \numsteps^{(2)}_{\ind+1} \leq
\tfun(\constfn\mydndelta^{\seqalpha{\ind+1}}).
\end{align}
\end{lemma}

\noindent See Appendix~\ref{ssub:proof_of_lemma_lemma:t_perturbation_error}
for the
proof of this lemma.\\


\subsubsection{Main argument}
\label{ssub:main_argument}
We claim that the sequence $\{\tvar_\obs^\iter \}_{\iter \geq 1}$
satisfies
\begin{subequations}
\begin{align}
  \label{eq:epoch_result_a}
  \Vert{ \tvar_n^{\totalnumsteps_{\ind}}}\Vert_2 &\leq
  \mydndelta^{\seqalpha{\ind}} \qquad\text{uniformly for all }
  \ind \in \braces{0, 1, \ldots, \imax}, \qquad \text{and} \\
  \label{eq:epoch_result_b}
  \enorm{\tvar_n^{\totalnumsteps_{\imax} +\iter }} &\leq
  2\mydndelta^{\parfinal_\star - \smallthreshold} \quad
  \mbox{uniformly for all  $\iter \in \braces{0, 1, 2, \ldots}$},
\end{align}
 with probability at least $1 - \tol$.
\end{subequations}
The quantities $\seqalpha{\ind}, \totalnumsteps_{\ind}$ and $\imax$
are defined in equations~\eqref{eq:time_one_and_two}
through equation~\eqref{eq:alpha_seq}. With these claims at our disposal, 
it remains to prove an upper bound on the scalar
$\totalnumsteps_{\imax}$.  Towards this end, doing some 
straightforward algebra we find that
\begin{align}
\label{eq:numsteps_bound}
\numsteps_{\ind} & \leq \numsteps_{\imax} \leq \unicon'
\mydndelta^{-\frac{\parfinal_\star}{\parbeta}}\quad
\text{for any}\quad 0 \leq \ind \leq \imax.
\end{align}
Combining the above bounds on  $\numsteps_{\ind}$ with the definition of
$\totalnumsteps_{\ind}$ from  equation~\eqref{eq:time_sequence} yields an
upper bound  on  $\totalnumsteps_{\imax}$.
Substituting the upper bound on $\totalnumsteps_{\imax}$ in 
inequality~\eqref{eq:epoch_result_b} yields the claimed bound~\eqref{eqn:slowsta_meta_result}
of Theorem~\ref{theorem:slow_meta_theorem}. 
We now prove the claims~\eqref{eq:epoch_result_a} and~\eqref{eq:epoch_result_b}
using induction.


\subsubsection{Proof of claim~\eqref{eq:epoch_result_a}} 
\label{par:proof_of_claim_eq:epoch_result_a_}

We condition on the event $\event$ defined in the
equation~\eqref{eqn:event_definition}, which occurs with probability at
least $1 - \tol$, and establish the claim using induction on the epoch
index $\ind$. The base case $\ind=0$ is immediate.
We now establish the inductive step, i.e., given
$\enorm{\tvar_n^{\totalnumsteps_{\ind}}} \leq
\mydndelta^{\seqalpha{\ind}}$ for some $\ind \leq \imax-1$, we show
that $\enorm{\tvar_n^{\totalnumsteps_{\ind + 1}}} \leq
\mydndelta^{\seqalpha{\ind + 1}}$.  We split the proof in two parts
(primarily to handle the constants):
\begin{subequations}
\begin{align}
\label{eqn:theta_first_bound_within_epochs}
\enorm{\tvar_n^{\totalnumsteps_{\ind} +
    \numsteps^{(1)}_{\ind+1}} } &\leq
\constfn\mydndelta^{\seqalpha{\ind+1}} \quad\text{and} \\
\label{eqn:theta_second_bound_within_epochs}
    \enorm{\tvar_n^{\totalnumsteps_{\ind} +
            \numsteps^{(1)}_{\ind+1} + \numsteps^{(2)}_{\ind+1}} } &
        \leq \mydndelta^{\seqalpha{\ind+1}},
 \end{align}
\end{subequations}
where $c'>1$ is a universal constant.
These claims together imply the induction hypothesis and thereby the
claim~\eqref{eq:epoch_result_a}.  


\paragraph{Proof of claim~\eqref{eqn:theta_first_bound_within_epochs}} 
\label{par:proof_of_claim_eqn:theta_first_bound_within_epochs}

Inequality~\eqref{eq:time_bounds_validity2} implies that $\numsteps^
{(1)}_{\ind+1} \leq \tfun(\mydndelta^{\seqalpha{\ind}})$, and hence we
can apply the bound~\eqref{eq:t_error} from
Lemma~\ref{lemma:t_perturbation_error} with $\radius =
\mydndelta^{\seqalpha{\ind}} \in \radii$ for any $\iter
\leq\numsteps^{(1)}_{\ind+1}$.  Applying the triangle inequality
yields
\begin{align}
  \notag
  \enorm{\tvar_n^{\iter + \totalnumsteps_{\ind}}} =
  \enorm{\nop^{\iter}(\tvar_n^{\totalnumsteps_{\ind}})} & \leq
  \enorm{\pop^{\iter} (\tvar_n^{\totalnumsteps_{\ind}})} +
  \enorm{\pop^{\iter}(\tvar_n^{\totalnumsteps_{\ind}}) - \nop^{\iter}
    (\tvar_n^{\totalnumsteps_{\ind}}) } \\
\label{eq:first_half_induction_intermediate}   
  & \stackrel{(i)}{\leq} \frac{1}{\iter^{ \parbeta}} +
\enorm{\pop^{\iter}(\tvar_n^{\totalnumsteps_ {\ind}}) -
  \nop^{\iter}(\tvar_n^{\totalnumsteps_{\ind}}) } \\
\label{eq:first_half_induction}
  & \stackrel{(ii)}{\leq} \frac{1} {\iter^{ \parbeta}} + \unicontwo(2
\mydndelta^{\seqalpha{\ind}})^\pargamma \mydndelta\iter,
\end{align}
for any $\iter \leq \numsteps^{(1)}_{\ind+1}$; where step~(i) follows
from the \slow-convergence~\eqref{eq:slow_convergence} of the operator
$\pop$ along with the assumption that $\tstar =0$, and step~(ii) follows by 
using the inductive hypothesis $\enorm{\tvar_n^
  {\totalnumsteps_{\ind}}} \leq \mydndelta^{\seqalpha{\ind}}$
and applying Lemma~\ref{lemma:t_perturbation_error} with
$\radius= \mydndelta^{\seqalpha{\ind}}$.  Note that in the final
bound~\eqref{eq:first_half_induction} the first term decreases with
iteration $\iter$ while the second term increases with $\iter$.  In
order to trade off these two terms,\footnote{We ignore the effect of
  the ceiling function $\ceil{\cdot}$ to simplify the computations}
  we set $\iter =
\numsteps^{(1)}_{\ind+1}$~\eqref{eq:time_one_and_two} in the
bound~\eqref{eq:first_half_induction}
and find that
\begin{align*}
  \enorm{\tvar_n^{\totalnumsteps_{\ind}+\numsteps^{(1)}_{\ind+1}}}  
  & \leq \frac{1}{{(\numsteps^{(1)}_{\ind+1})}^{\parbeta}}
  + \unicontwo(2 \mydndelta^{\seqalpha{\ind}})^\pargamma
  \mydndelta \numsteps^{(1)}_{\ind+1} \\
&= \underbrace{2(\unicontwo2^\pargamma)^{\frac{\parbeta}{1+\parbeta}}}_
             {\rdefn \constfn} \cdot
             \mydndelta^{1 - \frac{\seqalpha{\ind} \pargamma + 1}{1 + \parbeta}
             + \seqalpha{\ind}\pargamma}
             \\
             & = \constfn \mydndelta^{\frac{\seqalpha{\ind}
                 (\parbeta\pargamma)+ \parbeta}{1 + \parbeta}} \\
             & = \constfn \mydndelta^{\seqalpha{\ind+1}},
\end{align*}
where the last equality follows from the 
relation~\eqref{eq:alpha_seq} between $\seqalpha{\ind}$ and
$\seqalpha{\ind+1}$.  The claim~\eqref{eqn:theta_first_bound_within_epochs}
now follows.


\paragraph{Proof of claim~\eqref{eqn:theta_second_bound_within_epochs}} 
\label{par:proof_of_claim_eqn:theta_second_bound_within_epochs}
For any $\iter \leq \tfun(\constfn\mydndelta^{\seqalpha{\ind+1}})$, we have
\begin{align*}
\enorm{\tvar_n^{\iter + \totalnumsteps_{\ind} +
    \numsteps^{(1)}_{\ind+1} } } & \leq \enorm{\pop^{\iter}
  (\tvar_n^{\totalnumsteps_{\ind} + \numsteps^{(1)}_{\ind+1} })} +
\enorm{\pop^{\iter}(\tvar_n^{\totalnumsteps_{\ind} + \numsteps^{(1)}_
    {\ind+1} }) - \nop^{\iter}(\tvar_n^{\totalnumsteps_{\ind} +
    \numsteps^{(1)}_{\ind+1} }) } \\
    &\leq 
    \frac{1}{\iter^\parbeta}+ 
  \unicontwo (2\constfn \mydndelta^{\seqalpha{\ind + 1}})^\pargamma
  \mydndelta \iter,
\end{align*}
where the last inequality follows from arguments similar to those used
to establish the
inequalities~\eqref{eq:first_half_induction_intermediate}
and~\eqref{eq:first_half_induction} above. Next, recalling the
inequality $\numsteps^{(2)}_{\ind+1} \leq \tfun( \constfn\mydndelta^{
  \seqalpha{\ind+1} })$ from equation~\eqref{eq:time_bounds_validity2}
and plugging $\iter =
\numsteps^{(2)}_{\ind+1}$~\eqref{eq:time_one_and_two} in the above
inequality, we find that
\begin{align*}
    \enorm{\tvar_n^{\totalnumsteps_{\ind+1} } }
    &\leq \underbrace{2 (\unicontwo2^\pargamma)^{\frac{\parbeta}{1+\parbeta}}
    \constfn^{\frac{\parbeta\pargamma}{1+\parbeta}}}_{=:\widetilde{C}}\cdot
    \mydndelta^{\frac{\seqalpha{\ind+1}\parbeta\pargamma +
    \parbeta}{1 + \parbeta}}
    = \widetilde{C} \mydndelta^{\seqalpha{\ind+2}}.
\end{align*}
In order to complete the proof, it remains to show that last quantity is upper bounded by $\mydndelta^{\seqalpha{\ind+1}}$;
equivalently, we need to verify the following upper bound
\begin{align}
\label{eqn:main_inequality}
    \specialeps \leq
        \frac{1}{\widetilde{C}^{\seqalpha{\ind + 2} - \seqalpha{\ind + 1}}},
\end{align}
which is equivalent to the large sample-size assumption~\ref{item:sample_size}
(see condition~\eqref{eq:small_constant_c} for a more precise statement)
if we establish that
\begin{align}
\label{eqn:diff_lambda_lb}
    \seqalpha{\ind + 2} - \seqalpha{\ind + 1} \geq \smallthreshold_\star 
    \mydefn \frac{\smallthreshold
          (1 + \parbeta - \parbeta \pargamma)} {1+\parbeta}.
\end{align}
In order to do so, we use the fact~\eqref{eq:alpha_seq_defn} that
$\seqalpha{\ind} = \parnuone_\star (1 - \parnuone^\ind)$ and obtain that
\begin{align*}
    \seqalpha{\ind} \leq \parnuone_\star - \paralpha
    \quad\text{and consequently that}\quad
    \parfinal_\star \parnuone^\ind \geq \paralpha
\end{align*}
for 
all $\ind \in \{0, 1, \ldots, \imax \}$.
Putting together the pieces we have
\begin{align*}
    \seqalpha{\ind + 2} - \seqalpha{\ind + 1} 
    = \parfinal_\star \parnuone^{\ind + 1} (1 - \parnuone)
    \geq \paralpha (1 - \parnuone)
    = \smallthreshold_\star,
 \end{align*} 
which yields the claimed bound~\eqref{eqn:diff_lambda_lb} and we are done.

\subsubsection{Proof of claim~\eqref{eq:epoch_result_b}} 
\label{ssub:proof_of_claim_eq:epoch_result_b_}

The proof of this claim follows a similar road-map as that in the previous
Section, and hence we simply sketch it. Conditional on the event
$\event$, we claim that
\begin{align}
  \label{eq:epoch_result_c}
  \enorm{\tvar_n^{\totalnumsteps_{\imax} +\counter\numsteps_{\imax} }}
  & \leq \mydndelta^{\parfinal_\star - \smallthreshold} \quad
  \mbox{uniformly for all $\counter \in \braces{0, 1, 2, \ldots}$.}
\end{align}
Assuming this bound is given for now, we complete the
proof.  Invoking inequality~\eqref{eq:bound_on_fn} from the proof of
Lemma~\ref{lemma:t_perturbation_error}, we obtain that
\begin{align}
\label{eqn:relation_1}
  \enorm{\tvar_n^{\totalnumsteps_{\imax} +\counter\numsteps_{\imax} +\iter
  }} \leq 2\mydndelta^{\parfinal_\star - \smallthreshold} 
\end{align}
for all $\counter\in\braces{1, 2, \ldots}$ and $\iter
    \leq \tfun
    (\mydndelta^{\parfinal_\star - \smallthreshold})$.
Mimicking the arguments from claims~\eqref{eqn:theta_first_bound_within_epochs}
and \eqref{eqn:theta_second_bound_within_epochs}, and using
the large sample-size assumption~\ref{item:sample_size} 
(condition~\eqref{eq:small_constant_c}) yields the claim~\eqref{eqn:relation_1}
for any $\iter \leq \mydndelta^{-\frac{\parfinal_\star}
{\parbeta}}$.
Putting this together with the fact~\eqref{eq:numsteps_bound} 
that $\numsteps_{\imax} \leq \mydndelta^{-\frac{\parfinal_\star}{\parbeta}}$
implies the claim~\eqref{eq:epoch_result_b}.

Turning to the proof of claim~\eqref{eq:epoch_result_c}, we note
that the base case $\counter=0$ follows from the claim~\eqref{eq:epoch_result_a}
by plugging in $\ind =\imax$.  For the  inductive step, assuming
\mbox{$\enorm{\tvar_n^{\totalnumsteps_{\imax} +\counter\numsteps_{\imax} }}
\leq \mydndelta^{\parfinal_\star - \smallthreshold}$}, arguments
similar to that in the proof of
claims~\eqref{eqn:theta_first_bound_within_epochs} and
\eqref{eqn:theta_second_bound_within_epochs} yield
\begin{align*}
  \enorm{\tvar_n^{\totalnumsteps_{\imax} + \counter\numsteps_{\imax} +
      \numsteps^{(1)}_{\imax} } } & \leq
  \constfn\mydndelta^{\parfinal_\star-\smallthreshold}
  \quad\text{and}, \\ \enorm{\underbrace{\tvar_n^{\totalnumsteps_{\imax}
        +\counter\numsteps_{\imax} + \numsteps^{(1)}_{\imax}
        +\numsteps^{(2)}_{\imax}}}_{\tvar_n^{\totalnumsteps_{\imax}
        +(\counter+1)\numsteps_{\imax} }} } & \leq
        \mydndelta^{\parfinal_\star-\smallthreshold},
\end{align*}
thereby establishing the induction hypothesis.

  
\subsection{Proof of Theorem~\ref{theorem:fast_unstab_meta_theorem}}
\label{proof:theorem:fast_unstab_meta_theorem}

We divide the proof into two subsections, corresponding to parts \ref{item:fast}
and \ref{item:slow} of Theorem~\ref{theorem:fast_unstab_meta_theorem}.


\subsubsection{Proof of part~\ref{item:fast}}
\label{ssub:proof_of_part_a}
We introduce the shorthands $\noiseunstable(\obs, \tol) =
(\noise(\obs, \tol))^{\frac{1}{1+\pargamma}}$ and
\mbox{$T_{\mathrm{f}} = \frac{1}{(1+\pargamma)}\cdot
  \frac{\log(\lradius/\noise(\obs, \tol))} {\log(1/\fastcont)}$.}
Without loss of generality, we can assume that
\begin{align}
  \label{eq:unstable_fast_assumption}
  \enorm{\tvar_\obs^\iter-\tstar} >
  \frac{(2-\fastcont)}{(1-\fastcont)} \noiseunstable(\obs, \tol)\quad\text{ for
    all }\quad \iter \in \braces{0, \ldots, T_{\mathrm{f}}-1}.
\end{align}
Otherwise, the claim is immediate. Given the
condition~\eqref{eq:unstable_fast_assumption}, we prove the following
two claims:
\begin{subequations}
\begin{align}
\label{eq:unstable_iterates_disc}
    \tvar_\obs^\iter &\in \annulus (\tstar, \noiseunstable(\obs, \tol), \lradius)
    \quad\text{ for all}\quad \iter \in \braces{0, \ldots, T_{\mathrm{f}}-1}, \\
    \label{eq:unstable_fast_final}
    \quad\text{and}\quad
    \enorm{\tvar_\obs^{T_{\mathrm{f}}}-\tstar} &\leq \frac{(2-\fastcont)}{(1-\fastcont)}\noiseunstable(\obs, \tol).
\end{align}
\end{subequations}
The latter claim~\eqref{eq:unstable_fast_final} completes the proof of 
part~\ref{item:fast} of the theorem.

\paragraph{Proof of claim~\eqref{eq:unstable_iterates_disc}}
With the condition~\eqref{eq:unstable_fast_assumption} in hand, it remains 
to prove that \mbox{$\enorm{\tvar_\obs^\iter-\tstar} \leq \lradius$}. 
The base case of $\iter = 0$ is immediate from the initialization conditions.
For the induction step, assuming $\tvar_\obs^\iter \in \annulus (\tstar, \noiseunstable(\obs, \tol),
\lradius)$, we have
\begin{align}
  \enorm{\tvar_\obs^{\iter+1}-\tstar} =
  \enorm{\nop(\tvar_\obs^{\iter})-\tstar} &\leq
  \enorm{\nop(\tvar_\obs^{\iter})-\pop(\tvar_\obs^{\iter})} +
  \enorm{\pop(\tvar_\obs^{\iter})-\tstar}\notag \\
& \stackrel{(i)}{\leq} \sup \limits_{\tvar\in\annulus(\tstar,
    \noiseunstable(\obs, \tol) , \lradius)}\enorm{\nop(\tvar)-\pop(\tvar)} +
  \fastcont\enorm{\tvar_\obs^{\iter}-\tstar}\notag \\
& \stackrel{(ii)}{\leq} \noise(\obs, \tol) \max
  \braces{\frac{1}{\noiseunstable(\obs, \tol)^\pargamma},\lradius} +
  \fastcont\lradius\label{eq:fast_unstable_induction} \\
& = \frac{\noise(\obs, \tol) }{\noiseunstable(\obs, \tol)^\pargamma} +
  \fastcont\lradius \notag\\
&= \noise(\obs, \tol)^{\frac{1}{1+\pargamma}} + \fastcont\lradius
 \stackrel{(iii)}{\leq} \lradius,\notag
\end{align}
where the inequality~(i) follows from the induction hypothesis that
$\tvar_\obs^\iter \in \annulus (\tstar, \noiseunstable(\obs, \tol),
\lradius)$ and the fact that operator~$\pop$ is
$\fastcont$-contractive in the ball $\ballstar$; inequality~(ii)
follows from the first inequality from
condition~\eqref{eq:fast_unstable_thm_condition} that implies that
$\noiseunstable(\obs, \tol) = \noise(\obs, \tol)^{\frac{1}{1+\pargamma}}\geq
\innerradius$ and then invoking the instability
condition~\eqref{eq:sample_instability} with $\radius =
\noiseunstable(\obs, \tol)$ and $\lradius_2 = \lradius$. Finally,
the last inequality~(iii) follows from the second bound of
the condition~\eqref{eq:fast_unstable_thm_condition}. The inductive step
is thus established.

\paragraph{Proof of claim~\eqref{eq:unstable_fast_final}} 
We observe that
\begin{align}
\enorm{\tvar_\obs^{T_{\mathrm{f}}}-\tstar} & =
\enorm{\nop(\tvar_\obs^{T_{\mathrm{f}}-1})-\tstar} \\ 
&\leq
\enorm{\nop(\tvar_\obs^{T_{\mathrm{f}}-1})-\pop(\tvar_\obs^{T_{\mathrm{f}}-1})} +
\enorm{\pop(\tvar_\obs^{T_{\mathrm{f}}-1})-\tstar} \notag \\
& \stackrel{(i)}{\leq} \sup \limits_{\tvar\in \annulus(\tstar,
  \noiseunstable(\obs, \tol), \lradius)}
\enorm{\nop(\tvar_\obs^{T_{\mathrm{f}}-1})-\pop(\tvar_\obs^{T_{\mathrm{f}}-1})} +
\fastcont\enorm{\tvar_\obs^{T_{\mathrm{f}}-1}-\tstar} \notag \\
& \stackrel{(ii)}{\leq} \noise(\obs, \tol)
\max\braces{\frac{1}{\noiseunstable(\obs, \tol)^\pargamma},\lradius} +
\fastcont\enorm{\tvar_\obs^{T_{\mathrm{f}}-1}}
\definecolor{darkscarlet}{rgb}{0.34, 0.01, 0.1}, \label{eq:fast_unstable_key_recursion}
\end{align}
where inequality~(i) follows from our earlier
claim~\eqref{eq:unstable_iterates_disc} and the $\fastcont$-contractivity
of the operator $\pop$ on the ball $\ballstar$; inequality~(ii) follows 
from an argument similar to the one used to establish the
inequality~\eqref{eq:fast_unstable_induction}.
Finally, recursing equation~\eqref{eq:fast_unstable_key_recursion} $T_{\mathrm{f}}$
times, we obtain that
\begin{align*}
\enorm{\tvar_\obs^{T_{\mathrm{f}}}-\tstar} & {\leq} \noise(\obs, \tol)
\max\braces{\frac{1}{\noiseunstable(\obs, \tol) ^\pargamma},\lradius} \cdot ( 1 +
\fastcont + \ldots + \fastcont^{T_{\mathrm{f}}-1}) +
\fastcont^{T_{\mathrm{f}}}\enorm{\tvar_\obs^{0}-\tstar}\\ &\leq
\frac{\noise(\obs, \tol)}{(1-\fastcont)}
\max\braces{\frac{1}{\noiseunstable(\obs, \tol)^\pargamma},\lradius} +
\fastcont^{T_{\mathrm{f}}}\lradius \\
& \leq \frac{\noiseunstable(\obs, \tol)}{(1-\fastcont)} + \noiseunstable(\obs, \tol)
=\frac{(2-\fastcont)}{(1-\fastcont)}\noiseunstable(\obs, \tol),
\end{align*}
where the last step follows from the upper bound on iteration $T_{\mathrm{f}}$, which in turn
implies that $\fastcont^{T_{\mathrm{f}}}\lradius \leq \noiseunstable(\obs,
\tol)$. The proof is now complete.


\subsubsection{Proof of part~\ref{item:slow}}

The proof for 
Theorem~\ref{theorem:fast_unstab_meta_theorem}\ref{item:slow}
borrows ideas from the proof of
Theorem~\ref{theorem:slow_meta_theorem} as well as the proof of part
\ref{item:fast} of Theorem~\ref{theorem:fast_unstab_meta_theorem}.  We
introduce the following definitions:
\begin{align*}
    T_{\mathrm{s}} \defn [\noise(\obs, \tol)]^{-\frac{1-
    \pargammaMod \parfinal_\star}{1 + \parbeta}}, \quad \text{where} 
   \quad  \parfinal_\star \mydefn
  \frac{\parbeta}{1 + \parbeta - \pargamma \parbeta}. 
\end{align*}
In order to prove the result~\eqref{eq:slow_unstable_thm_statement}, 
we can, without loss of generality, assume that
\begin{align}
  \label{eq:unstable_slow_induction}
  \enorm{\tvar_\obs^\iter-\tstar} > 2[\noise(\obs, \tol)]^{\parfinal_\star}
  \quad\text{ for all }\quad \iter \in \braces{0, \ldots, T_{\mathrm{s}}-1},
\end{align}
and show that $\enorm{\tvar_\obs^{T_{\mathrm{s}}}-\tstar} \leq 2[\noise(\obs, 
\tol)]^{\parfinal_\star}$. We only prove the result for $\tstar=0$ as the more general
case can be derived in a similar fashion.

In order to proceed further, we make use of a result similar to 
Lemma~\ref{lemma:t_perturbation_error} adapted to the unstable case.
Given two positive scalars $\radius_1 < \radius_2$, we define
\begin{align}
        \label{eq:time_function_unstable}
    \tfunUnstable(\radius_1, \radius_2)
    &\defn \frac{\radius_2
  \radius_1^{\pargammaMod}}{ \noise(\obs, \tol)}.
\end{align}

\begin{lemma}
\label{lemma:t_perturbation_error_unstable}
Suppose that the assumptions for part \ref{item:slow} of Theorem~\ref{theorem:fast_unstab_meta_theorem}
hold. Further, suppose that the operator $\nop$ satisfies
$\enorm{\nop^\iter (\tvar)} \geq \radius_1$ for any point $\tvar$ such that
$\enorm{\tvar} \in [\radius_1, \radius_2]$ and for all $\iter \leq
\tfunUnstable(\radius_1, \radius_2)$, where $\innerradius \leq \radius_1
\leq \radius_2 \leq \lradius/2$. Then with probability at least
$1-\tol$, we have
\begin{align}
  \label{eq:t_error_unstable}
  \sup_{ \tvar \in \annulus(\tstar, \radius_1, \radius_2)} \enorm{
    \pop^\iter(\tvar) - \nop^\iter(\tvar)} 
    &\leq \iter
    \cdot \frac{\noise(\obs, \tol)}{\radius_1^{\pargammaMod}} \quad \text{
    for all }\quad \iter \leq \tfunUnstable(\radius_1, \radius_2).
\end{align}
\end{lemma}

\noindent See
Appendix~\ref{sub:proof_of_lemma_lemma:t_perturbation_error_unstable}
for its proof.

We are now ready for the main argument. We have
\begin{align}
  \enorm{\tvar_n^{\iter}} = \enorm{\nop^{\iter}(\tvarn^0)} & \leq
  \enorm{\pop^{\iter} (\tvarn^0)} + \enorm{\pop^{\iter}(\tvarn^0) -
    \nop^{\iter} (\tvarn^0)} \nonumber \\
& \stackrel{(i)}{\leq} \frac{1}{\iter^{ \parbeta}} +
  \enorm{\pop^{\iter}(\tvarn^0) - \nop^{\iter}(\tvarn^0) }
  \label{eq:first_half_induction_intermediate_unstable} \\
  & \stackrel{(ii)}{\leq} \frac{1} {\iter^{ \parbeta}} + \iter \cdot 
  \frac{\noise(\obs, \tol)}{[\noise(\obs, \tol)]^{\parfinal_\star \pargammaMod}},
        \quad \text{for all} \quad \iter \leq
        \tfunUnstable([\noise(\obs, \tol)]^{\parfinal_\star}, \rho),
    \label{eq:first_half_induction_unstable}
\end{align}
with probability at least $1-\tol$.
Here, inequality~(i) follows from the
\slow-convergence condition~\eqref{eq:slow_convergence} of 
the operator $\pop$
along with the assumptions that $\tstar =0$ and $\enorm{\tvar_n^{0}} \leq
\rho$;
inequality~(ii) follows by applying
Lemma~\ref{lemma:t_perturbation_error_unstable} with
$\radius_1 = [\noise(\obs, \tol)]^{\parfinal_\star}$ and $\radius_2 = \rho$
in light of the condition~\eqref{eq:unstable_slow_induction}.
In the final bound~\eqref{eq:first_half_induction_unstable},
the first term decreases with iteration $\iter$ while the second term
increases with $\iter$. In order to trade off the two terms, we plug 
in \mbox{$\iter = T_{\mathrm{s}} \stackrel{(\dagger)}{\leq} \tfunUnstable
([\noise
(\obs,
\tol)]^{\parfinal_\star}, \rho)$} (where the inequality~$(\dagger)$ holds
due
to the second bound in assumption~\eqref{eq:slow_unstable_thm_condition}),
and perform some algebra to obtain that
\begin{align*}
 \enorm{\tvar_n^{T_{\mathrm{s}}}} \leq \frac{1}{T_{\mathrm{s}}^{ \parbeta}} + T_{\mathrm{s}}
 \frac{\noise(\obs, \tol)}{[\noise(\obs, \tol)]^{\parfinal_\star \pargammaMod}} \leq 
 2[\noise(\obs, \tol)]^{\parfinal_\star},
\end{align*}
which yields the claim.

\section{Tightness of general results} 
\label{sec:lower_bounds}

In this appendix, we construct a simple class of problems to
demonstrate that the guarantees Theorems~\ref{theorem:slow_meta_theorem} and~\ref{theorem:fast_unstab_meta_theorem} in this paper are
unimprovable in general.


\subsection{Constructing the family of operators}
\label{par:gradient_descent_and_newton_s_method_}
We establish our lower bounds by considering the following pairs of optimization problems:
\begin{align}
  \label{eq:pop_polynomial_optimization}
&\min_{\tvar \in \Rspace^\usedim} \polyfunc ( \tvar), \ \ \text{where}
\ \polyfunc ( \tvar) := \frac{\enorm{ \tvar}^{p}}{ p}, \quad\text{and} \\
\label{eq:polynomial_optimization}
&\min_{\tvar \in \Rspace^\usedim} \polyfunc_{n} ( \tvar), \ \ \text{where } \polyfunc_{n} ( \tvar) := \polyfunc(\theta) -
  \noise_n \frac{\enorm{ \tvar}^{q}}{q},
\end{align}
where $p, q$ are positive reals satisfying $q \geq2$, and $p > q+1$, and the scalar $\noise_n$ is a perturbation term.  The perturbation $\noise_n$ is any non-increasing function in $n$, that decays to zero as the sample size $\obs$ increases, so that the problem~\eqref{eq:polynomial_optimization} can be seen as a noisy ``finite-sample instantiation'' of the ``population-level'' problem~\eqref{eq:pop_polynomial_optimization}.

\newcommand{\epsstat}{\epsilon^{\mathrm{stat}}_n}
To study the tightness of our general results, we study the guarantees for three different algorithms: (a) gradient descent method, (b) Newton's method, and (c) cubic-regularized Newton's
method (for $d = 1$). We note that for the population-level updates, there is a unique global optimal $\tstar = 0$, and for the sample-level objective, the
global minima $\tvar_{n}^{*}$ satisfies $\epsstat
\defn \enorm{ \tvar_{n}^{*}} = \noise_n^{\frac{1}{p - q}}$. 

For our lower bounds, we analyze the behavior of three different algorithms: (a) gradient
descent method, (b) Newton's method, and (c) cubic-regularized Newton's
method (for $d = 1$), with the population-level operators $\sampolygd$, $\sampolynm$, and $\sampolycnm$ defined as follows:
\begin{subequations}
  \begin{align}
    \label{eq:pop_gd_poly}    
    \polygd( \tvar) & = \tvar - \learnrate \nabla \polyfunc ( \tvar) 
    = \tvar \parenth{ 1 - \learnrate \enorm{\tvar}^{q - 2}}, \\
    \label{eq:pop_nm_poly}                 
    \polynm( \tvar) & = \tvar - \brackets{ \nabla^2
      \polyfunc ( \tvar)}^{-1} \nabla \polyfunc ( \tvar) =
    \parenth{1 - \frac{1}{p - 1}} \tvar, \quad \text{and}
    \\
\label{eq:pop_cnm_poly}                
\polycnm (\tvar) & = \mathop{ \arg \min}_{y \in \Rspace} \biggr\{
\nabla \polyfunc (\tvar) (y - \tvar) + \frac{1}{2} \nabla^2 \polyfunc
(\tvar) (y - \tvar)^2 + c_p \abss{y - \tvar}^3
\biggr\},
  \end{align}
  where $c_p := \frac16(p-1)(p-2)$, and $\learnrate > 0$ denotes the step-size of gradient descent
algorithm. 
\end{subequations}
The corresponding sample-level updates are generated by the operators
$\sampolygd$, $\sampolynm$, and $\sampolycnm$, as follows:
\begin{subequations}
  \begin{align}
    \label{eq:grad_poly}
    \sampolygd( \tvar) &\!=\! \tvar \!-\! \learnrate \nabla \polyfunc_{n}(
    \tvar) = \tvar - \learnrate \parenth{ \enorm{ \tvar}^{p - 2} -
      \noise \enorm{ \tvar}^{q - 2}} \tvar, \\
\label{eq:newton_poly}     
    \sampolynm( \tvar) &= \tvar - \brackets{ \nabla^2 \polyfunc_{n}(
      \tvar)}^{-1} \nabla \polyfunc_{n}( \tvar)
      = \frac{ (p - 2)
      \enorm{ \tvar}^{p - 2} - (q - 2) \noise \enorm{ \tvar}^{q -
        2}}{(p - 1) \enorm{ \tvar}^{p - 2} - \noise (q - 1) \enorm{
        \tvar}^{q - 2}} \tvar, 
        \\
    \sampolycnm(\tvar) &= \mathop{ \arg \min}_{y \in \Rspace}
    \braces{\nabla \polyfunc_{n} (\tvar) (y \!- \!\tvar) + \frac{1}{2} \nabla^2
    \polyfunc_{n} (\tvar) (y \!-\! \tvar)^2 
    \!+\! c_p
    \abss{y \!-\! \tvar}^3}.
  \end{align}
\end{subequations}

Standard algebra with the update 
equations~\eqref{eq:pop_gd_poly}-\eqref{eq:pop_cnm_poly}
yields the following properties with the population-level operators:
\begin{enumerate}[label=($\widehat{\mathrm P}$\arabic*)]
\item the operator $\polygd$ is $\slownotag
(\frac{1}{p - 2})$-convergent on the ball $\ballnotag (\tstar, 1)$ for small enough $\learnrate > 0$,
\item the operator $\polynm$ is $\fastnotag (\frac{p-2}{p -
      1})$-convergent towards $\tstar = 0$, and
\item the operator $\polycnm$ is
\mbox{$\slownotag (\frac{2}{p-3})$}-convergent on the ball $\ballnotag (\tstar, 1)$.
\end{enumerate}
Moving to the (in)-stability of sample-level operators, we can verify that:
\begin{enumerate}[label=($\widehat{\mathrm S}$\arabic*)]
\item the operator $\sampolygd$ is $\stabilitynotag(q - 1)$-stable
  over the Euclidean ball $\ballnotag (\tstar, 1)$;
\item the operator $\sampolynm$ is $\instabilitynotag(- p + q +
  1)$-unstable over the annulus $\annulus(\tstar, c_{1} \newdndelta,
  1)$, and
\item the operator $\sampolycnm$ is $\instabilitynotag (- \frac{p +
  1}{2} + q)$-unstable over the annulus $\annulus ( \tstar, c_{2}
  \newdndelta, 1)$ ($d=1$)
\end{enumerate}
with respect to the corresponding population-level operators, and the
noise function $\noise_n$.


\subsection{Lower bounds showing sharpness}
\label{subsec:guarantee_cookup_example}

\newcommand{\tnidx}{T_n}

In this section, we demonstrate the our general upper bounds on
statistical accuracy and the iteration count, when specialized to the
set-up above, are unimprovable.  More precisely, the following result
applies to the gradient descent updates~\eqref{eq:grad_poly} with step
size $\learnrate \in \big(0, \tfrac{1}{2} \big]$, along with the cubic
regularized and standard Newton updates.

\begin{proposition} 
\label{prop:lower_poly}
Let $p> q+1$ and $q \geq 2$, and define $\epsstat:=
\noise_n^{\frac{1}{p - q}}$, then for the
set-up~\eqref{eq:polynomial_optimization}, given an initialization
$\tvar^0$ with $\enorm{\tvar^0} = 1$, we have
    \begin{align}
  \enorm{ (\sampolygd)^{t}(\tvar^0) - \tstar } &
      \begin{cases}
      \leq 2 \epsstat\qtext{for all} t \geq c_1 \noise_n^{-\frac{p - 2}{p-q}},\\
      \geq \epsstat\ \ \qtext{for all} t \geq 1,\\
      \geq 2 \epsstat \qtext{for all} t\leq c_1' \noise_n^{-\frac{p - 2}{p-q}},
      \label{eq:gd_poly_bound}
      \end{cases}
      \\
      \enorm{
   (\sampolynm)^{t}(\tvar^0) - \tstar } &
   \begin{cases}
   \leq 2 \epsstat\qtext{for all} t \geq c_2 \log ( \noise_n^{ -1})),\\
      \geq \epsstat \ \ \qtext{for all} t \geq 1,\\
      \geq 2 \epsstat \qtext{for all} t\leq c_2' \log ( \noise_n^{ -1})),
   \end{cases} \quad\text{and}
   \label{eq:nm_poly_bound}
        \\
      \enorm{ (\sampolycnm)^{t}(\tvar^0) - \tstar } &
      \begin{cases}
   \leq 2 \epsstat\qtext{for all} t \geq
      c_3\noise_n^{-\frac{p - 3}{p - 1}},\\
      \geq \epsstat \ \ \qtext{for all} t \geq 1,
   \end{cases}
     \label{eq:cnm_poly_bound}
    \end{align}
    where $\tstar = 0$ denotes the fixed point of the operators $\polygd,\polynm$, and $\polycnm$, and $c_1>c_1', c_2>c_2',$ and $c_3$ denote universal constants depending on $p, q$ and independent of $n$.
\end{proposition}

It is worth understanding how Proposition~\ref{prop:lower_poly}
establishes the tightness of the general upper bounds given in
Theorems~\ref{theorem:slow_meta_theorem}
and~\ref{theorem:fast_unstab_meta_theorem}.  Note that the properties
$(\widehat{\mathrm P}1)-(\widehat{\mathrm P}3)$ and $(\widehat{\mathrm
  S}1)-(\widehat{\mathrm S}3)$, in conjunction with our general
results in Theorems~\ref{theorem:slow_meta_theorem}
and~\ref{theorem:fast_unstab_meta_theorem}, provide an upper bound on
the statistical error given sufficiently many iterations as summarized
in bounds~\eqref{eq:gd_poly_bound}, \eqref{eq:nm_poly_bound} and
\eqref{eq:cnm_poly_bound}, e.g., for the GD iterates, substituting
$\parbeta=\frac1{p-2}$ and $\pargamma = q-1$ in
Theorem~\ref{theorem:slow_meta_theorem}, we conclude that
\begin{align*}
  \enorm{ (\sampolygd)^{t}(\tvar^0) - \tstar } \leq
      2\noise_n^{\frac{1}{p - q}} = 2\epsstat \quad\text{for all} \quad t \geq
      c_1\noise_n^{-\frac{p - 2}{p-q}}.
\end{align*} 
Furthermore, Proposition~\ref{prop:lower_poly} guarantees that, up to a
the constant pre-factor $2$, this statistical error is the best
possible since
\begin{align*}
  \enorm{ (\sampolygd)^{t}(\tvar^0) - \tstar } \geq \epsstat
  \qtext{for all} t \geq 1.
\end{align*}
Finally, the proposition also asserts that the GD updates take at
least order $\noise_n^{-\frac{p - 2}{q - 2}}$ iterations to converge
by the following additional bounds, as we also have the following
bound from the display~\eqref{eq:gd_poly_bound}:
\begin{align*}
  \enorm{ (\sampolygd)^{t}(\tvar^0) - \tstar } \geq 2\epsstat
  \quad\text{for all} \quad t = c_2'\noise_n^{-\frac{p - 2}{p-q}}.
\end{align*}

A similar tightness of the statistical and computational guarantee can
be argued for fast unstable methods stated in
Theorem~\ref{theorem:fast_unstab_meta_theorem}\ref{item:fast} via the
guarantee~\eqref{eq:nm_poly_bound} for the Newton's method.  Finally,
for slow unstable operators, we establish the tightness for the
statistical error guarantee of
Theorem~\ref{theorem:fast_unstab_meta_theorem}\ref{item:slow} via the
bound~\eqref{eq:cnm_poly_bound} for the CNM algorithm. (Showing the
tightness of iteration complexity for this case requires fairly
involved technical analysis, and is left for future work.)  In a
nutshell, Proposition~\ref{prop:lower_poly} shows that the upper bound
on the final statistical errors, and the lower bound on the number of
iterations needed to obtain that final estimate, as stated in
Theorems~\ref{theorem:slow_meta_theorem}
and~\ref{theorem:fast_unstab_meta_theorem} are tight for the class of
problems~\eqref{eq:polynomial_optimization}.

\subsection{Proof of Proposition~\ref{prop:lower_poly}}

As noted earlier, the upper bounds on the statistical error, and the
corresponding lower bound on the number of iterations follow directly
by substituting appropriate $\parbeta$ and $\pargamma$ parameters from
the properties listed above in
Theorems~\ref{theorem:slow_meta_theorem} and
\ref{theorem:fast_unstab_meta_theorem}. Since the arguments are very
similar to those in Appendix~\ref{sec:proofs_of_corollaries}, we omit
a detailed derivation.

In order to see that the statistical error cannot decrease any
further, we note that in our example the iterates from gradient
descent and (cubic-regularized) Newton's methods always converge to
the global minima $\tvar_{n}^{*}$ of $\polyfunc_{n}$. Thus, we also
have
\begin{align*}
\enorm{\sampolygd( \tvar)} \geq \epsstat, \ \ \enorm{\sampolynm(
  \tvar)} \geq \epsstat, \ \text{and} \ \abss{\sampolycnm( \tvar)}
\geq \epsstat
\end{align*}
for all $\enorm{ \tvar} \geq \epsstat$. Consequently, we conclude that
the error for all iterations are lower bounded by $\epsstat$. \\

Next, we establish the lower bounds on the number of iterations to
converge to within $2\epsstat$ for Gradient descent and Newton's
method.  Introducing the shorthand $\noise \defn \noise_n$, and
rearranging terms in equations~\eqref{eq:grad_poly}
and~\eqref{eq:newton_poly}, we find that
\begin{align}
\label{eqn:GD-eqn}
\GD_n(\theta) &= \left(1 - \stepsize \enorm{\theta}^{p - 2} +
\stepsize \noise \enorm{\theta}^{q -2} \right)\theta, \\
\label{eqn:NM-rate}
\NM_n(\theta) & = \left(1 - \frac{\enorm{\theta}^{p - 2} -
  \enorm{\theta}^{q-2}\noise}{(p-1)\enorm{\theta}^{p - 2} - (q -
  1)\enorm{\theta}^{q-2}\noise} \right) \theta.
\end{align}

\paragraph{Proof for gradient descent iterates:} Recursing the
update~\eqref{eqn:GD-eqn}, we find that
\begin{align}
\label{eqn:GD-full-recursion}
    \theta^{t + T} = \theta^t \cdot \prod_{j = 1}^{T} \left(1 -
    \stepsize \enorm{\theta^{t + j}}^{p - 2} + \stepsize \noise
    \enorm{\theta^{t + j}}^{q -2} \right).
\end{align}
Note that it suffices to show that with $\enorm{\theta^t} = 2 \Delta
:= 4 \epsstat = 4 \noise^{\frac{1}{p - q}}$, the smallest $T_\Delta$
such that $\enorm{\theta^{T_\Delta + t}} \leq $ satisfies $T_\Delta =
\Omega(\Delta^{2-p})$.

Since the sequence $\{ \enorm{\theta^{t + j}} \}_{j = 1}^{T_\Delta}$
is a decreasing sequence, we find that
\begin{align*}
    \Delta \leq \enorm{\theta^{t + j}} \leq 2\Delta \qquad \text{for
      all} \;\; j = 1, \ldots T_\Delta.
\end{align*}
Using $\Delta \defn 2 \cdot \noise^{\frac{1}{p - q}}$ and the
update~\eqref{eqn:GD-full-recursion}, we have
\begin{align*}
  \enorm{\theta^{t + T_\Delta +1}} \geq \left(1 - c \eta \Delta^{p -
    2}\right)^{T_\Delta} \enorm{\theta^t} = 2 \Delta \cdot \left(1 - c
  \eta \Delta^2\right)^{T_\Delta}.
\end{align*}
where $c = 2^{2p-4}-2^{q-p}>0$ under the assumptions $p>q+1$ and $q
\geq2$.
In order to ensure that $\Vert\theta^{t + T_\Delta}\Vert \leq \Delta$,
we need to have
\begin{align*}
  \left(1 - c \eta \Delta^{p - 2}\right)^{T_\Delta} \leq \frac{1}{2}.
\end{align*}
Rearranging the last equation yields $T_\Delta \geq
\frac{c'}{\Delta^{p - 2}} \geq c'\noise^{-\frac{p -2}{p -q}}$, where
$c'$ is a universal constant which depends only on the pair $(p,
q)$. This completes the proof. \\


\paragraph{Proof for Newton's method iterates:} Following an
argument similar to the last paragraph and using $\enorm{\theta_t} = 2
\Delta \geq 2 \cdot \noise^{\frac{1}{p - q}}$, we find that
\begin{align*}
  \enorm{\theta^{t + T_\Delta}} \geq \left(1 - \frac{1}{p-q}
  \right)^{T_\Delta} \enorm{\theta^t} = 2 \Delta \cdot \left(1 -
  \frac{1}{p-q} \right)^{T_\Delta}.
\end{align*}
Recalling that $p - q \geq 2$, we have that $T_\Delta \geq \frac{\log
  2}{\log\left( \frac{p - q}{p-q - 1} \right)}$. Consequently, in
order to achieve an accuracy of $\frac{1}{\noise^{p - q}}$, we need at
least $\frac{\log 2}{\log\left( \frac{p - q}{p-q-1} \right)} \cdot
\log(\noise^{-(p - q)}) = c' \cdot \log(1/\noise)$ steps. Here, the
universal constant $c'$ only depends on $(p,q)$. This completes the
proof of the sharpness of the Newton's method.


\subsection{Undesirable behavior of unstable operators}
\label{AppCounterExample}

\begin{figure}[h]
  \begin{tabular}{cc}   
    \widgraph{0.45\textwidth}{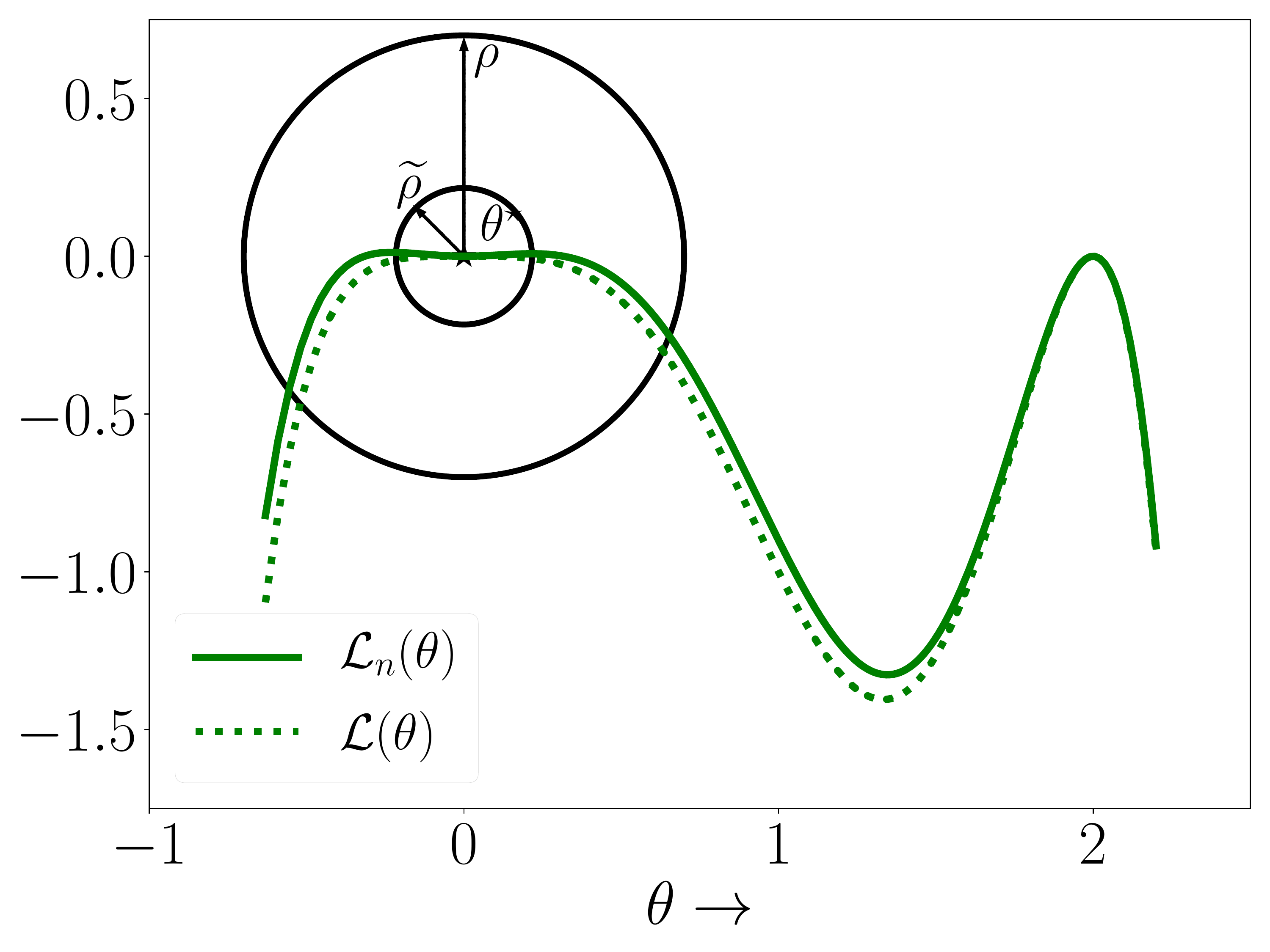}
    &
    \widgraph{0.45\textwidth}{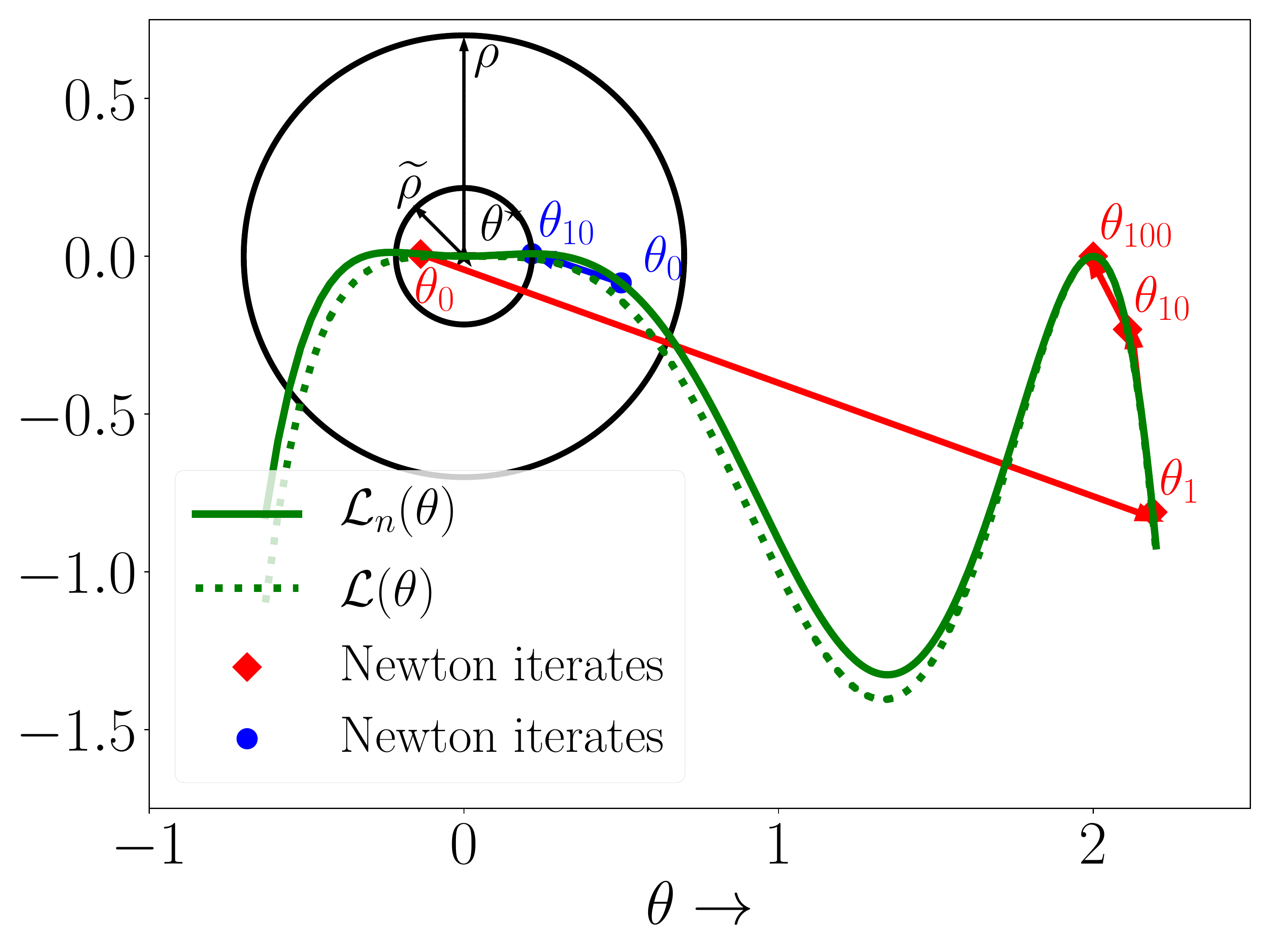}
    \\
    (a) & (b)
  \end{tabular}
  \caption{Instability of Newton's method for the example
    discussed above (figure best viewed in color). When the
    algorithm is initialized too close to $\tstar$ (red
    diamonds), the instability of Newton's method forces the
    iterates to jump too far away from $\tstar$ and converge to
    another fixed point. On the other hand, if the initial point
    is initialized in the annulus $\Ann(\tstar, \innerradius,
    \lradius)$, the Newton iterates (blue circles), do not leave
    this annulus and converge monotonically to a small
    neighborhood of~$\tstar$.  }
  \label{fig:counter_example}
\end{figure}

In this appendix, we prove that the minimum over all iterates $k \in
\{1, 2, \ldots, t \}$ in
Theorem~\ref{theorem:fast_unstab_meta_theorem} is necessary.  In
particular, we consider the following example
\begin{align*}
\mathcal{L}(\tvar) = -\tvar^4 (\tvar-2)^2 \quad \mbox{and} \quad
\mathcal{L}_\obs(\tvar) = -\parenth{\tvar^4 -
  \frac{\tvar^2}{\sqrt{\obs}}} (\tvar-2)^2.
\end{align*}
We let $\pop$ and $\nop$ denote the operators corresponding to the
Newton's method as applied to the functions $\mathcal{L}$ and
$\mathcal{L}_\obs$, respectively (Consequently, the operator $\pop$
has three fixed points). Following some simple algebra, it can be
verified there are universal constants $(c_1, c_2)$ such that that the
operators $\pop$ and $\nop$ defined above satisfy the conditions of
Theorem~\ref{theorem:fast_unstab_meta_theorem} (a) with $\tstar=0$ for
some $\fastcont<1$, $\pargamma=-1, \noise(\obs, \tol) =
\obs^{-\frac{1}{2}}, \innerradius= {c_1}\obs^{-\frac{1}{4}}$ and
$\lradius=c_2$. In panel (a) of Figure~\ref{fig:counter_example}, we
plot the two functions $\mathcal{L}$ and $\mathcal{L}_\obs$ and
illustrate the radii $\innerradius, \lradius$ (for a fixed $n$).  Some
additional algebra shows that there exists $\tvar^0_\obs \in \ball
(\tstar, \innerradius)$ such that the iterates corresponding to the
sequence $\tvar_\obs^ {\iter+1} = \nop(\tvar_\obs^\iter)$ satisfy
$\enorm{\tvar_\obs^{\iter}-\tstar} \geq 1 \gg \obs^{-\frac{1} {4}}$
for all iterations $\iter = 1, 2, \ldots$. See, in particular, the red
(diamond) iterates in panel (b) of Figure~\ref{fig:counter_example}
which are generated with a starting point $\tvar_n^0 =
c_3n^{-\frac14}$ (which is below the controlled instability threshold
$\innerradius$). Clearly, we see that the first iterate produced by
Newton's method escapes the local basin of attraction and the
subsequent iterates converge to a very different fixed point of the
function $\mathcal{L}_\obs$. On the other hand, when the Newton's
method is initialized in the annulus $\annulus(\tstar, \innerradius,
\lradius)$, the sequence $\tvar_n^\iter$ (blue circles) converges
quickly to the vicinity of $\tstar$ as guaranteed by
Theorem~\ref{theorem:fast_unstab_meta_theorem}.  Furthermore, the
iterates do not escape this local neighborhood.
Via this simple example, we have demonstrated that if no further
regularity assumptions are made, then starting an unstable algorithm
from a point that is too close to $\tstar$, the subsequent iterates
can be quite far from the true parameter.


\section{Proofs of auxiliary results}
\label{label:tehcnical_lemmas}

In this appendix, we collect the proofs of
Lemmas~\ref{lemma:t_perturbation_error}
and~\ref{lemma:t_perturbation_error_unstable} that are central to the
proofs of our main theorems.


\subsection{Proof of Lemma~\ref{lemma:t_perturbation_error}} 
\label{ssub:proof_of_lemma_lemma:t_perturbation_error}

We fix a radius $\radius \in \radii$.  Our proof is based on the
following auxiliary claim: conditioned on the event $\event$ from
equation~\eqref{eqn:event_definition}, we have
\begin{align}
\label{eq:bound_on_fn}
\sup_{\tvar\in\ball(\tstar, \radius)}\enorm{\nop^{\iter} (\tvar)} \leq
2 \radius \quad \text{for all}\quad \iter \leq \tfun(\radius)
=\frac{\radius^{1-\pargamma}}{2^\pargamma \unicontwo\mydndelta}.
\end{align}
Taking this claim as given for the moment, we now establish the
bound~\eqref{eq:t_error} claimed in the lemma.  We do so via induction
on the iteration $\iter \in \{0, 1, \ldots, \tfun(\radius)\}$.  Note
that the base-case $\iter = 0$ holds trivially, since
$\enorm{\pop^{0}(\tvar) - \nop^{0}(\tvar)} = \enorm{\tvar-\tvar} = 0$.
Given the induction hypothesis for $\iter$, we establish the claim for
$\iter'=\iter+1$. For any $\tvar \in \ball (\tstar, \radius)$, we have
\begin{align}
  \enorm{\pop^{\iter'}(\tvar) - \nop^{\iter'}(\tvar)} &=
  \enorm{\pop^{\iter + 1}(\tvar) - \nop^{\iter +
      1}(\tvar)} \label{eq:induction_t_error_first_step}\\
  & \leq \enorm{\pop( \pop^{\iter}(\tvar)) - \pop
    (\nop^{\iter}(\tvar))} + \enorm{\pop (\nop^{\iter}(\tvar)) -
    \nop(\nop^{\iter}(\tvar))} \notag\\ & \stackrel{(i)}{\leq}
  \enorm{\pop^{\iter}(\tvar) -\nop^{\iter}(\tvar)} +
  \sup_{\widetilde{\tvar} \in \ball(\tstar, 2\radius)}\enorm{
    \pop(\widetilde{\tvar}) - \nop (\widetilde{\tvar})}\notag\\ &
  \stackrel{(ii)}{\leq}\sup_{\tvar\in\ball(\tstar, \radius)} \enorm{
    \pop^\iter(\tvar) - \nop^\iter(\tvar)} + \unicontwo
  (2\radius)^\pargamma \mydndelta\notag\\
  & \stackrel{(iii)}{\leq} \unicontwo (2\radius)^\pargamma
  \mydndelta\iter + \unicontwo (2\radius)^\pargamma \mydndelta \notag \\
  & =(\iter+1) \unicontwo (2\radius)^\pargamma \mydndelta.  \notag
\end{align}
In the above sequence of inequalities, we have made use of the
following facts.  In step~(i), we have used the
$1$-Lipschitzness~\eqref{eq:lipschitz} of the operator $\pop$ for the
first term and the bound~\eqref{eq:bound_on_fn} on
$\nop^\iter(\theta)$ for the second term. In order to establish
step~(ii), we have used the fact that $\tvar \in
\ball(\tstar,\radius)$ for the first term, while for the second term
we have invoked the definition of the event~$\event$ in
equation~\eqref{eqn:event_definition} with radius $2\radius$ (note
that $2\radii \subset \radii'$ and the event $\event$ is defined for
all $\radius' \in \radii'$).  Finally step~(iii) follows directly from
the induction hypothesis.  Noting that the
bound~\eqref{eq:bound_on_fn} holds for any $\iter \leq \tfun(\radius)$
and taking supremum over $\tvar \in \ball(\tstar, \radius)$ on the LHS
of equation~\eqref{eq:induction_t_error_first_step}, we obtain the
desired proof of the inductive step.

\subsubsection{Proof of claim~\eqref{eq:bound_on_fn}}
\label{par:proof_of_claim_eq:bound_on_fn_}

We establish the claim~\eqref{eq:bound_on_fn} by proving the following
stronger result: For any fixed $\radius \in \radii$, and any $\tvar
\in \ball(\tstar, \radius)$, we have
\begin{align}
  \label{eq:more_general_bound_on_fn}
  \enorm{\nop^{\iter} (\tvar)} \leq \radius + \unicontwo
  (2\radius)^\pargamma \mydndelta \cdot \iter \quad \mbox{for all
    iterations $\iter = 0, 1, \ldots, \tfun(\radius)$.}
\end{align}
We note that the claim~\eqref{eq:bound_on_fn} is a direct application
of this result along with the definition
\mbox{$\tfun(\radius)=\frac{\radius^{1-\pargamma}}
  {2^\pargamma\unicontwo\mydndelta}$}.  We now use an induction
argument on the iteration $\iter$ (similar to the ones used in the
paragraph above) to establish the
claim~\eqref{eq:more_general_bound_on_fn}.  The base-case $\iter=0$
holds trivially.  Let us assume that $\enorm{\nop^{\iter} (\tvar)}
\leq \radius + \unicontwo (2\radius)^\pargamma \mydndelta \cdot \iter
$ and establish the claim~\eqref{eq:more_general_bound_on_fn} for
$\iter'=\iter+1$.  Note that since $\iter \leq \tfun(\radius)$, this
assumption trivially yields that $\enorm{\nop^{\iter} (\tvar)}\leq
2\radius$.  We have
\begin{align*}
  \enorm{\nop^{\iter+1}(\tvar)} &\leq
  \enorm{\pop(\nop^{\iter}(\tvar))}
  +\enorm{\pop(\nop^{\iter}(\tvar))-\nop(\nop^{\iter}(\tvar))}
  \\ &\stackrel{(i)}{\leq} \enorm{\nop^{\iter}(\tvar)} +
  \sup_{\widetilde{\tvar} \in \ball(\tstar, 2\radius)}\enorm{
    \pop(\widetilde{\tvar}) - \nop
    (\widetilde{\tvar})}\\ &\stackrel{(ii)}{\leq} (\radius +
  \unicontwo (2\radius)^\pargamma \mydndelta \cdot \iter) +
  \unicontwo (2\radius)^\pargamma \mydndelta\\ &= \radius +
  \unicontwo (2\radius)^\pargamma \mydndelta (\iter+1),
\end{align*}
where in step~(i), we have used the $1$-Lipschitzness~\eqref{eq:lipschitz}
of the operator $\pop$ for the first term and the observation that
$\enorm{\nop^{\iter} (\tvar)}\leq 2\radius$ for the second term.  On
the other hand, in step~(ii), we have used the induction hypothesis to
bound the first term, and invoked the definition of the event~$\event$
in equation~\eqref{eqn:event_definition} with radius $2\radius$ to
bound the second term.  Taking supremum over $\tvar \in \ball(\tstar,
\radius)$ completes the proof.

\subsubsection{Proof of claim~\eqref{eq:time_bounds_validity2}}

Combining the relation $\seqalpha{\ind} =
\parfinal_\star(1-\parnuone^\ind)$ with the two inequalities in
equation~\eqref{eq:time_bounds_validity2}, we find that it suffices to
prove the following two bounds:
\begin{align}
\label{eq:simpler_time_bounds_validity}
\mydndelta^{-\frac{\parbeta\parnuone^\ind}{1+\parbeta}} \geq
(2^\pargamma\unicontwo)^{\frac{\parbeta}{1+\parbeta}}
\quad\text{and}\quad
\mydndelta^{-\frac{\parbeta\parnuone^{\ind+1}}{1+\parbeta}} \geq
(2^\pargamma\unicontwo)^{\frac{\parbeta}{1+\parbeta}}
(\constfn)^{-\frac{\parbeta}{\parfinal_\star(1+\parbeta)}}.
\end{align}
Observe that $\seqalpha{\ind}\leq \parfinal_\star-\smallthreshold/4$;
consequently, we find that $1/\parnuone^\ell\leq
4\parfinal_\star/\smallthreshold$ for all $\ind \leq \imax$. Finally,
invoking assumption~\eqref{eq:sample_size} we find that
\begin{align}
\label{eq:small_constant_c}
\specialeps \leq
\frac{1}{(2^\pargamma\unicontwo)^{\frac{4\parfinal_\star}{\smallthreshold}}
  \cdot \max\braces{1, (\constfn)^{\frac{4}{\smallthreshold}}}}.
\end{align}
The rest of the proof follows by noting that the upper
bound~\eqref{eq:small_constant_c} implies the bounds in
equation~\eqref{eq:simpler_time_bounds_validity}.


\subsection{Proof of Lemma~\ref{lemma:t_perturbation_error_unstable}} 
\label{sub:proof_of_lemma_lemma:t_perturbation_error_unstable}

Fix an arbitrary pair of radii $\radius_1, \radius_2 \in \radii$.  Our
proof is based on the following intermediate claim
\begin{align}
  \|\nop^t(\tvar)\| \leq 2 \radius_2 \quad\text{ for all } \iter
  \leq \tfunUnstable(\radius_1, \radius_2).
  \label{eq:uniform_ub_unstable}
\end{align}
We prove this claim at the end of this appendix.  Assuming that this claim
is given at the moment, we now establish the
bound~\eqref{eq:t_error_unstable} claimed in the lemma.  We do so by
using induction on the iteration $\iter \in \{0, 1, \ldots,
\tfunUnstable(\radius_1, \radius_2)\}$ where we note that the base-case $\iter = 0$ holds trivially, since $\enorm{\pop^{0}(\tvar)
  - \nop^{0}(\tvar)} = \enorm{\tvar-\tvar} = 0$.  Turning to the
induction step (with $\iter' = \iter+1$), for any $\tvar$ with
$\|\tvar\| \in [\radius_1, \radius_2]$, we have
\begin{align}
  \label{eq:induction_t_error_first_step}  
  \enorm{\pop^{\iter'}(\tvar) - \nop^{\iter'}(\tvar)} & =
  \enorm{\pop^{\iter + 1}(\tvar) - \nop^{\iter + 1}(\tvar)} \\
& \leq \enorm{\pop( \pop^{\iter}(\tvar)) - \pop (\nop^{\iter}(\tvar))}
  + \enorm{\pop (\nop^{\iter}(\tvar)) - \nop(\nop^{\iter}(\tvar))}
  \notag \\
& \stackrel{(i)}{\leq} \enorm{\pop^{\iter}(\tvar)
    -\nop^{\iter}(\tvar)} + \sup_{ \radius_1 \leq
    \|\widetilde{\tvar}\| \leq 2\radius_2 }\enorm{
    \pop(\widetilde{\tvar}) - \nop (\widetilde{\tvar})}\notag\\ &
  \stackrel{(ii)}{\leq}\sup_{ \radius_1 \leq \|\tvar\| \leq 2
    \radius_2 } \enorm{ \pop^{\iter}(\tvar) - \nop^{\iter}(\tvar)} +
  \frac{\mydndelta}{\radius_1 ^{\pargammaMod}} \notag \\
& \stackrel{(iii)}{\leq} \iter \frac{\mydndelta}{\radius_1
    ^{\pargammaMod}} + \frac{\mydndelta}{\radius_1 ^{\pargammaMod}} =
  (\iter + 1) \cdot \frac{\mydndelta}{\radius_1 ^{\pargammaMod}}. \nonumber
\end{align}
In step~(i), we have used the $1$-Lipschitzness~\eqref{eq:lipschitz} of the
operator $\pop$ for the first term and the upper
bound~\eqref{eq:bound_on_fn} on $\nop^\iter(\theta)$ for the second
term.  In step (ii), the upper bound for the first term follows from
the sequence of inequalities
\begin{align*}
  \innerradius \leq \radius_1 \leq \|\tvar\| \leq \radius_2 \leq
  2\radius_2 \leq \lradius,
\end{align*}
whereas for the second term we have invoked the bound
$\|\widetilde{\tvar}\| \defn \nop^{\iter'}(\tvar) \leq 2
\radius_2$~\eqref{eq:uniform_ub_unstable} and applied the instability
condition~\eqref{eq:sample_instability}. Finally, step (iii) follows
from a direct application of the induction hypothesis.  Note that
the bound~\eqref{eq:bound_on_fn} holds for any $\iter \leq
\tfun(\radius)$. By taking supremum over $\tvar \in \ball(\tstar,
\radius)$ on the LHS of
equation~\eqref{eq:induction_t_error_first_step}, we obtain the
desired proof of the inductive step.

\subsubsection{Proof of bound~\eqref{eq:uniform_ub_unstable}}

We use an inductive argument to show that
\begin{align}
 \|\nop^t(\tvar)\| \leq t \cdot \frac{\mydndelta}{\radius_1
   ^{\pargammaMod}} + \radius_2 \quad \text{for all} \quad 1 \leq t
 \leq \tfunUnstable(\radius_1, \radius_2),
  \label{eqn:induction_ub}
\end{align}
which immediately implies the claim~\eqref{eq:uniform_ub_unstable}
once we plug in the definition of
$\tfunUnstable$~\eqref{eq:time_function_unstable}.

For the base-case $t = 0$, invoking the properties of the operators $\pop$ and
$\nop$ we have
\begin{align*}
    \enorm{\nop(\tvar)} \leq \enorm{\nop(\tvar) - \pop(\tvar)} +
        \enorm{\pop(\tvar)} &\stackrel{(i)}{\leq} \sup_{\radius_1 \leq
          \|\theta\| \leq \radius_2} \enorm{\nop(\theta) -
          \pop(\theta)} + \enorm{\tvar} \\ &\stackrel{(ii)}{\leq}
        \frac{\mydndelta}{\radius_1 ^{\pargammaMod}} + \radius_2,
 \end{align*}
where step (i) follows since $\enorm{\tvar} \in [\radius_1,
  \radius_2]$ and the operator $\pop$ is $1$-Lipschitz, and
step~(ii) follows from the instability
condition~\eqref{eq:sample_instability}.  This proves the base case of
the induction hypothesis~\eqref{eqn:induction_ub}.

Now we prove the inductive step. In particular, we assume that the
induction hypothesis~\eqref{eqn:induction_ub} holds for $\iter \leq
\tfunUnstable(\radius_1, \radius_2) - 1$ and show that the upper
bound~\eqref{eqn:induction_ub} holds for $\iter' = \iter + 1$. Towards
this end, unwrapping the expression for $ \enorm{\nop^{\iter +
    1}(\tvar)}$ we have
\begin{align*}
\enorm{\nop^{\iter'}(\tvar) } & \leq \enorm{\nop^{\iter + 1}(\tvar) -
  \pop(\nop^{\iter}(\tvar))} + \enorm{\pop( \nop^{\iter}(\tvar))} \\ &
\stackrel{(iii)}{\leq} \sup_{ \radius_1 \leq \enorm{\tvar} \leq
  2\radius_2} \enorm{\nop(\tvar) - \pop(\tvar)} +
\enorm{\nop^{\iter}(\tvar)} \\ & \stackrel{(iv)}{\leq}
\frac{\mydndelta}{\radius_1^{\pargammaMod}} + \iter
\frac{\mydndelta}{\radius_1^{\pargammaMod}} + \radius_2 \\ & = (\iter
+ 1) \frac{\mydndelta}{\radius_1 ^{\pargammaMod}} + \radius_2.
\end{align*}
Here, step (iii) follows from the fact that
$\enorm{\nop^{\iter}(\tvar)} \geq \radius_1$ and the $\lipschitz$
condition~\eqref{eq:lipschitz}; step (iv) stems from the instability
condition~\eqref{eq:sample_instability} and the induction hypothesis.
This completes the proof of the intermediate
claim~\eqref{eqn:induction_ub}.


\section{Proofs of corollaries} 
\label{sec:proofs_of_corollaries}   

We now collect the proofs of several corollaries stated in the paper.
As a high-level summary, our analysis in all three examples in
Section~\ref{sec:specific_models} involves applying
Theorem~\ref{theorem:slow_meta_theorem} to analyze gradient descent/ascent
and EM, both of which are stable algorithms and exhibit slow
convergence for the considered examples.  We invoke
Theorem~\ref{theorem:fast_unstab_meta_theorem}(b) to characterize the
cubic-regularized Newton algorithm, a slowly convergent and unstable
algorithm. Finally, the analysis of Newton's method in all the
examples relies on Theorem~\ref{theorem:fast_unstab_meta_theorem}(a).
Appendices~\ref{sub:proof_of_corollary_ref_grad_info} and \ref{ssub:proof_of_cor:Newton_mixture_model} 
are devoted to the
proofs of Corollaries~\ref{cor:grad_infor_response}
and~\ref{corollary:Newton_mixture}, respectively.  We then prove
Corollary~\ref{cor:grad_single_index} in
Appendix~\ref{subsec:proof:corollary:grad_single_index}. In this
section, the values of universal constants (e.g., $\unicon$,
$\unicon'$ etc.) can change from line-to-line.


\subsection{Proof of Corollary~\ref{cor:grad_infor_response}} 
\label{sub:proof_of_corollary_ref_grad_info}

In this appendix, we demonstrate the convergence and stability
of the gradient and Newton methods. The operators for the gradient method and Newton's method take the
following forms
\begin{subequations}
\begin{align}
  \label{eq:gradient_ascent_infor}
  \resga( \tvar) & = \theta + \learnrate \likelihoodres' (
  \tvar), \quad  \text{and} \quad \samresga( \tvar) = \theta +
  \learnrate \likelihoodres_{n}' ( \tvar), \\
  \label{eq:Newton_infor_response}
  \resnm( \tvar) & = \tvar - \brackets{ \frac{
      \likelihoodres' ( \tvar)} { \likelihoodres'' (
      \tvar)}}, \quad \text{and}\quad \samresnm( \tvar) =
  \tvar - \brackets{ \frac{
      \likelihoodres_n' ( \tvar)} { \likelihoodres_n'' (
      \tvar)}}.
\end{align}
\end{subequations}


\subsubsection{Proofs for the gradient operators} 
\label{ssub:proofs_for_the_gradient_descent_operator_infor_resp}

In lieu of the discussion around
Corollary~\ref{cor:grad_infor_response} it remains to establish that
(a) the operator $\resga$ exhibits a slow convergence condition
$\slownotag(\frac{1}{2})$ over the Euclidean ball $\ballnotag ( \tstar, 1/ 2)$ and (b) the operator
$\samresga$ satisfies a stability condition $\stabilitynotag(1 )$ over the Euclidean ball $\ballnotag ( \tstar, 1/ 2)$ with noise function
$\noise(\obs,\tol) = \sqrt{\log(1/\tol)/n}$ when $n \geq c \log(1 / \delta)$ for some universal constant $c > 0$.

\paragraph{Slow convergence of $\resga$}
Direct computation with the gradient of
population log-likelihood function $\likelihoodres$ leads to
\begin{align}
\label{eq:grad_loglihood_resp}
\likelihoodres' ( \tvar) &\mydefn \frac{ \tvar}{2 ( \tvar^2 + 1) (2
  \sqrt{1 + \tvar^2} - 1)} - \frac{ \tvar}{2}
\\ \quad\Longrightarrow\quad
\resga( \tvar) &= \tvar \brackets{1 - \learnrate \parenth{\frac{1}{2}
    - \frac{1}{2 ( \tvar^2 + 1)(2 \sqrt{1 + \tvar^2} - 1)}}}. \notag
\end{align}
Noting that the fixed point of the population operator is $\tstar = 0$
and that $\learnrate \leq 8/ 3$, we find that
\begin{align*}
\abss{\resga( \tvar) - \tstar} & = \abss{ \tvar} \brackets{1 -
  \learnrate \parenth{\frac{1}{2} - \frac{1}{2 ( \tvar^2 + 1)(2
      \sqrt{1 + \tvar^2} - 1)}}} \\
& \leq \abss{ \tvar} \brackets{1 - \learnrate \parenth{\frac{1}{2} -
    \frac{1}{2 ( \tvar^2 + 1)}}} \\ & \leq \abss{ \tvar} \parenth{1 -
  \frac{\learnrate \tvar^2}{4}}
  \quad\text{for all $\abss{ \tvar} \in [0, 1/ 2]$}.
\end{align*}
Thus the population operator $\resga$ satisfies a slow convergence
condition $\slownotag(\frac{1}{2})$ over the ball $\ballnotag ( \tstar, 1/ 2)$.


\paragraph{Stability of the sample operator $\samresga$}
\label{subsec:proof:sam_stab:cor:grad_infor_response}
We have
\begin{align*}
    \abss{\samresga( \tvar) - \resga( \tvar)} & = \learnrate
    \abss{\nabla \likelihoodres ( \tvar) - \nabla \likelihoodres_{n} (
      \tvar)} \\ & \leq \learnrate \biggr(\abss{\frac{ \tvar}{2
        (\tvar^2 + 1) \parenth{2 \sqrt{1 + \tvar^2} - 1}}
      \parenth{\frac{2}{n} \sum_{i = 1}^{n} (1 - R_{i}) - 1}}
    \\ & \hspace{ 12 em} + \abss{ \tvar \parenth{\frac{1}{2} -
        \frac{1}{n} \sum_{i = 1}^{n} R_{i} Y_{i}^2}}\biggr).
\end{align*}
Recall that, $R_{1}, \ldots, R_{n}$ are i.i.d. samples from Bernoulli
distribution with probability $1/ 2$.  Invoking Hoeffding's inequality
yields that
\begin{align}
\label{eq:grad_response_first_bound}
\abss{\frac{2}{n} \sum_{i = 1}^{n} (1 - R_{i}) - 1} \leq \unicon \;
\sqrt{\frac{\log(1/ \delta)}{n}},
\end{align}
with probability at least $1 - \delta$.  Additionally, as $Y_{1},
\ldots, Y_{n}$ are i.i.d. samples from standard Gaussian distribution
$\NORMAL(0, 1)$ and $R_{1}, \ldots, R_{n}$ are independent of $Y_{1},
\ldots, Y_{n}$, by following the same argument as that in the proof of
Lemma 1 from the paper~\citep{Raaz_Ho_Koulik_2018}, we can demonstrate
that
\begin{align}
\label{eq:grad_response_second_bound}
\abss{\frac{1}{n} \sum_{i = 1}^{n} R_{i} Y_{i}^2 - \frac{1}{2}} \leq
c_{1} \sqrt{\frac{\log(1/ \delta)}{n}},
\end{align}
as long as the sample size $n \geq c_{2} \log(1/ \delta)$ with probability at least $1 - \delta$ where $c_{1}$ and $c_{2}$ are some universal
constants.

Combining the inequalities~\eqref{eq:grad_response_first_bound}
and~\eqref{eq:grad_response_second_bound} yields the following bound
\begin{align*}
  & \hspace{-3 em} \sup_{\tvar \in \ball(\tstar, \radius)} \abss{\samresga( \tvar) -
    \resga( \tvar)} \\
    & \leq c_{3}\sqrt{\frac{\log(1/
      \delta)}{n}} \sup_{\theta \in \ball(\tstar,
      \radius)}\parenth{ \frac{ \abss{\tvar} }{2 (\tvar^2 + 1) \parenth{2
        \sqrt{1 + \tvar^2} - 1}} + \abss{\tvar}} 
    \\ & \leq \frac{3c_{3}r}{2},
\end{align*}
with probability at least $1 - 2 \delta$ for any $r > 0$. Here, the
second inequality in the above display follows from the fact that
$(\tvar^2 + 1) \parenth{2 \sqrt{1 + \tvar^2} - 1} \geq 1$ for all
$\tvar \in \Rspace$.  Thus, the sample-level operator
$\samresga$ is $\stabilitynotag(1 )$-stable over the Euclidean ball $\ballnotag ( \tstar, 1/ 2)$ with noise function
$\noise(\obs,\tol) = \sqrt{\log(1/\tol)/n}$ when $n \geq c \log(1/ \delta)$ for some universal constant $c > 0$.


\subsubsection{Proof for the Newton operators}
\label{ssub:proof_for_the_newton_operators}

Similar to the proof for Newton operators in over-specified Gaussian
mixtures (see Appendix~\ref{subsec:proof:Newton_mixture_model}), we
first verify the geometric convergence of population operator $\resnm$
and the instability condition of sample operator $\samresnm$. Then, we validate Assumption~\ref{item:unstab_control} by showing that
the Newton updates are monotone decreasing and satisfy the following
lower bound
\begin{align}
\abss{ \resnm( \tvar)} & \geq \abss{
  \tvar_{n}^{*}}, \label{eq:lower_infor_resp}
\end{align}
for all 
$\abss{ \tvar} \in [\abss{ \tvar_{n}^{*}}, 1/ 2]$ 
for any global maxima $\tvar_{n}^{*}$ 
of the sample log-likelihood function $\likelihoodres_{n}$ 
in equation~\eqref{eq:sam_likeli_infor}.
\paragraph{Geometric convergence of $\resnm$}
\label{subsec:geo_conver:corollary:Newton_infor_response}
We can verify that $ \likelihoodres'' ( \tvar) < 0$ for all $\tvar
\in \Rspace$.  Additionally, we have the following equation
\begin{align*}
\abss{\resnm( \tvar) - \tstar} 
= \abss{ \tvar - \tstar} \frac{\tvar^2 T_{2}( \tvar)}
{T_{1}( \tvar) + \tvar^2 T_{2}( \tvar)},
\end{align*}
where the functions $T_{1}$ and $T_{2}$ are defined as
\begin{subequations}
\begin{align*}
T_{1}(\tvar) & \mydefn \dfrac{1}{2} - \dfrac{1}{2 (\tvar^2 + 1) (2
  \sqrt{\tvar^2 + 1} - 1)}, \quad \mbox{and} \\
T_{2}(\tvar) & \mydefn \dfrac{1}{2 (\tvar^2 + 1)^2 (2 \sqrt{\tvar^2 +
    1} - 1)} \parenth{3 + \dfrac{1}{2 \sqrt{\tvar^2 + 1} - 1}}.
\end{align*}
\end{subequations}
From the earlier proof argument for slow convergence of $\resga$, we
have $T_{1} (\tvar) \geq \frac{ \tvar^2}{8}$ for all $\abss{ \tvar}
\in [0, 1/ 2]$. Given the above lower bound of $T_{1}$, we directly
obtain that
\begin{align*}
\abss{\resnm( \tvar) - \tstar} \leq \abss{ \tvar - \tstar} \frac{T_{2}
  ( \tvar)} {1/ 8 + T_{2} ( \tvar)} \leq \abss{\tvar - \tstar}
\frac{T_{2} ( 1/ 2)} {1/ 8 + T_{2}( 1/ 2)} \leq \frac{4}{5}
\abss{\tvar - \tstar},
\end{align*} 
for all $\abss{ \tvar} \in [0, 1/ 2]$ 
where the last inequality is due to the fact that 
$T_{2} ( \tvar)/ (\unicon + T_{2} ( \tvar))$ 
achieves its maximum value at $\abss{ \tvar} = 1/ 2$. 
Therefore, the population operator $\resnm$ is $\fastnotag (4/ 5)$-convergent on the ball $\ball(\tvar^{*}, 1/ 2)$. 

\paragraph{Instability of the sample Newton operator $\samresnm$}
Given the formulations of population operator $\resnm$ and sample
operator $\samresnm$ from Newton's method, we have the following
inequality
\begin{align*}
\abss{ \samresnm ( \tvar) - \resnm ( \tvar)} 
\leq \underbrace{\abss{\frac{ \likelihoodres' ( \tvar) 
- \likelihoodres_{n}' ( \tvar)}{ 
\likelihoodres'' ( \tvar)}}}_{\mydefn J_{1}} 
+ \underbrace{\abss{\likelihoodres_{n}' ( \tvar)
 \parenth{\frac{1}{\likelihoodres'' ( \tvar)} 
 - \frac{1}{ \likelihoodres_{n}'' ( \tvar)}}}}_{\mydefn J_{2}}.
\end{align*}
We claim the following upper bounds of $J_{1}$ and $J_{2}$:
\begin{align}
    \label{eq:upper_J1_response}
    J_{1} \leq c_{1} \frac{1}{\abss{ \tvar}} \sqrt{\frac{\log(1/ \delta)}{n}},
\end{align}
with probability at least $1 - 2 \delta$ as long as $\abss{ \tvar} \in [0, 1/ 2]$ and $n \geq c' \log(1/ \delta)$, and 
\begin{align}
J_{2} \leq c_{2} \cdot \frac{1}{
  \abss{ \tvar}} \sqrt{\dfrac{\log(1/
    \delta)}{n}}, \label{eq:upper_J2_response}
\end{align}
with probability at least $1 - 6 \delta$ when $\abss{ \tvar} \geq
\sqrt{2 \unicon} \parenth{\log( 1/ \delta)/ n}^{1/ 4}$.

With the upper bounds~\eqref{eq:upper_J1_response}
and~\eqref{eq:upper_J2_response} of $J_{1}$ and $J_{2}$ respectively,
we arrive at the following inequality
\begin{align*}
\abss{ \samresnm ( \tvar) - \resnm ( \tvar)} \leq c'' \abss{ \tvar}^{- 1} \sqrt{\log(1/ \delta)/ n},
\end{align*}
with probability at least $1 - 8 \delta$ as long as $\sqrt{2 \unicon}
\parenth{\log( 1/ \delta)/ n}^{1/ 4} \leq \abss{ \tvar} \leq 1/ 2$.
As a consequence, the sample operator $\samresnm$ satisfies
instability condition $\instabilitynotag (1 )$ over the annulus $\annulus ( \tstar, \sqrt{2 \unicon}
\parenth{\log( 1/ \delta)/ n}^{1/ 4}, 1/ 2)$ with noise function $\noise(\obs,
\tol) = \sqrt{\dfrac{\log(1/ \delta)}{n}}$ as long as $n \geq c' \log(1/ \delta)$.
\paragraph{Proof for the upper bound of $J_{1}$}
When $n \geq c' \log(1/ \delta)$, we can validate that
\begin{align*}
\abss{ \likelihoodres' ( \tvar) 
    - \likelihoodres_{n}' ( \tvar)} \leq c \abss{ \tvar}\sqrt{\frac{\log(1/ \delta)}{n}},
\end{align*}
for any $\abss{ \tvar} \in [0, 1/ 2]$ with probability at least $1 - 2 \delta$ where $c$ and $c'$ are some universal constants. 
Furthermore, based on the computations in Appendix~\ref{subsec:geo_conver:corollary:Newton_infor_response}, 
we find that
\begin{align}
\label{eq:upper_J1_first}
\abss{ \likelihoodres'' ( \tvar)} = T_{1}( \tvar) 
+ \tvar^2 T_{2}(\tvar) 
\geq \frac{\tvar^2}{8} + \tvar^2 T_{2}(1/ 2) \geq \frac{11 \tvar^2}{32}, 
\end{align}
for any $\abss{ \theta} \in [0, 1/ 2]$. Combining the previous inequalities, 
we have the following upper bound with $J_{1}$:
\begin{align*}
J_{1} \leq c_{1} \frac{1}{\abss{ \tvar}} \sqrt{\frac{\log(1/ \delta)}{n}},
\end{align*}
with probability at least $1 - 2 \delta$ as long as $\abss{ \tvar} \in [0, 1/ 2]$ and $n \geq c' \log(1/ \delta)$. 
\paragraph{Proof for the upper bound of $J_{2}$} In order to derive an upper bound for $J_{2}$, we make use of the following bounds:
\begin{subequations}
\begin{align}
    \abss{ \likelihoodres_{n}' ( \tvar)} & \leq c_{1} \parenth{
          \abss{ \tvar} \sqrt{\frac{\log(1/ \delta)}{n}} + \abss{
            \tvar}^3}, \label{eq:upper_J2_first_bound} \\ \abss{
          \likelihoodres_{n}'' ( \tvar) - \likelihoodres'' ( \tvar)} &
        \leq c_{2} \sqrt{\frac{\log(1/
            \delta)}{n}}, \label{eq:upper_J2_second_bound} \\ \abss{
          \likelihoodres_{n}'' ( \tvar)} & \geq c_{3} \parenth{
          \tvar^2 - \unicon \cdot \sqrt{\frac{\log(1/
              \delta)}{n}}}, \label{eq:upper_J2_third_bound}
\end{align}
\end{subequations}
for all $\abss{ \tvar} \in [0, 1/ 2]$ with probability at least $1 - 2 \delta$ when $n \geq c' \log(1/ \delta)$. Here, $\unicon, c_{1}, c_{2}, c_{3}$ in the above bounds are 
universal constants independent of $\delta$.

Deferring the proofs of these claims to later, we now proceed to give
an upper bound for $J_{2}$ based on the given bounds in the above
display.  In particular, from the formulation of $J_{2}$, we achieve
that
\begin{align*}
J_{2} & \leq \frac{32 c_{1} c_{2}}{11 c_{3}} \parenth{\abss{ \tvar}
  \sqrt{\frac{\log(1/ \delta)}{n}} + \abss{ \tvar}^3}
\dfrac{\sqrt{\frac{\log(1/ \delta)}{n}}} { \tvar^2 \parenth{\tvar^2 -
    \unicon \sqrt{\frac{\log(1/ \delta)}{n}}}} \\
    & \leq C \cdot \frac{1}{
  \abss{ \tvar}} \sqrt{\dfrac{\log(1/
    \delta)}{n}}
\end{align*}
with probability at least $1 - 6 \delta$ when $\abss{ \tvar} \geq
\sqrt{2 \unicon} \parenth{\log( 1/ \delta)/ n}^{1/ 4}$ where $C$ is
some universal constant.  Here, the last inequality is due to $\abss{
  \tvar} \sqrt{\frac{\log( 1/ \tol)}{n}} + \abss{\tvar}^3 \leq
\abss{\tvar}^3 \parenth{1 + \frac{1}{2c}}$ and 
\mbox{$\tvar^2 - \unicon
\sqrt{\frac{\log(1/ \delta)}{n}} \geq \abss{\tvar}^2/ 2$} as long as we have
$\abss{ \tvar} \geq \sqrt{2 \unicon} \parenth{\log( 1/ \delta)/ n}^{1/
  4}$.
\paragraph{Proof of claim~\eqref{eq:upper_J2_first_bound}} 
Invoking triangle inequality, when $n \geq c' \log(1/ \delta)$ we have
\begin{align*}
\abss{ \likelihoodres_{n}' ( \tvar)} \leq c \abss{ \tvar} \parenth{\sqrt{\frac{\log (1/ \delta)}{n}} 
+ \frac{1}{2} - \frac{1}{2 (\tvar^2 + 1) 
\parenth{2 \sqrt{ \tvar^2 + 1} - 1}}},
\end{align*}
with probability at least $1 - 2 \delta$ for any 
$\abss{ \tvar} \in [0, 1/ 2]$ 
where the inequality in the above display is 
due to the inequalities~\eqref{eq:grad_response_first_bound} 
and~\eqref{eq:grad_response_second_bound}. 
Furthermore, we can validate that
\begin{align*}
\frac{1}{2} - \frac{1}{2 (\tvar^2 + 1) 
\parenth{2 \sqrt{ \tvar^2 + 1} - 1}} 
\leq \frac{3 \tvar^2}{2}
\end{align*}
for any $\abss{ \tvar} \in [0, 1/ 2]$. In light of the previous inequalities, we arrive at the following inequality
\begin{align*}
\abss{ \likelihoodres_{n}' ( \tvar)}  
\leq \frac{3 c \abss{ \tvar}}{2} \parenth{\sqrt{\frac{\log (1/ \delta)}{n}} 
+ \tvar^2},
\end{align*}
with probability at least $1 - 2 \delta$ 
for all $\abss{ \tvar} \in [0, 1/ 2]$. 
As a consequence, we reach the conclusion of claim~\eqref{eq:upper_J2_first_bound}. 
\paragraph{Proof of claims~\eqref{eq:upper_J2_second_bound} and~\eqref{eq:upper_J2_third_bound}}
The proof of claim~\eqref{eq:upper_J2_second_bound} is a direct application of triangle inequality and the fact that $\abss{ \tvar} \in [0, 1/ 2]$. In addition, we have
\begin{align*}
\abss{ \likelihoodres_{n}'' ( \tvar) }
\geq \abss{ \likelihoodres'' ( \tvar)} 
- \abss{ \likelihoodres_{n}'' ( \tvar) - 
\likelihoodres'' ( \tvar)}
\geq c' \parenth{ \tvar^2 - c \sqrt{\dfrac{\log(1/ \delta)}{n}}},
\end{align*}
with probability at least $1 - 2 \delta$ for any 
$\abss{ \tvar} \in [0, 1/ 2]$ where $c, c'$ are universal constants independent of $\delta$ 
and the last inequality in the above display is due the results 
from equation~\eqref{eq:upper_J1_first} and claim~\eqref{eq:upper_J2_second_bound}.  
As a consequence, we achieve the conclusion of claim~\eqref{eq:upper_J2_third_bound}. 
\paragraph{Lower bound and monotonicity of Newton updates}
Now, we proceed to verify the lower bound of Newton updates in
claim~\eqref{eq:lower_infor_resp}.  In order to ease the ensuing
presentation, we denote $ f( \tvar) \mydefn \frac{1}{(\tvar^2 + 1) ( 2
  \sqrt{ \tvar^2 + 1} - 1)}$ for all $\tvar$. The global maxima
$\tvar_{n}^{*}$ of the sample log-likelihood function
$\likelihoodres_{n}$ are the solutions of the following equation
\begin{align*}
\tvar_{n}^{*} f( \tvar_{n}^{*}) 
\parenth{ \frac{1}{n} \sum_{i = 1}^{n} (1 - R_{i})} 
= \tvar_{n}^{*} \parenth{ \frac{1}{n} 
\sum_{i = 1}^{n} R_{i} Y_{i}^2}.
\end{align*}
The specific forms of $\tvar_{n}^{*}$ depend on 
the values of $R_{i}, Y_{i}$ for $i \in [n]$. 
In particular, when $\sum_{i = 1}^{n} R_{i} Y_{i}^2 
< \sum_{i = 1}^{n} (1 - R_{i})$, namely, 
the Hessian of sample likelihood function 
$\likelihoodres_{n}$ at 0 is positive, 
the function $\likelihoodres_{n}$ 
is bimodal and symmetric around 0. 
Additionally, $\tvar_{n}^{*}$ are different from 0 
and become the solution of the following equation
\begin{align}
\label{eq:global_maxima_infor_res}
f( \tvar_{n}^{*}) \parenth{ \frac{1}{n} \sum_{i = 1}^{n} (1 - R_{i})} = \parenth{ \frac{1}{n} \sum_{i = 1}^{n} R_{i} Y_{i}^2}.
\end{align}
On the other hand, when 
$\sum_{i = 1}^{n} R_{i} Y_{i}^2 > \sum_{i = 1}^{n} (1 - R_{i})$, 
the function $\likelihoodres_{n}$ is unimodal 
and symmetric around 0. 
Under this case, $\tvar_{n}^{*} = 0$ is the unique global maximum.

Without loss of generality, 
we assume that $\tvar > 0$ 
and the global maxima are solutions of equation~\eqref{eq:global_maxima_infor_res}. 
From the formulation of $\samresnm$, 
the inequality $\samresnm( \tvar) > 0$ is equivalent to
\begin{align*}
 \tvar f' ( \tvar) + f( \tvar) < f ( \tvar_{n}^{*}), 
\end{align*} 
which holds for all $\tvar \geq \abss{ \tvar_{n}^{*}}$ 
since $f( \tvar) < f( \tvar_{n}^{*})$ 
and $f' ( \tvar) < 0$ 
as $\tvar \geq \abss{ \tvar_{n}^{*}}$. 
Therefore, we have $\samresnm( \tvar) > 0$ 
for all $\tvar \geq \abss{ \tvar_{n}^{*}}$. 
Now, in order to demonstrate that 
$\samresnm( \tvar) \geq \abss{ \tvar_{n}^{*}}$ 
for $\tvar \geq \abss{ \tvar_{n}^{*}}$, it is equivalent to 
\begin{align}
\label{eq:equiv_lower_infor_res}
\parenth{ \abss{ \tvar_{n}^{*}} - \tvar} 
\tvar f' ( \tvar) + \abss{ \tvar_{n}^{*}} 
\parenth{ f( \tvar) - f( \tvar_{n}^{*})} \geq 0.
\end{align}
Invoking mean value theorem, 
we can find some constant 
$\bar{ \tvar} \in ( \abss{ \tvar_{n}^{*}}, \tvar)$ such that
\begin{align*}
f( \tvar) - f( \tvar_{n}^{*})  = f( \tvar) - f( \abss{ \tvar_{n}^{*}} )
= f' ( \bar{ \tvar}) ( \tvar - \abss{ \tvar_{n}^{*}}).
\end{align*}
Given the above equation, the inequality~\eqref{eq:equiv_lower_infor_res} can be rewritten as
\begin{align}
\label{eq:final_lower_infor_res}
\abss{ \tvar_{n}^{*}} f' ( \bar{ \tvar}) 
\geq \tvar f' ( \tvar) 
\end{align}
for all $\tvar \geq \abss{ \tvar_{n}^{*}}$.  Since the function $\tvar
f' ( \tvar)$ is a decreasing function in $(0, 1/ 2]$, we have $\tvar
  f' ( \tvar) \leq \bar{ \tvar} f' ( \bar{ \tvar})$ for any $\bar{
    \tvar} < \tvar$.  Since $f' ( \bar{ \tvar}) < 0$ and $\bar{ \tvar}
  > \abss{ \tvar_{n}^{*}}$, we find that $\bar{ \tvar} f' ( \bar{
    \tvar}) \leq \abss{ \tvar_{n}^{*}} f' ( \bar{ \tvar})$.  In light
  of these two inequalities, we achieve the
  inequality~\eqref{eq:final_lower_infor_res}.  As a consequence, we
  reach the conclusion of claim~\eqref{eq:lower_infor_resp}.


\subsection{Proof of Corollary~\ref{corollary:Newton_mixture}}
\label{ssub:proof_of_cor:Newton_mixture_model}
Under the model~\eqref{Eqnsingular_location_scale}, the sample EM
operator takes the form
\begin{align*}
  \nopem(\tvar) = \frac{1}{n} \sum_{i = 1}^{n} X_{i} \tanh(\tvar
  X_{i}),
\end{align*}
where $\tanh(x) = \frac{\exp( x) - \exp( - x)}{ \exp( x) + \exp( -
  x)}$ is the hyperbolic tangent.  In our prior work (cf. Theorem 3 in
the paper~\citep{Raaz_Ho_Koulik_2018}), we studied the sample EM
operator for this model.

Accordingly, in this paper, we limit our analysis to the Newton
updates; see Appendix~\ref{subsec:proof:Newton_mixture_model} for the
details.  The sample and population Newton updates take the form
\begin{subequations}
\label{eq:newton_em_operators}
\begin{align}
\label{eq:newton_em_sample}  
  \newem(\tvar) & = \tvar - \brackets{ \likelihood'' ( \tvar)}^{-1}
  \likelihood' (\tvar) = \tvar + \frac{\Exs \brackets{ X \tanh( X
      \tvar)} - \tvar} {\Exs \brackets{X^2 \tanh^{2} (X \tvar)}},
  \quad\text{and} \\
  \label{eq:newton_em}  
\samnewem(\tvar) & = \tvar - \brackets{ \samlikelihood'' (
  \tvar)} ^{-1} \samlikelihood' ( \tvar) \nonumber \\
  & = \tvar + \frac
         {\big(\frac{1}{n} \sum_{i = 1}^{n} X_{i} \tanh(X_{i}
           \tvar)\big) - \tvar} {\frac{1}{n}\sum_{i = 1}^{n} X_{i}^2
           \tanh^2(X_{i} \tvar) + 1 - \frac{1} {n} \sum_{i = 1}^{n}
           X_{i}^2}.
\end{align}
\end{subequations}

\subsubsection{Proofs for Newton operators}
\label{subsec:proof:Newton_mixture_model}

We begin by verifying the fast convergence of the operator $\newem$
and then the instability of the operator $\samnewem$ with respect to
$\newem$ in Theorem~\ref{theorem:fast_unstab_meta_theorem}. 
Then, we demonstrate that the Newton updates satisfy Assumption~\ref{item:unstab_control}. 
Noting that it can be done by establishing
that the Newton updates are monotone decreasing and admit the
following lower bound
\begin{align}
\abss{ \samnewem( \tvar)} & \geq \abss{
  \tvar_{n}^{*}} \label{eq:lower_mixture_model}
\end{align}
for all $\abss{ \tvar} \in [\abss{ \tvar_{n}^{*}}, 1/ 3]$ for any
global maximum $\tvar_{n}^{*}$ of $\samlikelihood$.

\paragraph{Fast convergence of the population-level operator $\newem$}
We provide the full proof for the case $\tvar \in (0, \tfrac{1}{3}]$;
  the proof for the case $\tvar \in [-\tfrac{1}{3}, 0)$ is analogous.
    We make use of the following known
    bounds~\citep{Raaz_Ho_Koulik_2018_second} on the hyperbolic
    function $x \mapsto x \tanh(x)$:
\begin{align}
\label{eq:bound_hyperbolic}
x^2 - \frac{x^{4}}{3} \leq x \tanh(x) \leq x^2 - \frac{x^4}{3} +
\frac{2 x^6}{15} \qquad \mbox{for all $x \in \real$.}
\end{align} 
Applying this bound, we find that
\begin{align*}
  \Exs \brackets{X \tanh(X \tvar)} & \leq \frac{1}{\tvar} \Exs
  \brackets{(X \tvar)^2 - (X \tvar)^4/3 + 2 (X \tvar)^6/ 15} = \tvar -
  \tvar^3 + 2 \tvar^5, \quad \mbox{as well as} \\
\Exs \brackets{X^2 \tanh^2(X \tvar)} & \leq \frac{1}{\tvar^2} \Exs
\brackets{(X \tvar)^4} = 3 \tvar^2,
\end{align*}
and consequently that
\begin{align*}
\frac{\tvar - \Exs \brackets{ X \tanh(X \tvar)}} {\Exs \brackets{X^2
    \tanh^2(X \tvar)}} \geq \frac{ \tvar - (\tvar - \tvar^3 + 2
  \tvar^5)}{ 3 \tvar^2} = \frac{ \tvar - 2 \tvar^3}{3} 
  \stackrel{(\tvar \in (0,
\tfrac{1}{3}])}{\geq}
  \frac{2\tvar}{9}.
\end{align*}
Noting that $\newem(\tvar) = \tvar - \frac{\tvar -
  \Exs \brackets{ X \tanh (X \tvar)}}{\Exs \brackets{X^2 \tanh^2(X
    \tvar)}} $ and $\tstar=0$, we conclude that the population Newton
operator $\newem$ is $\fastnotag (\tfrac{7}{9})$-convergent over the ball
$\ball(\tstar, \tfrac{1}{3})$.

\paragraph{Instability of the sample-level operator $\samnewem$}

Let us introduce the shorthand
\begin{align*}
A_{n} \mydefn \frac{1}{n} \sum_{i = 1}^{n} X_{i} \tanh(X_{i} \tvar),
\quad \mbox{and} \quad B_{n} \mydefn \frac{1}{n} \sum_{i = 1}^{n}
X_{i}^2 \tanh^2(X_{i} \tvar) + 1 - \frac{1}{n} \sum_{i = 1}^{n}
X_{i}^2.
\end{align*}
Using the definitions~\eqref{eq:newton_em} of the operators $\samnewem$
and $\newem$, we find that
\begin{align}
\label{eqn:key_ineq_deviate}
& \abss{ \samnewem( \tvar) - \newem( \tvar)} \\
        & = \abss{ \frac{ \Exs \brackets{ X \tanh(X \tvar)} - 
        \tvar} {\Exs \brackets{X^2 \tanh^2(X \tvar)}} - 
        \frac{A_{n} - \tvar}{B_{n}}} \nonumber \\
        & \leq \underbrace{\frac{\abss{\Exs \brackets{ X \tanh(X 
        \tvar)} - A_{n}}}{\Exs \brackets{X^2    
        \tanh^2(X \tvar)}}}_{ : = J_{1}}  + 
        \underbrace{\abss{A_{n} - \tvar} \abss{\frac{1}{\Exs 
        \brackets{X^2 \tanh^2(X 
        \tvar)}} - \frac{1}{B_{n}}}}_{ : = J_{2}}. \nonumber
\end{align}
Thus, in order to bound the difference $\abss{ \samnewem( \tvar) -
  \newem( \tvar)}$, it suffices to derive bounds for the terms $J_1$
and $J_2$.

\paragraph{Upper bound for $J_{1}$}

For a given $\tol \in (0, 1)$, as long as the sample size $n \geq C
\log( 1/ \tol)$ for some universal constant $C$, we can apply Lemma 1 from the
paper~\citep{Raaz_Ho_Koulik_2018} to assert that
\begin{align}
\label{eqn:simple_bound}
\abss{\Exs \brackets{ X \tanh(X \tvar)} - A_{n}} \leq \unicon \,
\abss{\tvar} \sqrt{\frac{\log( 1/ \tol)}{n}} \quad \mbox{for all
  $\abss{\theta} \in (0, \tfrac{1}{3})$}
\end{align}
with probability $1 - \delta$.  Moreover, the
bound~\eqref{eq:bound_hyperbolic} implies that
\begin{align*}      
\Exs \brackets{X^2 \tanh^2(X \tvar)} \geq \frac{1}{\tvar^2} \Exs
\brackets{\big({(X \tvar)^2 - \frac{(X \tvar)^4}{3}}\big)^2} = 3
\tvar^2 - 10 \tvar^4 + \frac{ 35\tvar^6}{33} \geq 2 \tvar^2,
\end{align*}
for $\tvar \in [-\tfrac{1}{3}, \tfrac{1}{3} ]$.  Combining the above
inequalities yields
\begin{align}
\label{eqn:first_key_bound}
J_{1} = \frac{\abss{\Exs \brackets{ X \tanh(X \tvar)} - A_{n}}}{\Exs
  \brackets{X^2 \tanh^2(X \tvar)}} \leq \unicon \frac{\abss{ \tvar}
  \sqrt{\frac{\log( 1/ \tol)}{n}}}{2 \tvar^2} \leq \unicon'
\frac{1}{\abss{ \tvar}} \sqrt{\frac{\log( 1/ \tol)}{n}},
\end{align}
for all $\abss{\theta} \in (0, 1/ 3)$ with probability at least $1 - \tol$. 
\paragraph{Upper bound for $J_{2}$} In order to obtain an upper 
bound for $J_{2}$, we 
claim the following key bounds appearing in its formulation:
\begin{subequations}
\begin{align}
 \abss{A_{n} - \tvar} 
    & \leq c_{1} \parenth{ \abss{\tvar} \sqrt{\frac{\log( 1/ \tol)}{n}} + 
    \abss{\tvar}^3}, 
    \label{eqn:second_key_bound} \\
 \abss{B_{n}} 
    & \geq c_{2} \parenth{ \tvar^2 - c\frac{\log^4(3 n/ \tol)}{\sqrt{n}}},  
    \label{eqn:third_key_bound} \\
\abss{\Exs \brackets{X^2 \tanh^2(X \tvar)} 
    - B_{n}} 
    & \leq c_{3} \frac{\log( n/ \tol)}{\sqrt{n}} ,
    \label{eqn:fourth_key_bound}
\end{align} 
\end{subequations}
for all $\abss{ \tvar } \in (0, 1/ 3]$ with probability at least $1 - 2 
    \tol$
as long as the sample size $n \geq c \log( 1/ \tol)$. 
Here, $c, c_{1}, c_{2}, c_{3}$ in the above probability bounds are universal constants 
independent of $\tol$. 
Assume that the above claims are given at the 
moment. The 
results in these claims lead to
\begin{align}
\label{eqn:fifth_key_bound}
J_{2} \
    & = \abss{A_{n}} \abss{\frac{\Exs \brackets{X^2 \tanh^2(X 
    \tvar)} - B_{n}}{B_{n} \Exs 
        \brackets{X^2 \tanh^2(X 
        \tvar)}}} \nonumber \\
        & \leq c' \parenth{ \abss{ \tvar} \sqrt{\frac{\log( 1/ 
    \tol)}{n}} + 
    \abss{\tvar}^3} \frac{\frac{\log( n/ \tol)}{\sqrt{n}}}
    {\tvar^2 \parenth{ \tvar^2 
    - c \frac{\log^{4}( 3 n/ \tol)}{\sqrt{n}}}} \nonumber \\
    & \leq c'' \frac{1}{\abss{ 
    \tvar}} \frac{\log( n/ \tol)}{\sqrt{n}}
\end{align}
with probability at least $1 - 5 \tol$. Here, the last 
inequality is due to the facts that 
\begin{align*}
\abss{ \tvar} \sqrt{\frac{\log( 1/ \tol)}{n}} + 
\abss{\tvar}^3 \leq \abss{\tvar}^3 \parenth{1 + \frac{1}{2c}}
\quad\text{and} \quad
\tvar^2 - c \frac{\log^{4}( 3 n/ \tol)}{\sqrt{n}} 
\geq \abss{\tvar}^2/ 2,
\end{align*}
as long as $\abss{\tvar} \geq \sqrt{2 c} \log^2(3 n / \tol)/n^{1/ 4}$.
Plugging the bounds~\eqref{eqn:first_key_bound}
and~\eqref{eqn:fifth_key_bound} into
equation~\eqref{eqn:key_ineq_deviate}, the operator
$\samnewem$ is $\instabilitynotag(- 1)$-unstable over the annulus $\annulus ( \tstar, \frac{\sqrt{2 c} \log^2(3 n / \tol)}{n^{1/ 4}}, 1/
3)$ with noise function $\noise(\obs, \tol) = \frac{\log( n/
  \tol)}{\sqrt{n}}$ as long as the sample size \mbox{$n \geq C
  \frac{\log^8(3 n / \tol)}{n^{1/ 4}}$.}


\paragraph{Proof of claim~\eqref{eqn:second_key_bound}} 
Invoking the concentration bound~\eqref{eqn:simple_bound} and applying
the triangle inequality,
we find that
\begin{align*}
\abss{ A_{n} - \tvar} 
    & \leq \abss{\frac{1}{n} \sum_{i = 1}^{n} X_{i} 
    \tanh(X_{i} \tvar) - \Exs \brackets{ X \tanh(X \tvar)}} 
    + \abss{\Exs \brackets{ X \tanh(X \tvar)} - \tvar} 
    \\
    & \leq c \parenth{ \abss{\tvar} \sqrt{\frac{\log( 1/ \tol)}{n}} 
    + \frac{1}{\abss{\tvar}} \abss{\Exs \brackets{ X \tvar 
    \tanh(X \tvar)} - \tvar^2}}
\end{align*}
for all $\abss{ \tvar} \in (0, 1/3]$ with probability $1 - \tol$.
Next, taking expectation on both sides in the bounds~\eqref{eq:bound_hyperbolic},
we find that
\begin{align*}
\Exs \brackets{ X \tvar \tanh(X \tvar)} - \tvar^2 
    & \leq \Exs \brackets{ (X \tvar)^2 - \frac{(X \tvar)^4}{3} 
    + \frac{2 (X \tvar)^6}{15}} - \tvar^2 
    \\
    & = - \tvar^4 + 2 \tvar^6 \leq - \frac{7 \tvar^4}{9}, \quad \text{and}\\
\Exs \brackets{ X \tvar \tanh(X \tvar)} - \tvar^2 
    & \geq \Exs \brackets{ (X \tvar)^2 
    - \frac{(X \tvar)^4}{3}} - \tvar^2 = - \tvar^4.
\end{align*}
Putting these pieces together yields the claim~\eqref{eqn:second_key_bound}.

\paragraph{Proof of claim~\eqref{eqn:third_key_bound}} 
Invoking standard chi-squared concentration bounds and applying triangle
inequality, we obtain that
\begin{align*}
\abss{B_{n}} & \geq \frac{1}{n} \sum_{i = 1}^{n} X_{i}^2 
    \tanh^2(X_{i} \tvar) - \abss{\frac{1}{n} \sum_{i = 1}^{n} 
    X_{i}^2 - 1} \\
    & \geq c \parenth{ \frac{1}{n} \sum_{i = 1}^{n} X_{i}^2 
        \tanh^2(X_{i} \tvar) - \sqrt{\frac{\log( 1/ \tol)}{n}}}
\end{align*}
with probability at least $1 - \tol$. 
Using the lower bound from inequality~\eqref{eq:bound_hyperbolic}, we find
that
\begin{align*}
\frac{1}{n} \sum_{i = 1}^{n} X_{i}^2 
    \tanh^2(X_{i} \tvar) 
    & \geq \frac{1}{n} \sum_{i = 1}^{n} \parenth{\tvar X_{i}^2 
    - \frac{\tvar^3 X_{i}^4}{3}}^2 \\
    & = \tvar^2 \parenth{\frac{1}{n} \sum_{i = 1}^{n} X_{i}^4} 
    - \frac{2 \tvar^4}{3} \parenth{\frac{1}{n} \sum_{i = 1}^{n} 
    X_{i}^6} + \frac{\tvar^6}{9} \parenth{\frac{1}{n} 
    \sum_{i = 1}^{n} X_{i}^8} \\
    & \stackrel{(i)}{\geq}
    \tvar^2 \parenth{3 - c' \frac{\log^2(3 n/ \tol)}
    {\sqrt{n}}} - \frac{2 \tvar^4}{3} \parenth{15 + 
    c' \frac{\log^3(3 n/ \tol)}{\sqrt{n}}} \\
    & \hspace{ 12 em} + \frac{\tvar^6}{9} 
    \parenth{105 - c' \frac{\log^4(3 n/ \tol)}{\sqrt{n}}} \\
    & \geq \tvar^2 - c' \frac{\log^4(3 n/ \tol)}{\sqrt{n}},
\end{align*}
with probability at least $1 - \tol$ for some universal 
constant $c$. Here step~(i) makes use of the following
concentration bound for higher moments of Gaussian random 
variables (Lemma~5~\citep{Raaz_Ho_Koulik_2018_second}):
\begin{align*}
\Prob \brackets{\abss{ \frac{1}{n} \sum_{i = 1}^{n} X_{i}^{2k} - \Exs
    \brackets{X^{2k}}} \leq c' \frac{\log^{k}(3 n/ \tol)}{n^\frac{1}{2}}}
    \geq
1-\frac{\tol}{3} \quad \mbox{ for $k \in \{2, 4, 6\}$}
\end{align*}
with probability at least $1 - \tol/3$ for $k \in \{2, 4, 6\}$.
Putting together the pieces yields the
claim~\eqref{eqn:third_key_bound}.


\paragraph{Proof of claim~\eqref{eqn:fourth_key_bound}} 

Applying the triangle inequality yields
\begin{align}
\label{eqn:sixth_key_bound}
& \abss{\Exs \brackets{X^2 \tanh^2(X \tvar)} - B_{n}} \\
& \leq
\abss{\frac{1}{n} \sum_{i = 1}^{n} X_{i}^2 \tanh^2(X_{i} \tvar) - \Exs
  \brackets{X^2 \tanh^2(X \tvar)}} + \abss{\frac{1}{n} \sum_{i =
    1}^{n} X_{i}^2 - 1} \nonumber \\
& \leq \abss{\frac{1}{n} \sum_{i = 1}^{n} X_{i}^2 \tanh^2(X_{i}
  \tvar) - \Exs \brackets{X^2 \tanh^2(X \tvar)}} + c \sqrt{\frac{\log(
    1/ \tol)}{n}} \nonumber
\end{align}
with probability at least $1 - \tol$.  By adapting the truncation
argument from the proof of Lemma 5 in the
paper~\citep{Raaz_Ho_Koulik_2018_second} for the random variable
$X\tanh(X)$ with $X\sim\NORMAL(0, 1)$, it follows that
\begin{align*}
\abss{\frac{1}{n} \sum_{i = 1}^{n} X_{i}^2 \tanh^2(X_{i} \tvar) - \Exs
  \brackets{X^2 \tanh^2(X \tvar)}} \leq \unicon' \frac{\log (n/ \tol)}{
  \sqrt{ n}},
\end{align*}
for all $\abss{ \theta} \in (0, 1/ 3]$ with probability at least $1 -
  \tol$. Putting the results together yields the
  claim~\eqref{eqn:fourth_key_bound}.

\paragraph{Lower bound and monotonicity of Newton updates}

We first make some observations about the structure of the
log-likelihood function $\samlikelihood$.  Define
\begin{align*}
f(\tvar) & \defn \tvar - \frac{1}{n} \sum_{i=1}^{n} X_{i} \tanh(X_{i}
\tvar).
\end{align*}
When $\sum_{i = 1}^{n} X_{i}^2 > n$, it can be shown (by computing the
gradient and Hessian) that the log-likelihood $\samlikelihood$ is
bimodal and symmetric around $0$.  It has multiple global maxima
$\tvar_{n}^{*}$ that are non-zero, and are solutions of the equation
$f( \tvar) = 0$.  On the other hand, when $\sum_{i = 1}^{n} X_{i}^2
\leq n$, the function $\samlikelihood$ is unimodal and symmetric
around $0$, and the point $\tvar_{n}^{*} = 0$ is the unique global
maximum of the log likelihood.


Next we verify the lower bound of Newton updates $\samnewem( \tvar)$
in claim~\eqref{eq:lower_mixture_model}; the proof of monotonicity can
be argued similarly.  Without loss of generality, we only consider the
setting when the global maxima $\tvar_{n}^{*}$ are different from 0
and $\tvar > 0$.  Under that case, the Hessian of the function
$\samlikelihood$ at $\abss{ \tvar_{n}^{*}}$ is negative.  A direct
computation with the gradient of the function $f$ leads to
\begin{align*}
f'(\tvar) = 1 - \frac{1}{n} \sum_{i = 1}^{n} X_{i}^2 \sech^2( X_{i}
\tvar) & = 1 - \frac{1}{n} \sum_{i = 1}^{n} X_{i}^2 \sech^2(
\abss{X_{i}} \abss{\tvar}) \\ & \geq 1 - \frac{1}{n} \sum_{i = 1}^{n}
X_{i}^2 \sech^2( \abss{X_{i}} \abss{\tvar_{n}^{*}}) \\ & = - \nabla^2
\samlikelihood ( \tvar_{n}^{*}) > 0
\end{align*}
for any $\tvar > \abss{ \tvar_{n}^{*}}$. 
Therefore, the function $f$ 
is a strictly increasing function 
when $\tvar > \abss{ \tvar_{n}^{*}}$. 
It leads to the inequality 
$f( \tvar) \geq f( \tvar_{n}^{*}) = 0$ 
for all $\tvar \geq \abss{ \tvar_{n}^{*}}$. 
Further computation with second derivative of $f$ 
yields that
\begin{align*}
 f''( \tvar) 
= \frac{2}{n} \sum_{i = 1}^{n} X_{i}^3 
\tanh( X_{i} \tvar) \sech^2( X_{i} \tvar) > 0
\end{align*} 
for all $\tvar > 0$.  The above inequality is due to $X_{i}
\tanh(X_{i} \tvar) > 0$ for all $\tvar > 0$ and $i \in [n]$.  Thus, the function
$f'$ is strictly increasing when $\tvar > 0$.

Now the inequality $\samnewem( \tvar) \geq \abss{ \tvar_{n}^{*}}$ for all
$\tvar \geq \abss{ \tvar_{n}^{*}}$ is equivalent to
\begin{align}
\label{eq:final_lower_mixture_model}
f'( \tvar) ( \tvar - \abss{ \tvar_{n}^{*}}) 
\geq f( \tvar) - f( \tvar_{n}^{*}).
\end{align}
Invoking the mean value theorem, we find that
\begin{align*}
f( \tvar) - f( \tvar_{n}^{*}) 
= f( \tvar) - f( \abss{ \tvar_{n}^{*}}) 
= f'( \bar{ \tvar}) ( \tvar - \abss{ \tvar_{n}^{*}})
\end{align*}
for some $\bar{ \tvar} \in ( \abss{ \tvar_{n}^{*}}, \tvar)$. 
Given that equality, the equality~\eqref{eq:final_lower_mixture_model} can be rewritten 
as $f'( \tvar) \geq f'( \bar{ \tvar})$ 
for all $\tvar \geq \abss{ \tvar_{n}^{*}}$. 
This inequality is true 
since $f'$ is an increasing function 
when $\tvar > 0$. As a consequence, we achieve 
the conclusion of claim~\eqref{eq:lower_mixture_model}.

\subsection{Proof of Corollary~\ref{cor:grad_single_index}}
\label{subsec:proof:corollary:grad_single_index}

In this appendix, we demonstrate the convergence and stability
properties of operators from gradient descent and (cubic-regularized)
Newton's methods in the non-linear regression model. The sample operators of these methods take the following forms
\begin{subequations}
    \begin{align}
        \label{eq:grad_single_index}
        \nopgd( \tvar) 
        &= \tvar - \learnrate  \likelihoodsing_{n}' ( \tvar)
        = \tvar - \learnrate \bigg({ \frac{2 p}{n} \sum_{i = 1}^{n} X_
        {i}^{4 p} \tvar^{4 p - 1} - \frac{2 p}{n} 
        \sum_{i = 1}^{n} Y_{i} X_{i}^{2 p} \tvar^{2 p -1}}\bigg),
        \\
        \nopnm( \tvar) 
        & = \tvar - \brackets{  \likelihoodsing_{n}'' ( \tvar)}^{-1}
        \likelihoodsing_{n}' ( \tvar) 
        \notag\\
        &= \tvar - \frac{\parenth{\frac{1}{n} \sum_{i = 1}^{n} X_{i}^{4 p}}
        \tvar^{2 p + 1} - \parenth{\frac{1}{n} \sum_{i = 1}^{n} Y_{i} X_{i}^{2 p}} \tvar}
        {\parenth{\frac{4 p - 1}{n} \sum_{i = 1}^{n} X_{i}^{4 p}}\tvar^{2 p} - \frac{2 p - 1}{n} \sum_{i = 1}^{n} Y_{i} X_{i}^{2 p}}, \quad\text{and} \label{eq:Newton_single_index} \\
        \nopcnm (\tvar) 
        & = \mathop{ \arg \min}_{y \in \Rspace} 
        \left\{ \likelihoodsing_{n}' (\tvar) (y - \tvar) 
        + \frac{1}{2} \likelihoodsing_{n}'' (\tvar) (y - \tvar)^2 
        + \regular \abss{y - \tvar}^3 \right\}, \label{eq:cubic_newton_single_index}
    \end{align}
\end{subequations}
where $\regular \mydefn (4 p - 1)!! (4 p - 1) p/ 3$.
Noting that the specific choice of $\regular$ 
in the formulation of the cubic-regularized Newton operator $\nopcnm$ 
arises because the second-order derivative of $\likelihoodsing_{n}$ 
is Lipschitz continuous with constant $\regular$. Similarly, the population-level operators are given by
\begin{subequations}
    \begin{align}
        \popgd( \tvar) & = \tvar - \learnrate \likelihoodsing' ( \tvar) 
        = \tvar \brackets{ 1 - (4 p - 1)!! (2 p) \learnrate \tvar^{4 p - 2}}, 
        \label{eq:pop_gd_single_idx}
        \\
        \popnm( \tvar) & = \tvar - \brackets{
        \likelihoodsing'' ( \tvar)}^{-1} \likelihoodsing' ( \tvar) 
        = \frac{(4 p - 2)}{4 p - 1} \tvar, \quad\text{and}
        \label{eq:pop_nm_single_idx} \\
        \popcnm (\tvar) 
        & = \mathop{ \arg \min}_{y \in \Rspace} 
        \left\{ \likelihoodsing' (\tvar) (y - \tvar) 
        + \frac{1}{2} \likelihoodsing'' (\tvar) (y - \tvar)^2 
        + \regular \abss{y - \tvar}^3 \right\}. \label{eq:pop_cnm_single_idx}
    \end{align}
\end{subequations}
\subsubsection{Proofs for the gradient descent operators} 
\label{ssub:proofs_for_the_gradient_descent_operator}

In order to achieve the conclusion of the corollary with convergence
rate of updates from gradient descent method, it is sufficient to
demonstrate that the sample gradient operator $\nopgd$ is $\stabilitynotag ( 2 p - 1)$-stable over the Euclidean ball $\ballnotag ( \tstar, 1)$ with noise function $\noise(n, \tol) = \frac{\log^{2 p} (n/ \tol)}{\sqrt{n}}$.  By using
the similar truncation argument as that
in equation~\eqref{eqn:sixth_key_bound}, we can verify the following
concentration bound
\begin{align}
\label{eq:key_con_bound_single_index}
\abss{ \frac{1}{n} \sum_{i = 1}^{n} Y_{i} X_{i}^{2 p}} \leq c
\log^{2 p}(n/ \tol)/ \sqrt{n},
\end{align}
with probability $1 - \tol$ where $c$ is some universal constant.  An application of triangle inequality
yields
\begin{align}
\label{eq:grad_single_index_bound}
\abss{\popga(\tvar) - \nopga(\tvar)} \leq \abss{\frac{1}{n}
  \sum_{i=1}^{n} X_{i}^{4 p} - (4 p - 1)!!}  \abss{ \tvar}^{4 p - 1} +
c \frac{\log^{2 p}(n/ \tol)}{\sqrt{n}} \abss{ \tvar}^{2 p - 1}.
\end{align}
Based on known concentration bounds for moments of Gaussian random
variables (cf. Lemma 5 in~\citep{Raaz_Ho_Koulik_2018_second}), we have
\begin{align}
\label{eq:moment_bound}
\abss{ \frac{1}{n} \sum_{i = 1}^{n} X_{i}^{4 p} - (4 p - 1)!!}
\leq c' \log^{2 p}(n/ \tol)/ \sqrt{n}
\end{align}
with probability $1 - \delta$ where $c'$ is some universal constant.  Substituting the inequality~\eqref{eq:moment_bound} into
equation~\eqref{eq:grad_single_index_bound} yields the
above claim with the stability of $\nopgd$.

\subsubsection{Proofs for the Newton operators} 
\label{ssub:proofs_for_the_newton_operators_index}

Moving to the convergence rates of updates from Newton's method, it is
sufficient to establish the instability of $\nopnm$ with respect to
$\popnm$, and moreover that, for any global minimum $\tvar_{n}^{*}$ of
the sample least-squares function $\likelihoodsing_{n}$ in
equation~\eqref{eq:sam_likeli_single_index}, we have
\begin{align}
\abss{ \nopnm( \tvar)} & \geq \abss{
  \tvar_{n}^{*}}, \label{eq:lower_single_index}
\end{align}
for all $\abss{ \tvar} \in [\abss{ \tvar_{n}^{*}}, 1]$. 


\paragraph{Instability of the sample Newton operator $\nopnm$}

Let us introduce the following shorthand notation:
\begin{align*}
A_{n} & \mydefn \parenth{\frac{2 p}{n} \sum_{i = 1}^{n} X_{i}^{ 4 p}}
\tvar^{4 p - 1} - \parenth{\frac{2 p}{n} \sum_{i = 1}^{n} Y_{i}
  X_{i}^{2 p}} \tvar^{2 p - 1}, \nonumber \\
B_{n} & \mydefn \parenth{\frac{2 p (4 p - 1)}{n} \sum_{i = 1}^{n}
  X_{i}^{4 p}} \tvar^{4 p - 2} - \parenth{\frac{2 p (2 p - 1)}{n}
  \sum_{i = 1}^{n} Y_{i} X_{i}^{2 p}} \tvar^{2 p - 2}.
\end{align*}
Applying the triangle inequality yields
\begin{align*}
\abss{\nopnm( \tvar) - \popnm( \tvar)} \leq \underbrace{\frac{\abss{(4
      p - 1)!! (2 p) \tvar^{4 p - 1} - A_{n}}}{(4 p - 1)!! (2 p) (4 p
    - 1) \tvar^{4 p - 2}}}_{\mydefn J_{1}} & \\ & \hspace{- 7 em} +
\underbrace{\abss{ A_{n}} \abss{\frac{1}{(4 p - 1)!! (2 p) (4 p - 1)
      \tvar^{4 p - 2}} - \frac{1}{B_{n}}}}_{\mydefn J_{2}}.
\end{align*}

\paragraph{Upper bound for $J_{1}$} 
Invoking triangle inequality, we obtain that
\begin{align*}
& \hspace{- 2 em} \abss{A_{n} - (4 p - 1)!! (2 p) \tvar^{4 p - 1}} \\
& \leq 2 p
\abss{\frac{1}{n} \sum_{i = 1}^{n} X_{i}^{4 p} - (4 p - 1)!!}  \abss{
  \tvar}^{4 p - 1} + \abss{ \frac{2 p}{n} \sum_{i = 1}^{n} Y_{i}
  X_{i}^{2 p}} \abss{ \tvar}^{2 p - 1} \\ & \leq c \frac{ \log^{2
    p}(n/ \tol)}{ \sqrt{n}} \parenth{ \abss{ \tvar}^{4 p - 1} + \abss{
    \tvar}^{2 p - 1}},
\end{align*}
where the last inequality is due to concentration bounds for moments
of Gaussian random
variables~\eqref{eq:key_con_bound_single_index}. With the above
inequality, we have
\begin{align}
\label{eq:final_upper_single_index_j1}
J_{1} \leq \frac{c \log^{2 p}(n/ \tol) \parenth{ \abss{ \tvar}^{4 p
      - 1} + \abss{ \tvar}^{2 p - 1}}}{(4 p - 1)!! (2 p) (4 p
    - 1) \sqrt{n} \abss{ \tvar}^{4 p -
    2}} \leq \frac{2c}{\abss{ \tvar}^{2 p - 1}} \frac{\log^{2 p}(n/
  \tol)}{\sqrt{n}},
\end{align}
for all $\abss{ \tvar} \leq 1$ with probability at least $1 - 2 \tol$.

\paragraph{Upper bound for $J_{2}$} In order to obtain 
an upper bound for $J_{2}$, we exploit the following concentration
bounds
\begin{subequations}
\begin{align}
\abss{A_{n}} \leq c_{1} \parenth{ \abss{ \tvar}^{4 p - 1} + \frac{\log^{2 p}(n/
  \tol)}{\sqrt{n}} \abss{ \tvar}^{2 p -
  1}}, \label{eq:upper_single_index_j2_first} \\
\abss{ B_{n} - (4 p - 1)!! (2 p) (4 p - 1) \tvar^{4 p - 2}} \leq
c_{2} \frac{\log^{2 p}(n/
  \tol)}{\sqrt{n}}, \label{eq:upper_single_index_j2_second} \\
\abss{ B_{n}} \geq c_{3} \parenth{ (4 p - 1)!! (2 p) (4 p - 1) \tvar^{4 p - 2} -
\unicon \frac{\log^{2 p}(n/
  \tol)}{\sqrt{n}}}, \label{eq:upper_single_index_j2_third}
\end{align}
\end{subequations}
for all $\abss{ \tvar} \leq 1$ with probability at least $1 - 2
\delta$.  Here, $\unicon, c_{1}, c_{2}, c_{3}$ are universal constants independent of
$\tol$.  The proofs of the above claims are direct applications of
triangle inequalities and concentration bounds we utilized earlier
with gradient descent operators in
Appendix~\ref{ssub:proofs_for_the_gradient_descent_operator};
therefore, they are omitted.  In light of the above bounds, we can
bound $J_{2}$ as follows:
\begin{align}
\label{eq:final_upper_single_index_j2}
J_{2} & \leq \frac{c_{1} c_{2}}{c_{3}} \parenth{ \abss{ \tvar}^{4 p - 1} + \frac{\log^{2
      p}(n/ \tol)}{\sqrt{n}} \abss{ \tvar}^{2 p - 1}} \nonumber \\
      & \hspace{6 em} \times
\frac{\frac{\log^{2 p}(n/ \tol)}{\sqrt{n}}} {\tvar^{4 p - 2} \parenth{
    (4 p - 1)!! (2 p) (4 p - 1) \tvar^{4 p - 2} - \unicon
    \frac{\log^{2 p}(n/ \tol)}{\sqrt{n}}}} \nonumber \\ & \leq \frac{2 c_{1} c_{2}}{c_{3} c}
\frac{1}{\abss{ \tvar}^{2 p - 1}} \frac{\log^{2 p}(n/ \tol)}{\sqrt{n}},
\end{align}
for all $\abss{ \tvar} \in [C \cdot \log^{p/ (2 p - 1)}(n/ \tol)/ n^{1/4
    (2 p - 1)}, 1]$ with probability $1 - 6 \tol$ where $C$ is solution of the equation $(4 p - 1)!! (2 p) (4 p - 1) \tvar^{4 p - 2} = 2 \unicon
    \frac{\log^{2 p}(n/ \tol)}{\sqrt{n}}$.  Combining the results
from equations~\eqref{eq:final_upper_single_index_j1}
and~\eqref{eq:final_upper_single_index_j2}, we achieve that
\begin{align}
\abss{\nopnm( \tvar) - \popnm( \tvar)} \leq c' \frac{1}{\abss{
    \tvar}^{2 p - 1}} \frac{\log^{2 p}(n/ \tol)}{\sqrt{n}}
\end{align}
for all $\abss{ \tvar} \in [C \log^{p/ (2 p - 1)}(n/ \tol)/ n^{1/4
    (2 p - 1)}, 1]$ with probability $1 - 8 \tol$ where $c'$ is some universal constant.  
    
As a consequence, the sample operator $\nopnm$ is $\instabilitynotag (- 2 p
    + 1)$-unstable over the annulus $\annulus (\tstar, c_{1} \log^{p/ (2 p - 1)}(n/ \tol)/ n^{1/4 (2 p - 1)},
    1)$ with noise function $\noise(\obs, \tol) = \dfrac{\log^{2
        p}(n/ \delta)}{\sqrt{n}}$.


\paragraph{Lower bound and monotonicity of Newton updates}

Moving to the claim~\eqref{eq:lower_single_index}, we first study the global minima $\tvar_{n}^{*}$ of the sample least-squares function
$\likelihoodsing_{n}$ in equation~\eqref{eq:sam_likeli_single_index}.  In
particular, they satisfy the equation $\nabla \likelihoodsing_{n}(
\tvar_{n}^{*}) = 0$, which is equivalent to
\begin{align*}
\parenth{\frac{1}{n} \sum_{i = 1}^{n} X_{i}^{4 p}} (\tvar_{n}^{*})^{4
  p - 1} - \parenth{\frac{1}{n} \sum_{i = 1}^{n} Y_{i} X_{i}^{2 p}}
(\tvar_{n}^{*})^{2 p - 1} = 0.
\end{align*}
Given the above equation, the specific form of $\tvar_{n}^{*}$ depends
on the sign of second derivative of $\likelihoodsing_{n}$ at 0.  In
particular, when $\sum_{i = 1}^{n} Y_{i} X_{i}^{2 p} > 0$, the
function $\likelihoodsing_{n}$ is bimodal and symmetric around 0.
Additionally, global mimima $\tvar_{n}^{*}$ have the form
\begin{align}
\label{eq:global_min_single_index}
\parenth{ \tvar_{n}^{*}}^{2 p} = \parenth{\frac{1}{n} \sum_{i = 1}^{n}
  Y_{i} X_{i}^{2 p}} \bigg/ \parenth{\frac{1}{n} \sum_{i = 1}^{n}
  X_{i}^{4 p}}.
\end{align}  
On the other hand, when $\frac{1}{n} \sum_{i = 1}^{n} Y_{i} X_{i}^{2
  p} \leq 0$, the function $\likelihoodsing_{n}$ is unimodal and
symmetric around 0.  Furthermore, it has only global minimum
$\tvar_{n}^{*} = 0$.

Now, we focus on the case $\tvar > 0$ and $\sum_{i = 1}^{n} Y_{i} X_{i}^{2
  p} > 0$, i.e., the global minima $\tvar_{n}^{*}$ are different from
0 and the solutions of equation~\eqref{eq:global_min_single_index}.  A simple
calculation demonstrates that $B_{n} > 0$ and $\nopnm (\tvar) > 0$ as
long as $\tvar > \abss{ \tvar_{n}^{*}}$.  Now, the inequality $\nopnm(
\tvar) \geq \abss{ \tvar_{n}^{*}}$ is equivalent to
\begin{align*}
\parenth{\frac{4 p - 2}{n} \sum_{i = 1}^{n} X_{i}^{4 p}} \tvar^{2 p +
  1} + \parenth{ \frac{2 p - 1}{n} \sum_{i = 1}^{n} Y_{i} X_{i}^{ 2
    p}} \abss{ \tvar_{n}^{*}} \\
    \hspace{- 3 em} \geq \parenth{ \frac{4 p - 1}{n} \sum_{i
    = 1}^{n} X_{i}^{4 p}} \tvar^{2 p} \abss{ \tvar_{n}^{*}} + \parenth{ \frac{2 p - 2}{n} \sum_{i = 1}^{n}
  Y_{i} X_{i}^{ 2 p}} \tvar
\end{align*}
for $\tvar \geq \abss{ \tvar_{n}^{*}}$.  In light of the closed form
expression of $\abss{ \tvar_{n}^{*}}$
in equation~\eqref{eq:global_min_single_index}, a simple algebra with the above
inequality leads to the inequality
\begin{align*}
(4 p - 2) \tvar^{2 p + 1} + (2 p - 1) \abss{ \tvar_{n}^{*}}^{2 p + 1}
  \geq (2 p - 2) \parenth{ \tvar_{n}^{*}}^{2 p} \tvar + (4 p - 1)
  \abss{ \tvar_{n}^{*}} \tvar^{2 p},
\end{align*}
which holds true due to AM-GM inequality.  Thus, we have established
the claim~\eqref{eq:lower_single_index}.


\subsubsection{Proofs for the cubic-regularized Newton operators}
\label{ssub:corollary:cubic_newton_single_index}

Our proof is divided into three separate steps. First, we establish
the slow convergence of operator $\popcnm$. Then, we proceed to
establishing the instability of operator $\nopcnm$. Finally, we demonstrate the monotonicity of
cubic-regularized Newton updates and their lower bound
\begin{align}
\label{eq:lower_cubic_single_index}  
\abss{ \nopcnm( \tvar)} & \geq \abss{ \tvar_{n}^{*}},
\end{align}
for all $\abss{ \tvar} \in [\abss{ \tvar_{n}^{*}}, 1]$ for any global
minima $\tvar_{n}^{*}$ of the sample least-squares function
$\likelihoodsing_{n}$ in equation~\eqref{eq:sam_likeli_single_index}.


\paragraph{Slow convergence of $\popcnm$}

Without loss of generality, we assume that $\tvar \in (0, 1]$.  Direct
  computation leads to
\begin{align*}
\popcnm(\tvar) & = \tvar + \tvar^{4 p - 2} - \sqrt{ \tvar^{8 p - 4} +
  \frac{2}{4 p - 1} \tvar^{4 p - 1}} \\ & = \tvar - \frac{\frac{2}{4 p
    - 1} \tvar^{4 p - 1}}{ \tvar^2 + \sqrt{ \tvar^{8 p - 4} +
    \frac{2}{4 p - 1} \tvar^{4 p - 1}}} \leq \tvar \parenth{1 - c_{1}
  \tvar^{(4 p - 3)/ 2}},
\end{align*}
for any $\tvar \in (0, 1]$ where $c_{1} < 1$ is some universal
  constant.  As a consequence, the operator $\popcnm$ satisfies slow
  convergence condition $\slownotag(2/ (4 p - 3))$ over the Euclidean
  ball $\ballnotag (\tstar, 1)$.

  
\paragraph{Instability of the sample operator $\nopcnm$}

Suppose that $\tvar > \abss{\tvar_{n}^{*}}$, where $\tvar_{n}^{*}$ are
global minima of the sample least-squares function
$\likelihoodsing_{n}$.  With this condition, direct computation of
$\nopcnm( \tvar)$ leads to
\begin{align*}
\nopcnm(\tvar) & = \tvar - \frac{ 2 \likelihoodsing_{n}' ( \tvar)}{
  \likelihoodsing_{n}'' (\tvar) + \sqrt{ \parenth{
      \likelihoodsing_{n}'' (\tvar)}^2 + 12 \regular \cdot
    \likelihoodsing_{n}' (\tvar)}} \mydefn \tvar - \frac{ 2
  \likelihoodsing_{n}' ( \tvar)}{ T_{n}}.
\end{align*}
Similar to the previous proofs with cubic-regularized Newton
operators, we find that
\begin{align*}
\abss{ \popcnm( \tvar) - \nopcnm (\tvar)} & \leq 2 \frac{ 
  \likelihoodsing' ( \tvar) \abss{T_{n} - T} + T \abss{ 
    \likelihoodsing_{n}' ( \tvar) -  \likelihoodsing' ( \tvar)}}{T
  T_{n}},
\end{align*}
where $T \mydefn  \likelihoodsing'' (\tvar) + \sqrt{
  \parenth{ \likelihoodsing'' (\tvar)}^2 + 12 \regular \cdot 
  \likelihoodsing' (\tvar)} \geq \sqrt{ 12 \regular \likelihoodsing' (
  \tvar)} \geq C \cdot \tvar^{(4 p - 1)/ 2}$ for some universal constant $C > 0$. Additionally, we have
\begin{align*}
\abss{T_{n} - T} \leq c' \cdot \tvar^{-1/ 2} \frac{\log^{2 p}(n/ \tol)}{ \sqrt{n}}
\end{align*}
when $\tvar \geq c \cdot \max \left\{ \abss{ \tvar_{n}^{*}},
\frac{\log^{p/ (2 p - 1)} (n/ \delta)}{n^{1/ 4 (2 p - 1)}}\right\}$
with probability $1 - 10 \delta$ for some universal constants $c$ and $c'$. Furthermore, we can check that
$T_{n} \geq \sqrt{12 \regular \cdot \likelihoodsing_{n}' ( \tvar)} \geq
c'' \tvar^{(4 p - 1)/ 2}$ as long as $\tvar \geq c \cdot \max \left\{ \abss{
  \tvar_{n}^{*}}, \frac{\log^{p/ (2 p - 1)} (n/ \delta)}{n^{1/ 4 (2 p
    - 1)}}\right\}$ with probability $1 - 2 \delta$ for some universal constant $c''$.  These
inequalities guarantee that
\begin{align*}
\abss{ \popcnm( \tvar) - \nopcnm (\tvar)} \leq c_{1} \tvar^{-1/ 2}
\frac{\log^{2 p}(n/ \tol)}{ \sqrt{n}}
\end{align*}
for all $\tvar \geq c \cdot \max \left\{ \abss{ \tvar_{n}^{*}},
\frac{\log^{p/ (2 p - 1)} (n/ \delta)}{n^{1/ 4 (2 p - 1)}}\right\}$
with probability $1 - 14 \delta$. As a consequence, we conclude that
the operator $\nopcnm$ is $\instabilitynotag( - 1/ 2)$-unstable over the annulus $\annulus( \tstar, c
\frac{\log^{p/ (2 p - 1)} (n/ \delta)}{n^{1/ 4 (2 p - 1)}},
1)$ with noise function $\noise = \frac{\log^{2
    p} (n/ \delta)}{\sqrt{n}}$ where $c$ is some universal constant.

\paragraph{Lower bound and monotonicity of cubic-regularized Newton updates}
To simplify the presentation, we only consider $\tvar > 0$ and the
setting when global minima $\tvar_{n}^{*}$ are different from 0. As
$\tvar \geq \abss{ \tvar_{n}^{*}}$, the inequality $\nopcnm( \tvar)
\geq \abss{ \tvar_{n}^{*}}$ is equivalent to
\begin{align*}
\likelihoodsing_{n}'' (\tvar) + \sqrt{ \parenth{
    \likelihoodsing_{n}'' (\tvar)}^2 + 12 \regular
  \likelihoodsing_{n}' (\tvar)} > 2 \likelihoodsing_{n}''
(\widetilde{ \tvar})
\end{align*}
for some $\widetilde{ \tvar} \in (\abss{ \tvar_{n}^{*}}, \tvar)$. This
inequality holds since $ \likelihoodsing_{n}'$ and $
\likelihoodsing_{n}''$ are positive and strictly increasing when $\tvar
> \abss{ \tvar_{n}^{*}}$, thereby completing the proof of claim~\eqref{eq:lower_cubic_single_index}.


\section{Extension to multivariate settings}
\label{sec:extension_multivariate}

In this appendix, we discuss some extensions of the theoretical
results in Section~\ref{sec:specific_models} to multivariate
settings. Here we state detailed theoretical results for the EM
algorithm and gradient descent for multivariate versions of the
over-specified mixture model and the non-linear regression model. We
explore the behavior of Newton's method via experimental studies in
both Figures~\ref{FigEMEmpirical} and~\ref{FigPREmpirical}.

\subsection{Over-specified mixture model} 
\label{sec:multivariate_mixture}

We denote by $\normDensity(\cdot; \tvar, \sd^2 I_{d})$ the density of
$\NORMAL(\tvar, \sd^2 I_{d})$ random variable, i.e.,
\begin{align*}
\normDensity(x; \tvar, \sd^2 I_{d}) = (2 \pi \sd^2)^{- d/ 2}e^{-
  \frac{\enorm{x - \tvar}^2}{2 \sd^2}}.
\end{align*}
Assume that $X_{1},\ldots,X_ {n}$ be $n$ are i.i.d. samples from
$\NORMAL(0, I_{d})$.  We then fit a two-component symmetric Gaussian
mixture with equal fixed weights whose density is given by:
\begin{align}
\label{Eqnsingular_location_scale_multivariate}
    \FitDensity(x) = \frac{1}{2}\normDensity(x;-\tvar, I_{d}) +
    \frac{1}{2}\normDensity(x;\tvar, I_{d}),
\end{align}
where $\tvar \in \Rspace^{d}$ is the parameter to be estimated. Given
the model, the true parameter is unique and given by
$\thetastar=0$. Similar to the univariate setting in
Section~\ref{sub:over_specified_gaussian_mixture_models}, we also use
the EM algorithm to estimate $\thetastar=0$. Direct calculation of the
sample EM operator yields that
\begin{align*}
\nopem(\tvar) = \frac{1}{n} \sum_{i = 1}^{n} X_{i} \tanh(\tvar^{\top}
X_{i}),
\end{align*}
where $\tanh(x) = \frac{\exp( x) - \exp( - x)}{ \exp( x) + \exp( -
  x)}$ for all $x \in \Rspace$. The result characterizing the behavior
of sample EM operator in the multivariate
setting~\eqref{Eqnsingular_location_scale_multivariate} is already
proven in our prior work~\citep{Raaz_Ho_Koulik_2018} (see Theorem 3 in
that paper). So as to keep our discussion self-contrained, we restate
it here:
\begin{corollary}
\label{eq:multivariate_EM}
For the over-specified Gaussian mixture
model~\eqref{Eqnsingular_location_scale_multivariate} with $\tstar =
0$, given some $\tol \in (0,1)$ and for any fixed $\smallthreshold \in
(0, 1/4)$ and initialization $\tvar^0 \in \ballnotag (\tstar, 1 )$,
with probability at least $1 - \tol$ the sequence $\tvar^\iter \defn
(\nopem)^\iter(\tvar^0)$ of EM iterates satisfies the bound
\begin{align*}
  \enorm{\tvar^\iter - \tstar } & \leq c_{1} \parenth{\frac{d + \log(
      \frac{\log(1/\smallthreshold)}{\tol})}
    {\obs}}^{\frac{1}{4}-\smallthreshold} \qquad \mbox{for all
    iterates $\iter \geq c_{1}' \sqrt{\frac{\obs}{d}} \log\frac{1}
    {\smallthreshold}$,}
\end{align*}
as long as $n \geq c_1'' (d +
\log\frac{\log(1/\smallthreshold)}{\tol})$.
\end{corollary}
The result of Corollary~\ref{eq:multivariate_EM} shows that the EM
iterates converge to a radius of convergence $(d/ n)^{1/4}$ around the
true parameter $\tstar = 0$ after $\sqrt{n/ d}$ number of
iterations. Note that our simulation results for EM, as shown in
Figure~\ref{FigEMEmpirical}, are consistent with this theoretical
prediction.

\subsection{Non-linear regression model} 
\label{sec:multivariate_nonlinear_regression}

We now turn to the multivariate instantiation of the non-linear
regression model considered in the main text.  Suppose that we observe
pairs $(X_{i}, Y_{i}) \in \real^{d} \times \real$ generated from the
model
\begin{align}
\label{eq:multivariate_single_index}
Y_{i} = g \parenth{X_{i}^{\top} \tvar^{*}} +
\newnoise_{i}\quad\text{for }\quad \quad i = {1, \ldots, \obs},
\end{align}
where $\newnoise_{i} \sim \NORMAL(0, 1)$.

We assume that the covariate vectors $X_{i}$ are drawn i.i.d. from the
multivariate Gaussian $\NORMAL(0, I_{d})$. As in our study of the
univariate case, we consider the family of link functions $g(x) =
x^{2p}$ for $p \geq 1$ and the unknown parameter $\tvar^{*} =
\bold{0}$.  With this set-up, the maximum likelihood estimate for
$\tvar^{*}$ is based on the minimization problem
\begin{align}
\min_{\theta \in \real^{d}} \mullikelihoodsing_{n} (\tvar)
\quad\text{where} \quad \mullikelihoodsing_{n} ( \tvar) \mydefn
\frac{1}{2 \obs} \sum_{i = 1}^{\obs} \parenth{Y_{i} -
  \parenth{X_{i}^{\top} \tvar}^{2
    p}}^2. \label{eq:multivariate_objective_single_index}
\end{align}
By taking the expectation of $\mullikelihoodsing_{n}$ with respect to
$X_{1}, \ldots, X_{n} \sim \NORMAL(0, I_{d})$, we find that the
corresponding population version of $\mullikelihoodsing$ takes the
form
\begin{align}
\mullikelihoodsing (\tvar) \mydefn \frac{1}{2} \Exs
\brackets{\parenth{Y - \parenth{X^{\top} \theta}^{2p}}^{2}} = \frac{1
  + (4p - 1)!! \enorm{\tvar -
    \tvar^{*}}^{4p}}{2}. \label{eq:pop_multivariate_objective_single_index}
\end{align}
The sample operator for the gradient method is given by
\begin{align}  
  \nopgd(\tvar) & = \tvar - \learnrate \nabla
  \mullikelihoodsing_{n}(\tvar) \nonumber \\
\label{eq:multi_grad_single_index}  
  & = \tvar - \learnrate \bigg(\frac{2p}{n} \sum_{i = 1}^{n} X_{i}
(X_{i}^{\top} \tvar)^{4p - 1} - \frac{2 p}{n} \sum_{i = 1}^{n} Y_{i}
X_{i} (X_{i}^{\top} \tvar)^{2p - 1} \biggr),
\end{align}
whereas the population level operator corresponding to the operator
$\nopgd$ takes the form
\begin{align}
\label{eq:multi_pop_grad_single_index}  
  \popgd(\tvar) & = \tvar - \learnrate \nabla
  \mullikelihoodsing(\tvar) = \tvar \parenth{1 - (4p - 1)!! 2 p
    \learnrate \enorm{\tvar}^{4p - 2}},
\end{align}
where $I_{d}$ denotes the identity matrix in $d$ dimension.

\begin{figure}[t!]
\begin{adjustbox}
{width=0.95\textwidth,center=\textwidth} 
  \begin{tabular}{cc}
    \widgraph{0.5\textwidth, trim={0, 0, 0, 0}, clip}{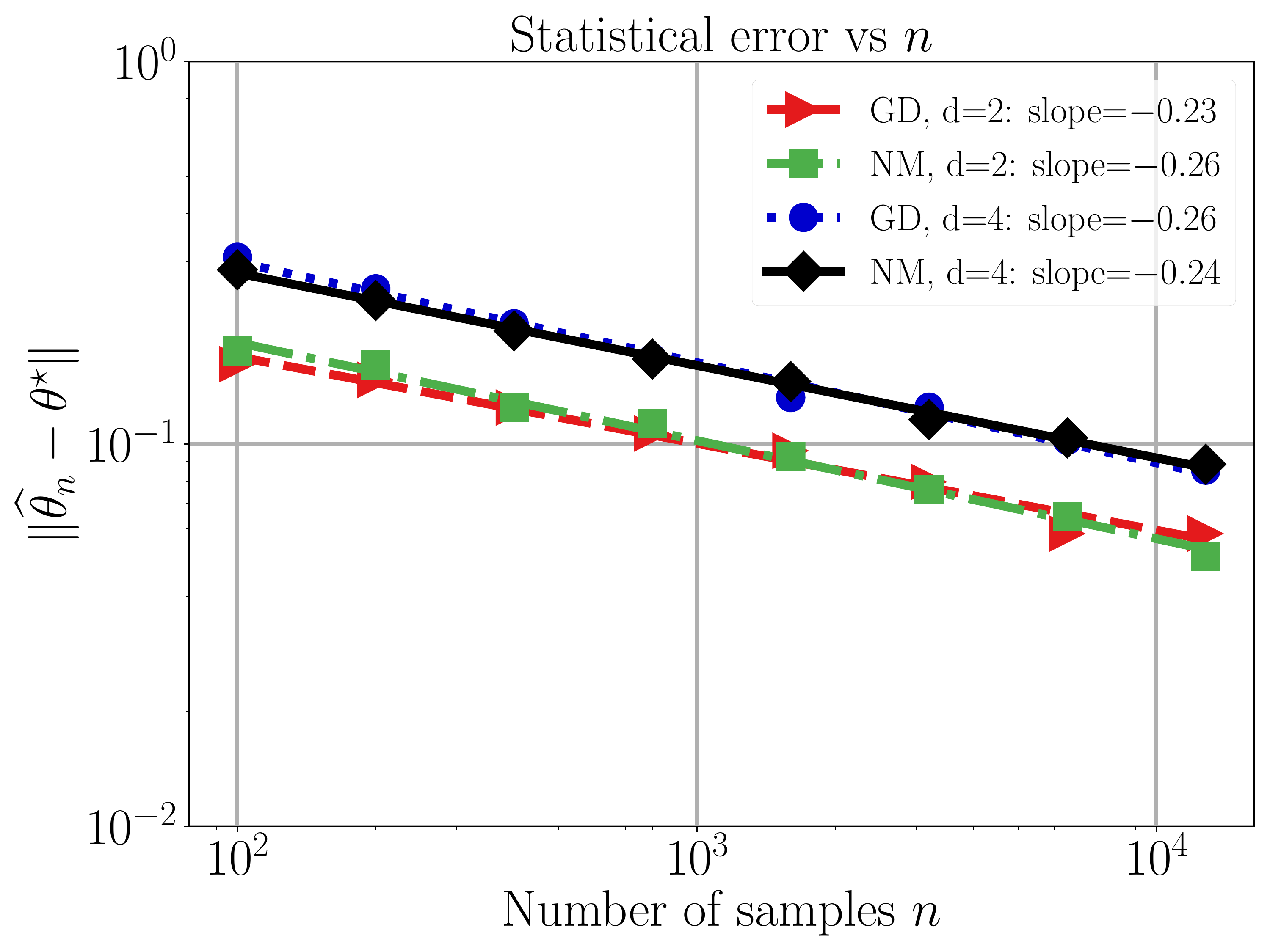}
    &
    \widgraph{0.5\textwidth, trim={0, 0, 0, 0}, clip}{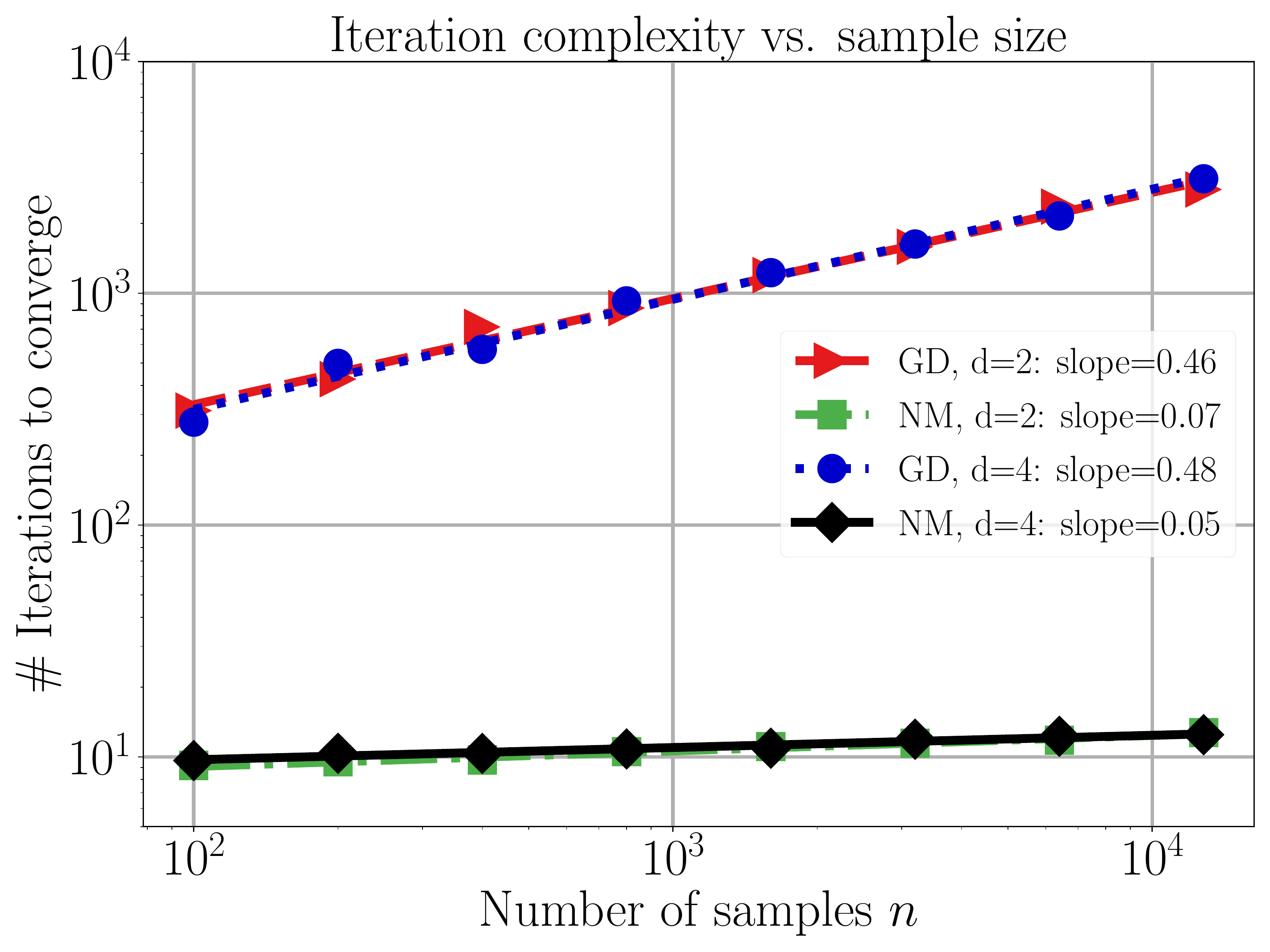} \\ (a) &
    (b)
  \end{tabular}
  \end{adjustbox}
  \caption{Plots characterizing the behavior of Gradient Descent (GD) and Newton's method (NM) for the non-linear regression with $p=1$ for $d=2$ and $d=4$ dimensions. (a) Log-log plots of the Euclidean
    distance $\|\widehat{\theta}_\obs - \tstar \|_2$ versus the sample
    size.  It shows that all the algorithms converge to an estimate at
    Euclidean distance of the order $n^{-1/4}$ from the true parameter
    $\tstar$.  (b) Log-log plots for the number of iterations taken by
    different algorithms to converge to the final estimate.}
  \label{FigPREmpirical}
\end{figure}

We first state a result concerning the contraction and stability
properties of the population and sample operators $\popgd$ and
$\nopgd$.
\begin{lemma}
\label{lemma:contraction_multivariate_index}
(a) For any step size $\learnrate \in (0, \frac{1}{(4 p - 1)!!(2
  p)}]$, the gradient operator $\popgd$ is
$\slownotag(\frac{1}{4p-2})$-convergent over the ball $\ball(\tstar,
1)$. \\
(b) The operator $\nopgd$ is $\stabilitynotag (2 p - 1)$-stable over
the ball $\ball(\tstar, 1)$ with noise function $\noise(n, \delta) =
\sqrt{\frac{d + \log(1/ \delta)}{n}}$.
\end{lemma}
\noindent
The proof of Lemma~\ref{lemma:contraction_multivariate_index} is deferred to the end of this appendix. Based on the result of that lemma, we have the following result characterizing the behavior of the updates from the gradient descent algorithm for solving $\mullikelihoodsing_{n}$.
\begin{corollary}
\label{eq:multivariate_nonlinear}
For the non-linear regression
model~\eqref{eq:multivariate_single_index} with $\tstar = 0$, given
some \mbox{$\tol \in (0,1)$} and for any fixed \mbox{$\smallthreshold
  \in (0, 1/4)$} and initialization $\tvar^0 \in \ballnotag (\tstar, 1
)$, with probability at least $1 - \tol$ the sequence $\tvar^\iter
\defn (\nopgd)^\iter(\tvar^0)$ generated by gradient descent satisfies
the bound
\begin{align*}
  \enorm{\tvar^\iter - \tstar } & \leq c_{1} \parenth{\frac{d + \log(
      \frac{\log(1/\smallthreshold)}{\tol})}
    {\obs}}^{\frac{1}{4p}-\smallthreshold} \ \ \mbox{for all iterates
    $\iter \geq c_{1}' \parenth{\frac{\obs}{d}}^{\frac{2p-1}{2p}}
    \log\frac{1} {\smallthreshold}$,}
\end{align*}
as long as $n\geq c_1'' (d +
\log\frac{\log(1/\smallthreshold)}{\tol})^{4p}$.
\end{corollary}
Based on the result of Corollary~\ref{eq:multivariate_nonlinear}, the
updates from the gradient method converge to a ball of radius of the
order of $(d/n)^{1/4p}$ around the true parameter $\tstar = 0$ after
an order of $(n/d)^{(2p - 1)/2p}$ number of iterations. We further
illustrate these behaviors of the gradient method when $p = 1$ in
Figure~\ref{FigPREmpirical}. Based on these results, the computational
complexity of the gradient method is at the order of
$n^{\frac{4p-1}{2p}}d^{\frac{1}{2p}}$.
For the Newton's method, the experimental results in
Figure~\ref{FigPREmpirical} show that the Newton iterates also
converge to the similar radius of convergence $(d/n)^{1/4p}$ after
$\log(n)$ number of iterations. Since each iteration of the Newton's
method takes an order of $n \cdot d + d^3$ arithmetic operations where
$d^3$ is computational complexity of computing inverse of an $d \times
d$ matrix via Gauss-Jordan elimination, the overall complexity
required to reach to the final estimate scales as $(n d+ d^3) \log
n$. Thus, when $d^{\frac{6p - 1}{4p - 1}} \ll n$, Newton's method is
computationally more efficient than the gradient descent method.

\subsubsection{Proof of Lemma~\ref{lemma:contraction_multivariate_index}}

The slow contraction of the population gradient operator $\popgd$
follows immediately from its definition. Furthermore, the proof of the
stability of the sample operator $\nopgd$ follows from the
concentration bound in Corollary 3 in~\citep{mou_diffusion}. In fact,
from the proof of Corollary 3 in~\citep{mou_diffusion}, as long as $r
\leq 1$ we have
\begin{align*}
\sup_{\theta \in \ball(\tvar^{*}, r)} \abss{\nopgd (\tvar) - \popgd
  (\tvar)} \leq C r^{2p - 1} \sqrt{\frac{d + \log(1/ \delta)}{n}},
\end{align*}
as long as $n \geq C' (d + \log(d/ \delta))^{4 p}$ where $C$ and $C'$
are some universal constants. As a consequence, we obtain the
conclusion of the lemma with the contraction and stability of the
operators $\popgd$ and $\nopgd$.







\bibliography{Nhat}
\end{document}